\documentclass[lettersize,journal]{IEEEtran}
\usepackage{amsmath,amsfonts}
\usepackage{algorithmic}
\usepackage{algorithm}
\usepackage{array}
\usepackage[caption=false,font=normalsize,labelfont=sf,textfont=sf]{subfig}
\usepackage{textcomp}
\usepackage{stfloats}
\usepackage{url}
\usepackage{verbatim}
\usepackage{graphicx}
\usepackage{cite}
\usepackage{bookmark}
\usepackage[english]{babel}
\usepackage{xcolor}
\usepackage[numbers]{natbib}
\usepackage{float}
\usepackage{multirow}
\usepackage{placeins}
\usepackage{amssymb}
\usepackage{chngcntr}
\usepackage{authblk}

\graphicspath{ {./high-res-fig/} }

\begin{document}

\title{A comprehensive survey on semantic facial attribute editing using generative adversarial networks}


\author[1]{Ahmad Nickabadi}
\author[1]{Maryam Saeedi Fard}
\author[1]{Nastaran Moradzadeh Farid}
\author[1]{Najmeh Mohammadbagheri}

\affil[1]{Computer Engineering Department, Amirkabir University of Technology, Tehran, Iran \authorcr {\tt \{nickabadi, m.saeedifard, nmoradzadehf, n.m.bagheri77\}@aut.ac.ir}\vspace{1.5ex}}



\maketitle

\begin{abstract}
Generating random photo-realistic images has experienced tremendous growth during the past few years due to the advances of the deep convolutional neural networks and generative models. Among different domains, face photos have received a great deal of attention and a large number of face generation and manipulation models have been proposed. Semantic facial attribute editing is the process of varying the values of one or more attributes of a face image while the other attributes of the image are not affected. The requested modifications are provided as an attribute vector or in the form of driving face image and the whole process is performed by the corresponding models.   
In this paper, we survey the recent works and advances in semantic facial attribute editing. We cover all related aspects of these models including the related definitions and concepts, architectures, loss functions, datasets, evaluation metrics, and applications. Based on their architectures, the state-of-the-art models are categorized and studied as encoder-decoder, image-to-image, and photo-guided models. The challenges and restrictions of the current state-of-the-art methods are discussed as well.
\end{abstract}

\begin{IEEEkeywords}
Facial Attribute Editing, Generative Models, Generative Adversarial Networks, Encoder-Decoder, Deep Neural Networks, Convolutional Neural Networks 
\end{IEEEkeywords}

\section{Introduction}

\begin{figure*}[htb]
\centering
\subfloat[]{\includegraphics[width=2.5in]{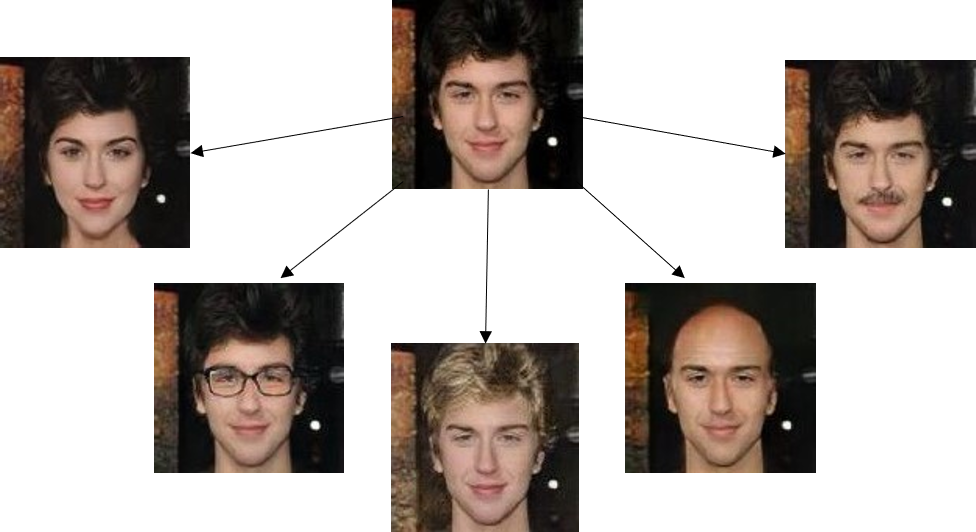}%
\label{fig_first_case}}
\hfil
\subfloat[]{\includegraphics[width=2.5in]{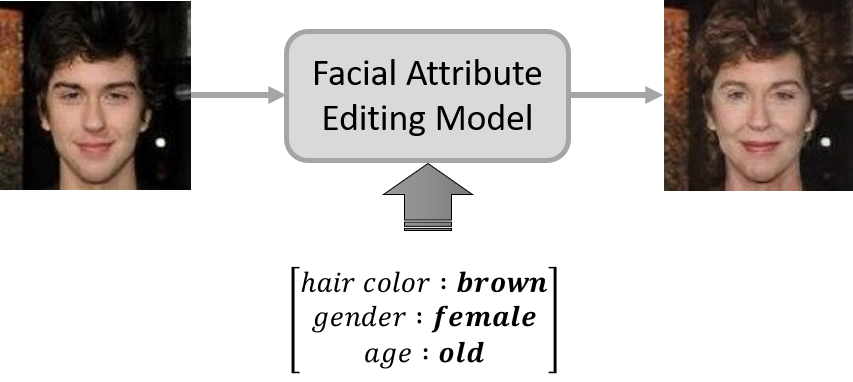}%
\label{fig_second_case}}
\caption{Facial attribute editing. (a) Examples of attribute editings applied to a face image. (b) The overall structure of a face attribute editing model.}
\label{fig:definition}
\end{figure*}

Facial attribute editing is the task of manipulating single or multiple attributes of a face image while the other attributes remain untouched and has found applications in many areas including data augmentation in facial image processing tasks,  adding effects to facial images in the social networks, improving the performance of automatic face recognition systems, creating and animating faces in game and animation industries. Examples of facial attribute editing are shown in Figure \ref{fig:definition}. As shown in this figure, the aim of the overall process is to change the values of some target attributes of the input face image, e.g. to change the sex, age, skin color or add glasses to the image. 

The face of a human can be digitally represented in many forms including 1) 2D images of grayscale or RGB pixels, 2) parametric or nonparametric 3D models, 3) low-level feature vectors extracted by image processing models, and 4) high-level descriptors given as the values of some facial attributes (e.g. gender, age, hair color, and face shape). Face editing may happen at any of these representation levels. For example, while it is possible to make a face look older by manually adding some wrinkles to the face image in a drawing software or application, it is also possible to do this by requesting a high-level semantic facial attribute manipulation model to just increase the age. The focus of this survey is on the semantic facial attribute editing models in which the required changes are given as the values of the target attributes (e.g. “the hair color should be brown”) or in the form of a driving face image with the desired attribute values and the whole image modification task is then performed by the model without any user intervention. In the past few years, the progress of deep learning and generative models has revolutionized the semantic image manipulation models.

Generative and discriminative models are two distinct branches of machine learning \cite{wang2017generative}. On one hand, generative models learn the joint probability \(p(x,y)\) of the input \(x\) and the intended corresponding information (such as label) \(y\). A generative model models the actual distribution of observations and can sample new random data from the estimated distribution. On the other hand, discriminative methods compute the posterior \(p(y|x)\) and learn a direct map from input \(x\) to the target \(y\). In general, a discriminative method models the decision boundary between the classes and is commonly used as a classifier \cite{jordan2002discriminative}.

The above division is applicable to the facial image processing techniques, too. Facial image processing models can be divided into two major groups: facial analysis methods that result in some information about a face image and facial manipulation methods that generate new face images. Given an input face image, facial analysis methods estimate some attributes of the input face, such as age, identity, gender, face shape, etc. On the other side, facial manipulation aims to generate a face image, either a random face from an input noise \cite{karras2019style} or a modified face from a given image\cite{pumarola2018ganimation}. 
As mentioned earlier, a face editing model takes a face image and a set of desired attribute values as input and generates an edited version of the face image with the requested attributes. There are various approaches for doing this with generative adversarial networks which are thoroughly studied in the rest of this paper. 

Generative facial attribute editing models vary in three ways: 1) the attributes that are (can be) edited by the model, 2) the way the desired values of the target attributes are expressed, and 3) the generation process used to produce the modified face image.

Regarding the attributes of a face image that can be edited, several categorizations have been proposed so far. According to \cite{chandaliya2019scdae}, facial attributes are generally categorized into global and local attributes. A global attribute relates to the whole area of the input image. On contrary, to obtain the value of a local attribute, a specific part of the face should be considered. For example, age is a global attribute that is determined by looking at the image globally, while hair color is a local attribute that its value is determined by the hair part of the image.  Chandaliya et al. \cite{chandaliya2019scdae} introduce another categorization for facial attributes. They subdivide these attributes into categorical and binary attributes based on the number of possible values of an attribute. Based on this categorization, hair color is a categorical attribute that can have multiple values (e.g., Brown, Blond, and Black), and gender is a binary attribute that has only two values (Male and Female). 
Facial attributes can also be classified as continuous or discrete. The values of a continuous attribute interpolate between two values while discrete attributes can only adopt a finite set of values. Finally, facial attributes are either identity-relevant or identity-irrelevant. Identity-relevant attributes are those attributes that uniquely determine a person's face and are not affected by the expression or pose of the face. For example, smiling is a continuous identity-irrelevant attribute while ethnicity is a discrete identity-relevant attribute. 

The desired values of the target attributes are determined in three different ways: 1) an attribute vector, 2) a textual description of the target attributes, and 3) a driving image. Moreover, as stated in \cite{zheng2020survey}, facial manipulation methods can be divided into single and multiple attribute editing methods. For example, \cite{fang2020triple} works only on age editing and \cite{lee2020byeglassesgan} only removes eyeglasses from face images, while RelGAN \cite{wu2019relgan} can simultaneously edit hair color, age and gender attributes.
However, in both cases, the desired attribute(s) can be fed into the model in all of the aforementioned ways. 
An attribute vector is a vector of elements, each of which corresponds to a perceptible facial attribute. To specify an attribute, the related element is accordingly set to the desired value. It is also possible to set several values, simultaneously, in the case of multiple attribute editing.

A human-defined description is a high-level textual description of a face usually provided by a human. In this case, a natural language processing module is required to parse the input text and extract the attribute values in the form of an attribute vector. 
For example, Vijay et al. \cite{vijay2020dialog} have proposed a facial attribute editing model using a speech-based description of the face and Liu et al. \cite{liu2020describe} utilize a textual description in their model. 
A driving image is a face image with the desired attributes whose values are to be extracted and applied to the target face image-a process which is also known as style transfer \cite{park2020swapping}. 

The most important aspect of a face attribute manipulation method is the way the attributes are edited. While some researchers have attempted to train supervised deep convolutional neural networks for editing face attributes\cite{yim2015rotating,zhang2018face}, semantic face editing methods are generally based on generative models including Variational Autoencoders (VAEs) and Generative Adversarial Networks (GANs). 

Variational autoencoder is a special case of encoder-decoder architecture in which the encoder learns the posterior distribution $p(z|x)$ and encodes the input image $x$ to a latent space variable $z$. Decoder samples new images from the latent vector $z$ and generates new samples similar to the input image. Nie et al. \cite{nie2020improved} use semi-supervised learning based on a VAE model to edit attributes of face images. 

However, most of the current facial attribute modification methods are based on GANs \cite{goodfellow2014generative}. These models consist of a generator network \(G\) and a discriminator network \(D\) playing a mini-max game against each other. The generator aims to learn the distribution of input data and generates new samples imitating training samples. On contrary, the discriminator learns to distinguish the real samples from the outputs of the generator. GANs are used in different domains, but image synthesis is the origin and one of the best targets of these models. 
To merge the advantages of both models, some methods such as \cite{huang2018introvae} use a hybrid model that combines VAEs and GANs.

In this survey, we present a comprehensive study on existing semantic facial attribute editing methods, especially focusing on the GAN-based models. Compared to the other similar surveys, the distinguishing characteristics of this survey are as follows: 
\begin{itemize}
    \item The current study is completely focused on semantic facial attribute editing task and has deeply and thoroughly reviewed the works in this field. 
    \item Due to the high quality of the face images generated by GAN models in recent years, image editing and especially semantic facial attribute editing have experienced a rapid growth and many new works have been published in the past few years.  In this survey, we have tried to cover all of the novel methods of this area. 
    \item In this survey, we provide exact definitions of the related concepts and a relatively novel grouping of the face image attributes (Section \ref{sec2}), a novel categorization and a comprehensive study of the architectures of different models (Section \ref{sec3}),  a review of the applications of the related methods (Section \ref{sec5}) and finally a summary of the datasets and evaluation metrics (Section \ref{sec4}).
\end{itemize} 

In the following, Section \ref{sec2} introduces preliminaries on basic concepts and definitions related to semantic facial attribute editing. In Section \ref{sec3}, we survey the existing methods from architecture and loss function perspectives. Section \ref{sec4} is dedicated to discussing datasets and evaluation metrics used in facial attribute editing solutions. And, at the last, we investigate the facial attribute editing applications in Section \ref{sec5}. Finally, Section \ref{sec6} concludes the paper.

\section{Preliminaries} \label{sec2}

As mentioned earlier, semantic facial attribute editing refers to altering some human-perceptible facial attributes while the others remain unchanged by just describing the values of the target attributes. 
Two main questions about facial attribute editing are: 1) what are these target attributes? and 2) how are they edited? In this section, we answer the first question by providing a comprehensive list of facial attributes that has been the subject of edit in the related literature and in the following sections we will review the GAN-based methods proposed for this purpose. However, as there are other tasks and applications in computer vision field that are similar to semantic facial attribute editing in some aspects, we provide the exact definitions of these tasks in this section and delineate their relation to semantic facial attribute editing.
Finally, at the end of this section, some related surveys are reviewed and compared with this survey to see how it is different from previous studies. 

\begin{table*}[!htbp]
\centering
\caption{The set of facial attributes used in the literature of semantic facial attribute editing.}
\label{tab:attributes}
\resizebox{\textwidth}{!}{%
\begin{tabular}{|c|cc|c|c|c|c|}
\hline
\textbf{Category} &
  \multicolumn{2}{c|}{\textbf{Attribute}} &
  \textbf{Value Type} &
  \textbf{Values} &
  \textbf{Global/Local} &
  \textbf{Attribute Type} \\ \hline
\multirow{22}{*}{Face Parts} &
  \multicolumn{1}{c|}{\multirow{6}{*}{hair}} &
  hair type &
  categorical &
  curly, wavy, straight &
  local &
  identity \\ \cline{3-7} 
 &
  \multicolumn{1}{c|}{} &
  hair color &
  categorical &
  black, blond, brown, gray &
  local &
  identity \\ \cline{3-7} 
 &
  \multicolumn{1}{c|}{} &
  bang &
  binary &
  0/1 &
  local &
  identity \\ \cline{3-7} 
 &
  \multicolumn{1}{c|}{} &
  bald &
  binary &
  0/1 &
  local &
  identity \\ \cline{3-7} 
 &
  \multicolumn{1}{c|}{} &
  receding hairline &
  binary &
  0/1 &
  local &
  identity \\ \cline{3-7} 
 &
  \multicolumn{1}{c|}{} &
  sideburns &
  binary &
  0/1 &
  local &
  identity \\ \cline{2-7} 
 &
  \multicolumn{1}{c|}{\multirow{4}{*}{facial hair}} &
  5 o'clock shadow &
  binary &
  0/1 &
  local &
  identity \\ \cline{3-7} 
 &
  \multicolumn{1}{c|}{} &
  goatee &
  binary &
  0/1 &
  local &
  identity \\ \cline{3-7} 
 &
  \multicolumn{1}{c|}{} &
  mustache &
  binary &
  0/1 &
  local &
  identity \\ \cline{3-7} 
 &
  \multicolumn{1}{c|}{} &
  no beard &
  binary &
  0/1 &
  local &
  identity \\ \cline{2-7} 
 &
  \multicolumn{1}{c|}{\multirow{3}{*}{eye}} &
  narrow eye &
  binary &
  0/1 &
  local &
  identity \\ \cline{3-7} 
 &
  \multicolumn{1}{c|}{} &
  bags under eyes &
  binary &
  0/1 &
  local &
  identity \\ \cline{3-7} 
 &
  \multicolumn{1}{c|}{} &
  eye color &
  binary &
  brown &
  local &
  identity \\ \cline{2-7} 
 &
  \multicolumn{1}{c|}{eyebrow} &
  eyebrow type &
  categorical &
  bushy, arched &
  local &
  identity \\ \cline{2-7} 
 &
  \multicolumn{1}{c|}{nose} &
  nose shape &
  categorical &
  pointy, big &
  local &
  identity \\ \cline{2-7} 
 &
  \multicolumn{1}{c|}{\multirow{2}{*}{mouth}} &
  big lips &
  binary &
  0/1 &
  local &
  identity \\ \cline{3-7} 
 &
  \multicolumn{1}{c|}{} &
  string mouth line &
  binary &
  0/1 &
  local &
  identity \\ \cline{2-7} 
 &
  \multicolumn{1}{c|}{forehead} &
  forehead type &
  categorical &
  fully-visible partially-visible, obstructed &
  local &
  identity \\ \cline{2-7} 
 &
  \multicolumn{1}{c|}{\multirow{2}{*}{cheeks}} &
  rosy cheeks &
  binary &
  0/1 &
  local &
  identity \\ \cline{3-7} 
 &
  \multicolumn{1}{c|}{} &
  high cheekbones &
  binary &
  0/1 &
  local &
  identity \\ \cline{2-7} 
 &
  \multicolumn{1}{c|}{jaw} &
  round jaw &
  binary &
  0/1 &
  local &
  identity \\ \cline{2-7} 
 &
  \multicolumn{1}{c|}{chin} &
  double chin &
  binary &
  0/1 &
  local &
  identity \\ \hline
\multirow{8}{*}{Global Attributes} &
  \multicolumn{2}{c|}{face shape} &
  categorical &
  oval, square, round, flushed &
  global &
  identity \\ \cline{2-7} 
 &
  \multicolumn{2}{c|}{skin} &
  categorical &
  pale, shiny &
  global &
  identity/environment \\ \cline{2-7} 
 &
  \multicolumn{2}{c|}{attractive} &
  binary &
  0/1 &
  global &
  identity \\ \cline{2-7} 
 &
  \multicolumn{2}{c|}{age} &
  numerical &
   &
  global &
  identity \\ \cline{2-7} 
 &
  \multicolumn{2}{c|}{gender} &
  binary &
  0/1 &
  global &
  identity \\ \cline{2-7} 
 &
  \multicolumn{2}{c|}{race} &
  categorical &
  asian, white, black &
  global &
  identity \\ \cline{2-7} 
 &
  \multicolumn{2}{c|}{ethnicity} &
  categorical &
  \begin{tabular}[c]{@{}c@{}}African American, East Asian, \\ Caucasian Latin, Asian Indian\end{tabular} &
  global &
  identity \\ \cline{2-7} 
 &
  \multicolumn{2}{c|}{chubby} &
  binary &
  0/1 &
  global &
  identity \\ \hline
\multirow{2}{*}{Makeup} &
  \multicolumn{2}{c|}{makeup} &
  binary &
  0/1 &
  global &
  accessory \\ \cline{2-7} 
 &
  \multicolumn{2}{c|}{wearing lipstick} &
  binary &
  0/1 &
  local &
  environment \\ \hline
\multirow{3}{*}{Expression} &
  \multicolumn{2}{c|}{expression} &
  categorical &
  \begin{tabular}[c]{@{}c@{}}neutral, happy, sad, fearful, angry, awed \\ surprised, disgusted,  happily surprised, \\ happily disgusted, sadly fearful, hatred, \\ sadly angry, sadly surprised, sadly disgusted,\\  fearfully angry, fearfully surprised,\\  fearfully disgusted, angrily surprised, \\ angrily disgusted, disgustedly surprised, appalled\end{tabular} &
  global &
  expression \\ \cline{2-7} 
 &
  \multicolumn{2}{c|}{smiling} &
  binary &
  0/1 &
  global &
  expression \\ \cline{2-7} 
 &
  \multicolumn{2}{c|}{frowning} &
  binary &
  0/1 &
  global &
  expression \\ \hline
\multirow{3}{*}{Pose} &
  \multicolumn{2}{c|}{head pose} &
  numerical &
  yaw, pitch, roll &
  global &
  pose \\ \cline{2-7} 
 &
  \multicolumn{2}{c|}{mouth open} &
  binary &
  0/1 &
  local &
  pose \\ \cline{2-7} 
 &
  \multicolumn{2}{c|}{eye open} &
  binary &
  0/1 &
  local &
  pose \\ \hline
\multirow{7}{*}{Accessories} &
  \multicolumn{2}{c|}{no eyewear} &
  binary &
  0/1 &
  local &
  environment \\ \cline{2-7} 
 &
  \multicolumn{2}{c|}{eyeglasses} &
  binary &
  0/1 &
  local &
  environment \\ \cline{2-7} 
 &
  \multicolumn{2}{c|}{sunglasses} &
  binary &
  0/1 &
  local &
  environment \\ \cline{2-7} 
 &
  \multicolumn{2}{c|}{wearing earrings} &
  binary &
  0/1 &
  local &
  environment \\ \cline{2-7} 
 &
  \multicolumn{2}{c|}{wearing hat} &
  binary &
  0/1 &
  local &
  environment \\ \cline{2-7} 
 &
  \multicolumn{2}{c|}{wearing necktie} &
  binary &
  0/1 &
  local &
  environment \\ \cline{2-7} 
 &
  \multicolumn{2}{c|}{wearing necklace} &
  binary &
  0/1 &
  local &
  environment \\ \hline
\multicolumn{1}{|l|}{\multirow{4}{*}{Image Attributes}} &
  \multicolumn{2}{c|}{illumination} &
  numerical &
   &
  global &
  environment \\ \cline{2-7} 
\multicolumn{1}{|l|}{} &
  \multicolumn{2}{c|}{occlusion} &
  numerical &
   &
  global &
  environment \\ \cline{2-7} 
\multicolumn{1}{|l|}{} &
  \multicolumn{2}{c|}{background} &
  numerical &
   &
  global &
  environment \\ \cline{2-7} 
\multicolumn{1}{|l|}{} &
  \multicolumn{2}{c|}{blurry} &
  numerical &
   &
  global &
  environment \\ \hline
\end{tabular}%
}
\end{table*}
\subsection{Facial Attributes}

The input of a facial attribute editing model is a 2D image of a human's face. But, what do we mean by the attributes of a face that can be edited by these models? Table \ref{tab:attributes} summarizes all facial attributes that have been  considered in the related datasets and manipulation models. Here, we have divided these attributes into seven groups. It is clear that a human's face is exactly defined by the shape (3D structure) and appearance (color and texture) of its different parts. These are the main attributes of a face that determine the identity of a person and are represented with the two first attribute categories of Table \ref{tab:attributes}, i.e. the local part-specific or global facial attributes. For example, the eye color is related to a specific part of the face while the gender is affected by different characteristics of the face from its different parts and regions. It is assumed that changing the value of any of these attributes alters the identity of the corresponding face. What we see in a face photo is not merely determined by the intrinsic characteristics of the face but there are other factors affecting the appearance of the face and the other facial attributes of Table \ref{tab:attributes} are defined to describe these factors.

The expression and pose of a human face, also known as the face style, are two sets of attributes characterizing the status of the face in front of the camera and are easily controlled by the movement of muscles. Expressions reveal the inner feelings of a person and are usually described with a categorical attribute that shows different human expressions detailed in Table \ref{tab:attributes} (e.g. happy, sad, angry). In some works, special attributes are devoted to more important expressions which are not different from the overall expression attribute. The pose attributes show the situation of the head in terms of its yaw, pitch and roll, i.e. rotations around the neck, ears and nose axes, respectively. The open/close situation of the mouth and eyes are also considered as two attributes of the face than can be categorized as part of the pose attributes. 

In addition to the above factors, there are some external objects and cosmetic materials that are added to faces and change their appearance.  Makeup and accessories are two prominent examples of such factors that are mentioned under the categories with the same names in Table \ref{tab:attributes}. Makeup attributes are more studied in the face editing models developed for the beauty industries and among the accessory attributes, glasses have received more attention. 

Finally, the environment and the camera used to capture a photo may affect the final face image. The background image, occlusion, illumination, and blurriness status of the image are some attributes related to these aspects of a face image.
 
In addition to the above categorization, some related works organize the facial attributes in different ways. FLAME which is a3D head model, formulates a face three sets of parameters, namely facial shape parameters \(\beta\), pose parameters \(\theta\), and facial expression parameters \(\psi\) as \(\Theta=(\beta,\theta,\psi)\). Also based on the possible values that an attribute can take, the facial attributes are categorized into three major categories: binary, categorical, and numerical attributes. Binary attributes (e.g. mustache, goatee, and high cheekbone) can accept only two values representing the presence and absence of the corresponding attribute in an image. Categorical attributes are those whose values belong to a set of specific values. For example, forehead type is a categorical attribute and only accepts one of the values of the set \{fully-visible, partially-visible, or Obstructed\}. The last category is the numerical attributes whose values are real positive numbers, like age which can accept any number greater than zero. Numerical attributes are usually divided into successive bins and treated as categorical attributes. 

In another categorization of facial attributes, these attributes are divided into local and global classes. The local attributes (e.g. rosy cheek) focus on specific parts or bounded sub-region of the face while global attributes (e.g. face shape and attractiveness) are related to the total face or a major part of it. 

Due to the different nature of the hair-related attributes, some studies (like \cite{tan2020michigan}) separate these sort of attributes (e.g. Hair Color, Hair Type, Hair Structure, Hair length, and Bang) from face-related attributes. Contrary to almost all other parts of the face, variable geometry and texture of the hair makes it difficult to edit the attributes related to this part.

In this study, we survey papers on semantic facial attribute editing. Some of them are special-purpose models focusing on a single attribute while the others are capable of editing multiple attributes either separately or simultaneously. For example, \cite{he2019s2gan, zhu2020look,fang2020triple,yao2020high} are special-purpose models developed for manipulating the age of a face image. There are other single-attribute models for editing makeup \cite{gu2019ladn}, hair \cite{tan2020michigan}, and expression\cite{Wang2019dft}. 
As examples of models that can simultaneously edit more than one attribute, \cite{tewari2020stylerig} can edit pose, illumination, and expression of the input image, \cite{wu2019relgan} manipulates hair color, gender, and age attributes, and \cite{yin2019instance} edits eyeglass, mustache, and goatee. 
There are many other works are that are able to edit almost all of the facial attributes as in \cite{chen2020harnessing,xiao2018elegant,tao2019resattr,choi2020stargan}.

\subsection{Definitions}
In this section, we provide exact definitions for tasks and concepts related to the scope of this paper, i.e. semantic facial attribute editing, in three subsections: face image generation, face image manipulation, and facial attribute editing. Face generation is the process of creating a new face image from a random noise. The distribution of the generated images is learned from a set of training face images. Although these models are not primarily developed for face editing tasks, they have formed the base of some facial attribute editing models. Face image manipulation refers to any process that is applied to an input face image to generate a modified version of that image. As we will see in the following, face image inpainting and colorizing are two tasks of this kind. In fact, semantic facial attribute editing is a special form of face image manipulation but as the focus of this survey is on this task, we review its related concepts and sub-tasks in a separate subsection.

\subsubsection{Face image generation}
As stated before, face editing is not among the primary design goals of face image generators. However, they have been widely used as the main building blocks of many semantic facial attribute editing models. In the following, some tasks and topics related to face image generators are defined.

\textbf{General face image generation} 
A generator is a model that can produce samples from a specific distribution. Generating random vectors from standard probability distributions such as normal or uniform distributions is much simpler than generating more complex data such as face images as there are more dependencies between different elements of an image. Generative Adversarial Networks (GANs), designed by Ian Goodfellow and his colleagues \cite{goodfellow2014generative}, are a powerful tool for random image data generation. Trained on a set of face images, a GAN can generate similar face photos based on input noise vectors. Since the introduction of GANs, many successful models have been proposed for face image generation. PgGAN \cite{karras2017progressive} is an extension of the GAN training that allows for more stable training of generator models capable of producing large, high-quality images. It begins with a very low-resolution image and progressively adds layers to increase the generator model's output size and the discriminator model's input size until the desired image size is reached. The Style Generative Adversarial Network (StyleGAN) as another face generator model \cite{karras2019style}, is an improvement to the pgGAN architecture that allows for control over the created images' disentangled style properties by introducing an intermediate latent space. StyleGAN generates high-resolution and realistic images, but sometimes unnatural spots (artifacts) appear on the resulted images. In StyleGAN2 \cite{karras2020analyzing}, these artifacts are exposed and analyzed. StyleGAN2 is now the most successful model in face generation.

Some semantic facial attribute editing models are based on pre-trained face generator models. For example, Tewari et al. \cite{tewari2020stylerig} use StyleGAN as the base of their proposed method to transfer face pose, expression and scene illumination from a source face image to a target face image. Shen et al. \cite{shen2020interfacegan} also employ StyeGAN's and pgGAN's latent spaces to edit facial attributes. TunaGAN \cite{mao2019tunagan} is another example of the aforementioned models that use an auxiliary network on top of StyleGAN to gain control over the latent space of StyleGAN and perform requested editing.

\textbf{Attribute-based face image generation} 
In the basic form of GANs introduced in the previous section, generators synthesize images without any specific condition and there is no control over the generated images. In conditional GANs, the generator is conditioned on one or more specific features, e.g. facial attributes, as input and the generated image has all of the determined attributes. The other attributes not mentioned in the condition are produced randomly and form the diversity in the generated results. These models are more appropriate for doing semantic facial attribute editing in which the conditions are used for controlling (or editing) the attributes of the generated images \cite{zhang2019adversarially,yan2016attribute2image}.

\textbf{Feature disentanglement} 
Unfortunately, the facial attributes are entangled with each other in the latent or intermediate latent space of the current generators. As a result, manipulating the value of an attribute in the latent space of these models may change the values of other irrelevant attributes as well. For example, makeup is highly correlated with gender and by adding makeup, gender is also affected \cite{mao2019tunagan}. 
In addition to semantic connection between the attributes, their representation in the generator's latent space is a cause of this phenomenon known as feature entanglement \cite{ning2020fegan,ning2021continuous}. Recently, many researches, e.g. \cite{nitzan2020disentangling,zhu2020learning}, have been trying to provide image representation spaces with separate and independent attributes so that the attribute of interest can be modified without affecting the other attributes. If this feature disentanglement is completely achieved in the latent space of a face generator, the facial attribute editing task will be easily and directly mapped to simple vector manipulation operations in this latent space. 

\textbf{Latent space embedding} 
As stated before, generative models can create face images from input random noises without providing any control over the content of the generated images. The process of locating the best and the closest representation point in the latent space of a generator model for a given face image is called latent space embedding. By having this compact representation, one can recover the initial face image by feeding the generative model with the obtained latent vector. Embedding an image into the generator's latent space and manipulating the resultant latent representation can lead to the editing of the intended attribute on the output image. Also known as GAN inversion methods, latent space embedding methods fall into two main groups \cite{zhu2020domain}: learning-based \cite{richardson2020encoding} and optimization-based \cite{abdal2020image2stylegan++} methods. The former learns a deterministic mapping from the image space back into the latent space based on the pairs of an image and its corresponding latent code while the latter performs an image-wise optimization to find the latent code with the minimum reconstruction loss.

\textbf{Latent space interpolation} 
The latent space of a generative model is a multi-dimensional continuous space and each point in this space represents an image in the image space. It is easy to explore this vector space by applying different vector arithmetic operations such as addition, subtraction, and multiplication on scalars. Latent space interpolation refers to the task of creating a series of points for traversing from a start to an end point in the latent space. The images corresponding to these points are expected to resemble gradual changes from an initial to a final image in the image space. Latent space interpolation may be used in different image manipulation tasks such as image morphing \cite{abdal2019image2stylegan} and facial attribute editing \cite{he2019attgan}. In facial image editing, latent space interpolation may be used to generalize the binary attributes to continuous attribute values -known as attribute intensity control. For example, the mouth of a person can be edited to smoothly change from the completely close to completely open state \cite{he2019attgan}.

\subsubsection{Face image manipulation}
This section explains some face image manipulation tasks in which an input face image is modified in a way that the result is still a face image but with different characteristics. As mentioned earlier, the facial attribute editing tasks are covered in the following subsections and here other applications such as face image inpainting, morphing, translation, swap and de-identification are discussed.  

\textbf{Face inpainting}
A face inpainting algorithm aims to reconstruct the missing parts of an input face image based on the existing parts. The most important point in such algorithms is homogeneousness and coherence between reconstructed parts and available face content from every perspective like color, texture, etc. Face completion is much more complicated from general image inpainting task, because face portions are highly structured and variant at the same time. Some methods use facial geometry information in a supervised reconstruction procedure \cite{song2019geometry} to preserve appropriate structure. Another solution is to consider another face image as reference for generating missing parts \cite{deng2020reference}. 

\textbf{Face to face translation}
Face-to-face translation task is a specific instance of the more general image to image translation which concentrates on the face images. Face-to-face translation maps an input face image from source face domain \(A\) to the target face domain \(B\). Based on source and target domains, face-to-face translation can be employed in different applications. 
Sketch-to-photo conversion is a common example of these applications where the input domain is the black and white sketches from faces and the goal is to convert the input sketch to a  colorful realistic face photo. Sketch to photo translation can be used in criminal tracking and mobile-app entertainment \cite{hu2020facial,richardson2021encoding}. Another form of face-to-face translation is image colorizing which takes a grayscale  image as input and converts it into a colorized image that reflects the input's semantic colors and tones \cite{guan2020collaborative,ramya2019face}.
Some researches treat the facial attribute editing  as a face-to-face translation task in which the input face image is translated into another face image with the same identity but different in some attribute like expression, pose, or age \cite{cao2019makeup,he2019deliberation,choi2018stargan}.

\textbf{Face morphing}
The aim of face morphing, as a subfield of image morphing, is to create a new face image that resembles the biometric information of two or more face images\cite{scherhag2019face}. The morphed face images that are weighted combinations of two subjects may be verified against probes of both relevant subjects by various face recognition systems. This technique is considered a critical threat for face recognition systems as if a morphed face image is added to the database of a face recognition system, both persons will be verified against this manipulated face photo \cite{damer2018morgan}. However, face morphing is more known for its application in the entertainment industry. Face morphing is not usually treated as a face attribute editing task and is not covered in this paper.

\textbf{Face swapping}
In face swapping, the face region in an image is completely replaced  with a different face while the other components of the image remain untouched \cite{nirkin2019fsgan}. Face swapping is a well-known form of creating deepfake videos, and various models have been proposed for this task (See \cite{ngo2020unified,yang2021deep,petrov2020deepfacelab,li2019faceshifter} for some examples). Since face swapping  modifies the whole face not some attributes, it is not considered as a face editing task in this survey.

\textbf{De-identification}
De-identification aims to manipulate the identity of a person in an image to make it unrecognizable for human users or automatic face recognition systems. The most straightforward solution is the use of face-swapping methods \cite{sun2018natural}. Besides, face editing methods can also be utilized to manipulate a face image to completely change the identity information \cite{li2019anonymousnet}. 

\subsubsection{Facial attribute editing}
This section explains  some concepts, applications, and challenges  directly connected with the facial attribute editing problem. 

\textbf{Facial attribute editing/manipulation}
Facial attribute editing or facial attribute manipulation task (both terms are used interchangeably through the paper) which is the main focus of this study, refers to a process that takes a face image as input and generates a corresponding output face image with specific attribute values, e.g. converting  a neutral face to a smiling one or removing eyeglasses from a face photo. As mentioned earlier, editing an attribute can be done in different levels (spaces) including pixel space, 2D or 3D geometrical representation, low-level feature vector, or high-level semantic descriptors.

\textbf{Face reenactment}
In face reenactment an attribute, usually expression or pose, is conveyed from a driving face image to a source image\cite{tripathy2022single}. This task is also known as style transfer which transfers styles from a face image to another one. If a style transfer or face reenactment method is applied to consecutive frames of a video, a \textit{deepfake} video will be generated. Deepfake which has become very popular nowadays, refers to generating fake face images using deep learning models (usually face reenactment methods) to apply desired poses and expressions from the frames of a video to an image of the target person. Deepfake is a particular application of facial attribute editing tasks and  will be explained in the Section \ref{sec5} in more detail.

\textbf{Attribute interpolation}
In semantic facial editing, a face image is edited by changing an attribute from an initial value to a target value. In some cases, intermediate states of an attribute can also result in valid face images\cite{liu2021smoothing}. Genrating such intermediate images is known as attribute interpolation and is done in a similar way as latent space interpolation explained earlier. Age and skin color are two examples of such attributes.

\subsection{Related Works}

To show the necessity of current survey despite the publication of other surveys in the field of facial attribute editing, in this section, the more related studies are reviewed. 
Wang et al. \cite{wang2020survey} has studied existing face data augmentation works with an emphasis on the generative adversarial networks as one of the most powerful and effective tools in this field. However, the focus of this study is neither on the facial attribute editing nor on the GAN-based models.
Akhtar and his colleagues \cite{akhtar2019face} have studied the face manipulation works from three different perspectives of generation, detection and recognition.  They have briefly reviewed a small number of face manipulation methods. 
\cite{wang2020state} has reviewed the recent researches on GANs including image semantic editing, super-resolution, inpainting, carton generation, etc., with a special focus on image synthesizing approaches. Though a lot of the examples of the paper belong to the face domain, the study is not devoted to face-related researches, and the major parts of the work are dedicated to non-facial image synthesis tasks. 
\cite{liu2021generative} is another similar study that has reviewed some algorithms and applications of GAN-based image/video synthesizing networks including face images. 

As mentioned earlier, GAN-based models have been widely used in generating deepfakes. GAN-based models can create fake images or videos by manipulating the facial information including identity and expression in images or videos. So, prominent studies have been done on GAN-based models for deepfake generation and detection. \cite{juefei2021countering,tolosana2020deepfakes} have covered the recent techniques for manipulating face images. These studies have divided deepfake approaches into four groups including entire face synthesis, identity swap, attribute manipulation, and expression swap. \cite{mirsky2020creation} has surveyed plenty of models used for deepfake creation. \cite{mirsky2020creation} focuses on the newest works on deepfake field and describes how the related architectures work. \cite{zhang2020deep} is another related survey that divides face image synthesis methods into three categories: face-reenactment, face-swap, and face-generation. The main focus of all these studies is on the current trends and advancements in the deepfake domain not facial attribute editing.

\cite{zheng2020survey,chen2020overview} are the two most related studies to ours.
\cite{zheng2020survey} surveys facial attribute manipulation in addition to facial attribute estimation as two sub-tasks of facial attribute analysis tasks. Facial attribute estimation recognizes the presence of an attribute in a face image. Facial attribute manipulation methods are categorized into two subcategories: model-based methods which translate images from a source domain to a target domain, and extra-condition methods that take conditions in the format of another image or attribute vector.
\cite{chen2020overview} focuses on generating attributes that do not exist on the face image. In this work, only a few prominent GAN-based methods, including AttGAN, SGGAN, STGAN, StarGAN, and SC-FEGAN have been discussed. 

None of the above surveys has dedicatedly concentrated on the facial attribute editing methods to provide an in-depth review of the related works and cover all of the state-of-the-art models in this field. In this paper, we have tried to provide a comprehensive and up-to-date study of the facial attribute editing methods based on generative adversarial networks from different perspectives including architecture, loss function, evaluation method, and application – most of them not covered by the previous reviews. 

\section{Approaches} \label{sec3}

During the past few years, generative adversarial networks have experienced a very fast growth leading to an explosion of image generation and manipulation methods. Several facial attribute editing methods have been proposed so far. This section reviews these models in terms of the architectures and the loss functions of the models in two following subsections.

\subsection{Architectures}
\begin{figure*}[htb]
    \centering
    \includegraphics[width=\textwidth]{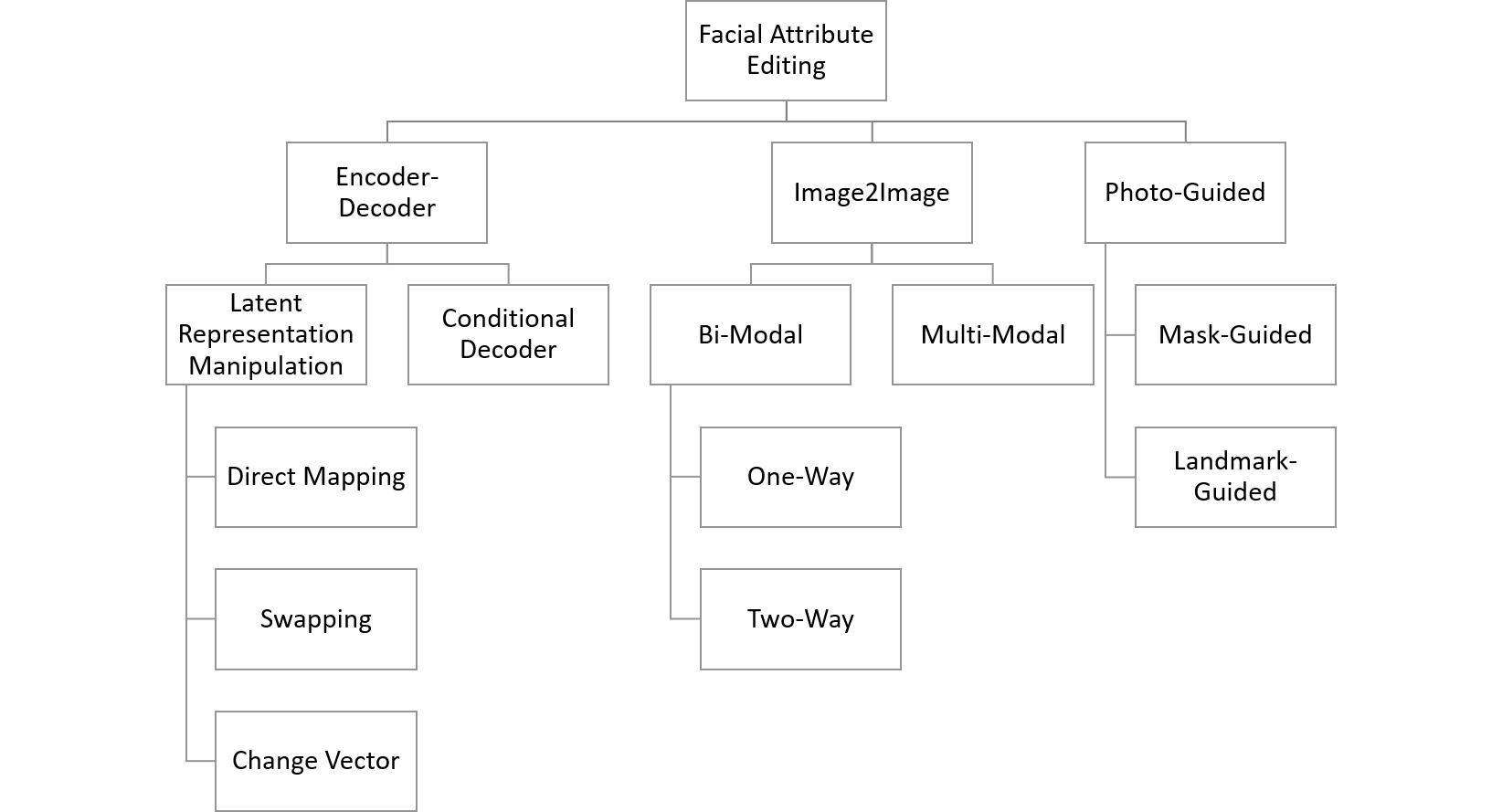}
    \caption{The categorization of semantic facial attribute editing models based on their architectures.}    \label{fig:arch:architectures}
\end{figure*}

Regardless of the attributes that are to be edited, we first categorize and study the proposed facial attribute editing models based on the architectures of these models.  As shown in Fig. \ref{fig:arch:architectures}, there main approaches have been adopted by researchers. In encoder-decoder models, the input image is mapped into a latent space and a latent space manipulation method or a conditional decoder is employed for applying the requested changes. The image-to-image translation methods treat face images with different values of an attribute as different modalities of images and propose networks for mapping images from one modality to the other.  Finally, in photo-guided models, extra information from the input image, e.g. the face landmarks, is used to guide the attribute editing process.  These three groups of methods along with the related models are described in the following. 

\subsubsection{Encoder-Decoder}
A common architecture for a facial attribute editing model is the encoder-decoder structure in which the encoder maps the input image to a latent space vector representing the features of the input face and the decoder reconstruct the a face image from each point of the latent space. The desired attribute modification can then be encoded as some vector manipulation operators or as a conditional decoder conditioned on the target attribute vector. 

\begin{figure}[htb]
    \centering
    \includegraphics[width=0.48\textwidth]{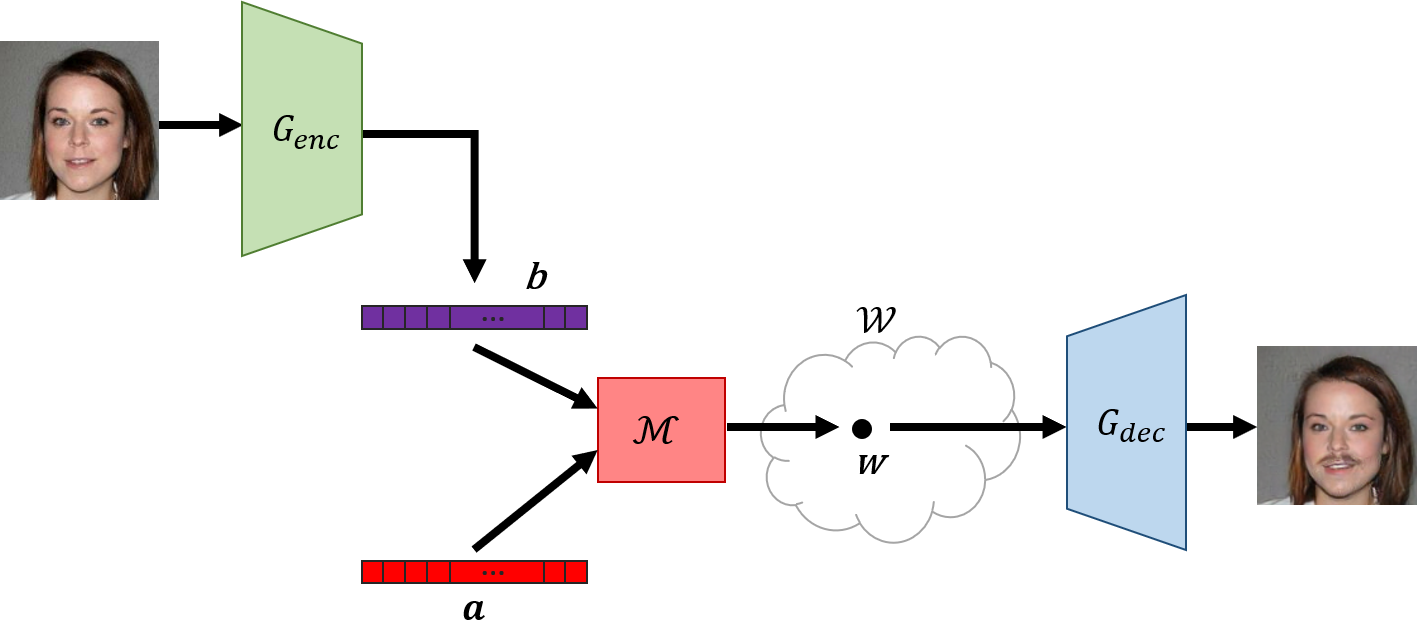}
    \caption{The overall architecture of Latent Representation Manipulation models.}    \label{fig:arch:lrmoverall}
\end{figure}

\textbf{Latent Representation Manipulation:}
All facial attribute editing models reviewed in this study use a generator to synthesize new edited face images. The generator network maps its input to a face image in the RGB space. The inputs of the generator are the points of the latent space. In this case, each point of the latent space corresponds to a unique face with specific attributes decoded by the generator. This correspondence between the latent points and face images has motivated facial attribute manipulation methods in which the input face images are mapped to a point in the latent space in which the edits are applied. The resulted point is then mapped to a face image with the desired attributes in the RGB space using the aforementioned generator. The typical pipeline of the models of this category is shown in Figure \ref{fig:arch:lrmoverall}. As shown in this figure, the input image $I$ is first encoded as vector $b$. The requested edits are given as the target attribute vector $a$. The mapping module $M$ gives the final point in the latent space which is converted to the edited image by $G_{dec}$. The input image can be directly encoded as a point of latent space ($w \in W$). In this case, some edits are simplified as a vector addition, $w = a + \Delta w$, where $\Delta w$ is determined based on the required changes. Moreover, when the requested changes are given as a driving image, the target attribute vector $a$ can be a point of the latent space as well.    
Prominent models of this category are studied in the following. 

\begin{figure}[htp]
    \centering
    \includegraphics[width=0.45\textwidth]{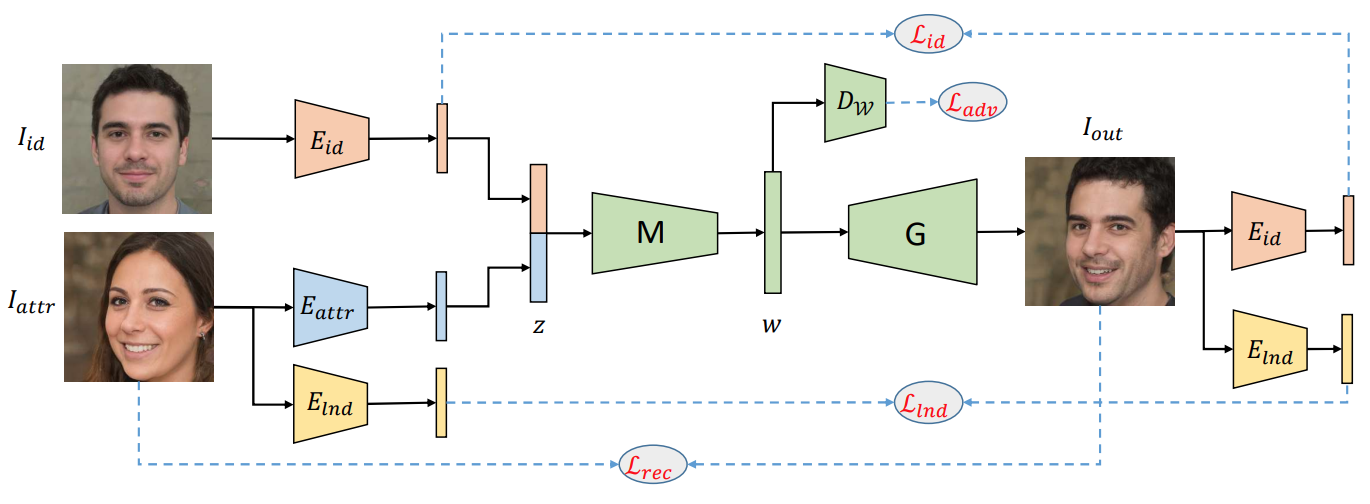}
    \caption{The architecture of the model proposed in \cite{nitzan2020disentangling}, this model combines two facial images by disentangling the identity and other facial attributes in the latent space.}
    \label{fig:arch:nitzan2020disentangling}
\end{figure}

Nitzan et al. \cite{nitzan2020disentangling} have proposed a facial attribute editing model which takes two images \(I_{id}\) and \(I_{attr}\) as input and produces a face image whose identity is taken from \(I_{id}\) and its all other attributes including pose, expression and illumination are coming from \(I_{attr}\). Figure \ref{fig:arch:nitzan2020disentangling} demonstrates the architecture of this model. As depicted in this figure, the model consists of two encoders \(E_{id}\) and \(E_{attr}\) for encoding identity and non-identity attributes of the input images, a mapping network \(M\) for converting the disentangled representations of these two sets of attributes to a point in the latent space of a generator,  and a pre-trained generator network \(G\) which produces the final edited face image. The other components of the model, i.e. \(E_{lnd}\) encoder and \(D_{W}\) discriminator, are only used in training the model. StyleGAN is used as the pre-trained generator for this model. This method had concentrated on disentangling identity from other facial attributes. Moreover, attribute editing is performed in the mapping stage and no other manipulation of the points in the latent space is required. The utilized losses for training the model include reconstruction, identity, landmark, and adversarial losses.

\begin{figure}[htb]
    \centering
    \includegraphics[width=0.48\textwidth]{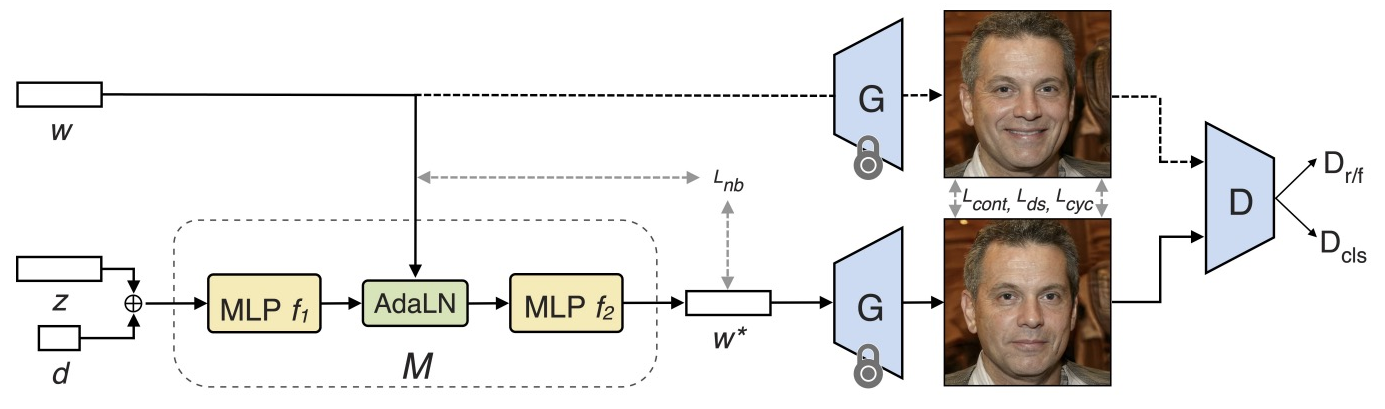}
    \caption{ISF-GAN's architecture \cite{liu2021isf}.}    
    \label{fig:arch:liu2021isf}
\end{figure}

ISF-GAN \cite{liu2021isf} exploits the general architecture of the latent space manipulation methods for image to image translation. As shown in Fig. \ref{fig:arch:liu2021isf}, the source image is encoded into the latent space of a pre-trained generator and then the modified representation ($w^*$) is generated by an Implicit Style Function ($M$) based on the target attributes ($d$) and random noise ($z$). The style transfer module consists of two MLP networks with the adaptive layer normalization applied to the output of the first network. Different loss functions are used to train the model.

\begin{figure}[htb]
    \centering
    \includegraphics[width=0.48\textwidth]{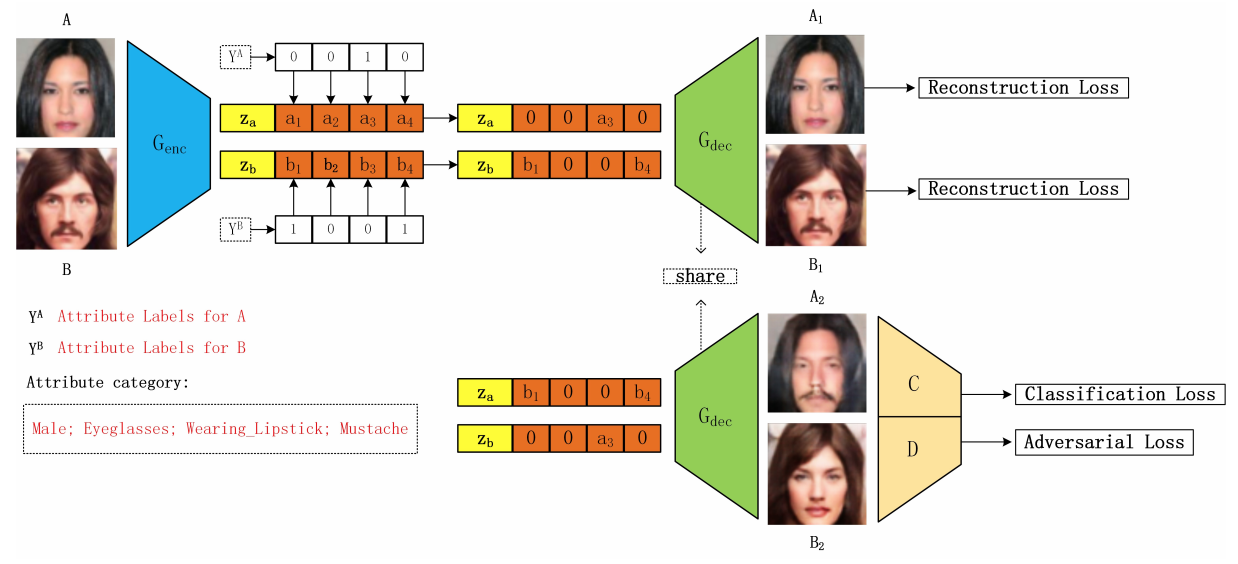}
    \caption{MulGAN's architecture \cite{guo2019mulgan}.}    \label{fig:arch:guo2019mulgan}
\end{figure}

MulGAN \cite{guo2019mulgan} (Figure \ref{fig:arch:guo2019mulgan}) is a generative model for editing special attributes of an input image based on an exemplar face image. As shown in Figure \ref{fig:arch:guo2019mulgan}, the model takes two images \(A\) and \(B\) along with two binary attribute tag vectors  \(Y^{A}\) and \(Y^{B}\) as input. Using the encoder module of the generator (\(G_{enc}\) ), the input images are encoded in two vectors each consisting of two attribute-relevant and attribute-irrelevant parts. The attribute-relevant section encodes the values of four attributes of the corresponding image (e.g. $(a_1,a_2,a_3,a_4)$ for Male, Eyeglasses, Wearing Lipstick, and Mustache, respectively) while the attribute-irrelevant section encompasses the other attributes of the input image. The decoder module of the generator (\(G^{dec}\)) produces face images using the same representation. Manipulation and replacement of the target attributes are then possible by the use of attribute tag vectors. The output of the encoder may be directly fed into the decoder to reconstruct the input image or the attributes can be exchanged using the attribute tag vectors to create edited versions of the input images. Reconstruction, classification, and adversarial losses are used for training the model.

\begin{figure}[htb]
    \centering
    \includegraphics[width=0.48\textwidth]{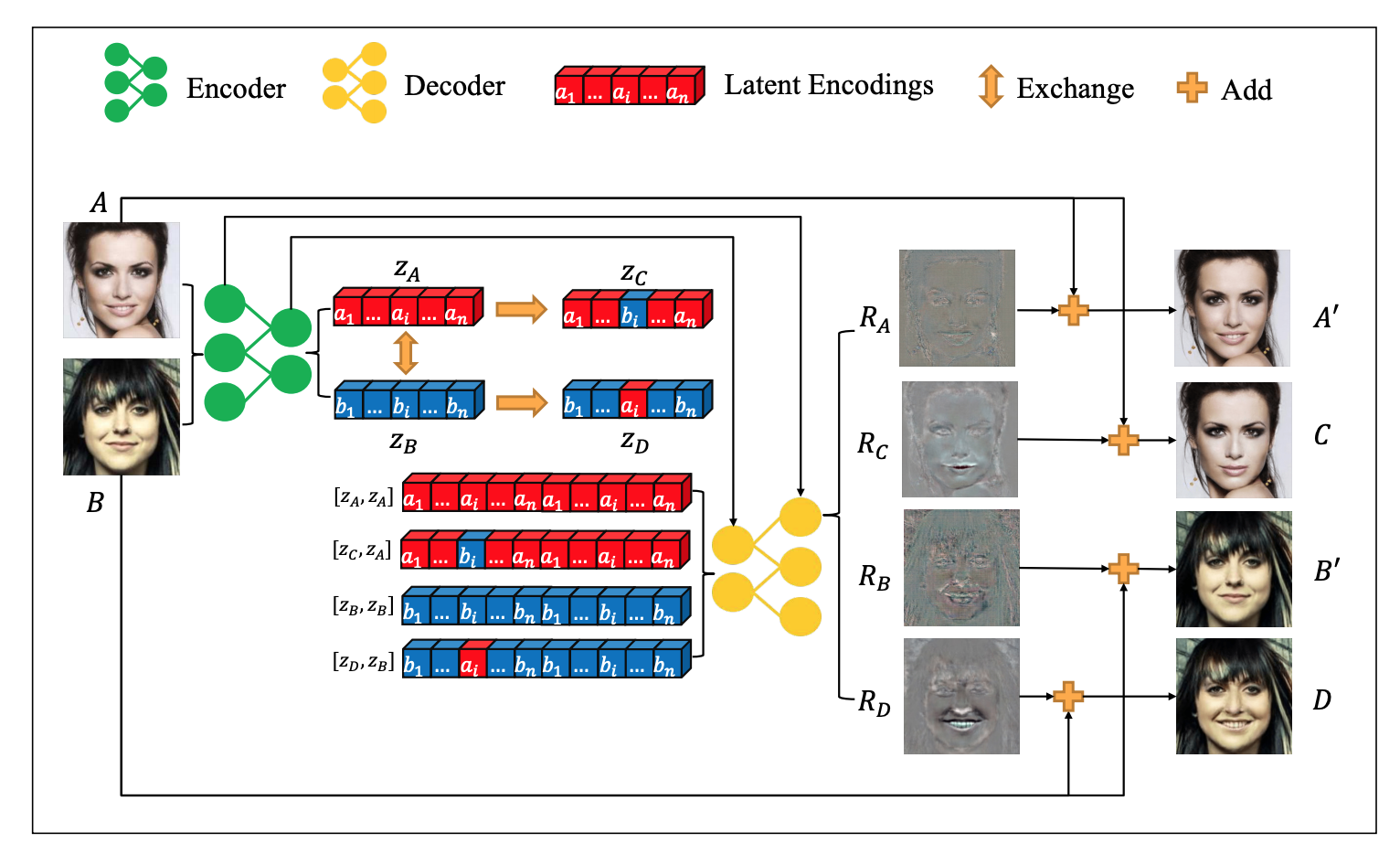}
    \caption{Elegant's architecture \cite{xiao2018elegant}.}
    \label{fig:arch:xiao2018elegant}
\end{figure}

Xiao et al. \cite{xiao2018elegant} proposed Elegant, a method for Exchanging Latent Encodings with GAN for transferring multiple facial attributes. This model takes two images as input and can transfer attributes from one image to another by exchanging certain parts of latent encodings of the two images. The model follows the encoder-decoder architecture. The encoder network learns a disentangled representation which guarantees that each attribute is encoded into a different part of the embedded vector. To do this, Elegant receives two images from two sets at each iteration of training: a positive set and a negative set. The image \(A\) from the former set has the attribute, while the image \(B\) from the negative set does not. The encoder is then utilized to get the latent encodings of images \(A\) and \(B\), denoted by \(z_A\) and \(z_B\), respectively. Then, the $i$th parts of the latent encodings of the input images are exchanged to obtain new encodings \(z_C\) and \(z_D\). It is expected that \(z_C\) and \(z_D\) are the encodings of the version of image A without the supposed attribute and the version of image B with the supposed attribute, respectively. After that, a decoder is trained that receives two different representations of a person and generates the residual image required for applying the same editing in the RGB space. By adding residuals to the corresponding input images, the output images are constructed with/without the target attribute. The whole procedure of this model is shown in Figure \ref{fig:arch:xiao2018elegant}. Elegant is based on a U-Net structure with a symmetrical encoder and decoder and shortcut connections between their intermediate layers. The model requires labeled data for determining positive and negative samples for each attribute at the training step. Also, discriminators are used for adversarial training and multi-scale discriminators can be utilized to improve the resolution of the generated images. The loss functions of this model include adversarial and reconstruction losses.

\begin{figure}[htb]
    \centering
    \includegraphics[width=0.48\textwidth]{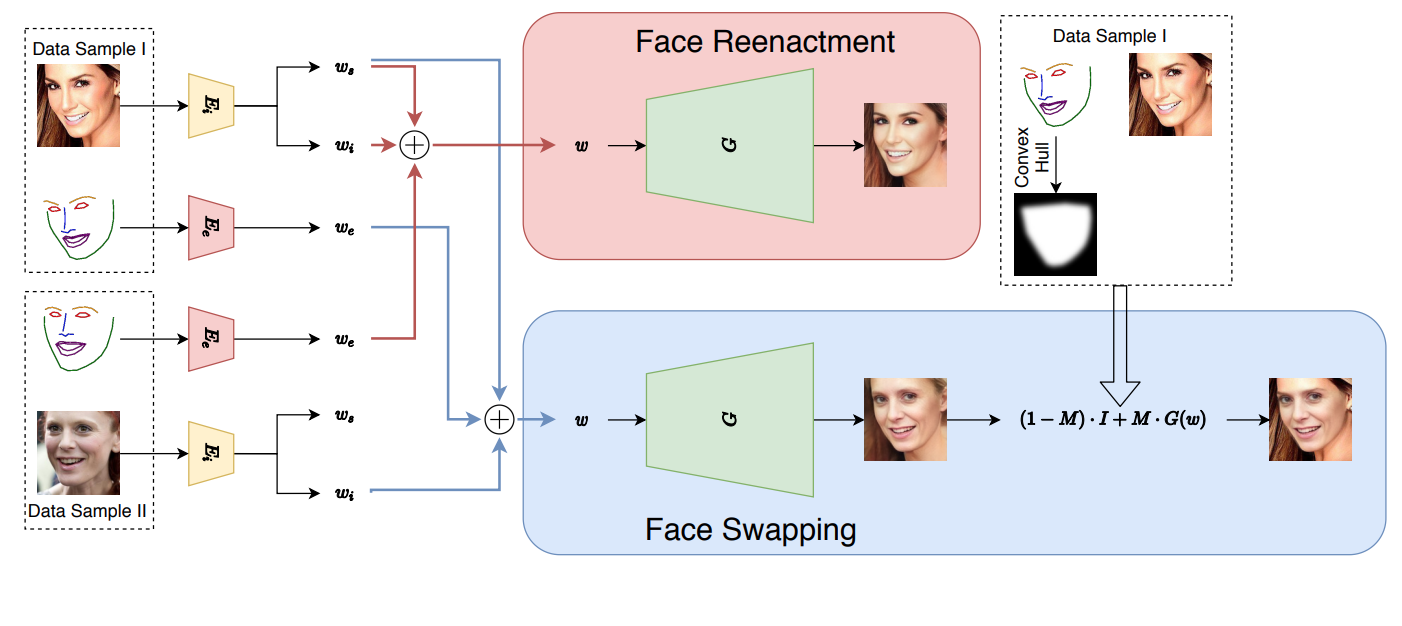}
    \caption{Architecture of the proposed model in \cite{ngo2020unified} }
    \label{fig:arch:ngo2020unified}
\end{figure}

\cite{ngo2020unified} proposed a unified end-to-end pipeline for face swapping and reenactment based on disentangled representation learning of specific visual attributes. This model employs a pre-trained generator (StyleGAN2) for generating face images and constructs the latent representations used by this generator by linear addition of four components: the mean of the generator's latent space \(\mu_G\), identity attributes \(w_i\), style attributes \(w_s\), and pose/facial expression attributes \(w_pe\). The three last components are extracted from given face images and their corresponding landmarks. For the face swapping task, \(w_i\) is taken from the source, and other attributes are taken from the target. For the face reenactment task, \(w_pe\) and \(w_s\) are taken from the source and the identity \(w_i\) from the target. This model's architecture is demonstrated in Figure \ref{fig:arch:ngo2020unified}. As shown in this figure, the source and the target faces with their facial landmarks are used as inputs to the two encoders \(E_i\) and \(E_pe\), respectively. The identity encoder \(E_i\) encodes the identity and style attributes and the pose/expression encoder extracts the pose/expression features. The used loss functions in this model include reconstruction, perceptual, landmark, and two identity losses.

\begin{figure}[htb]
    \centering
    \includegraphics[width=0.48\textwidth]{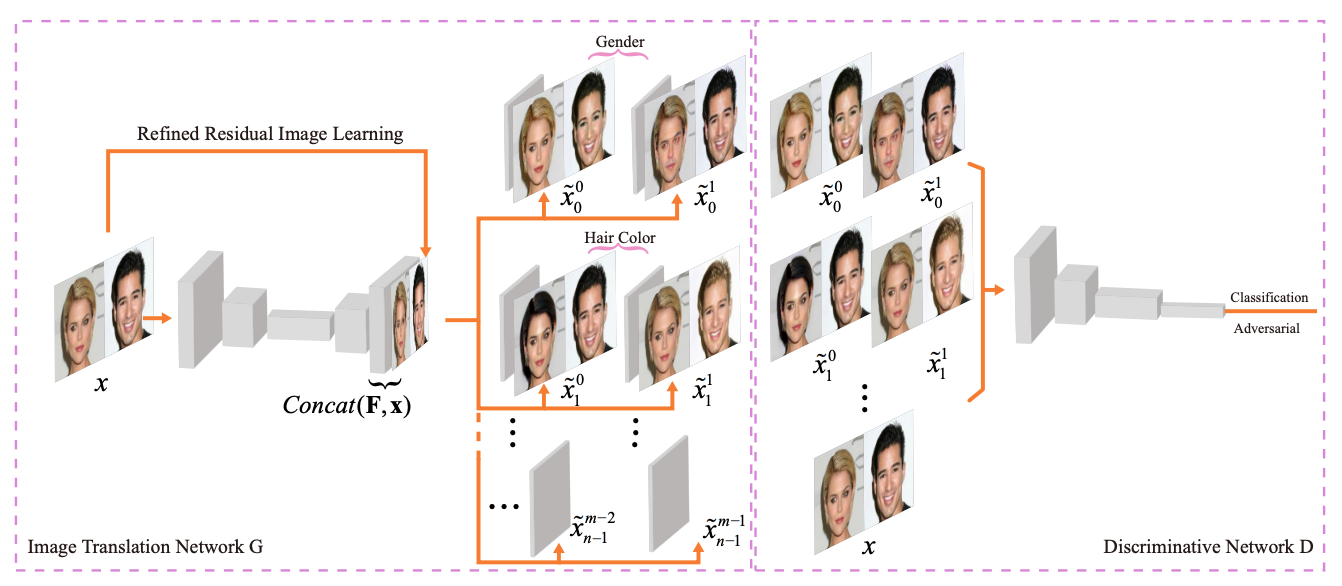}
    \caption{SG-GAN's architecture \cite{zhang2018sparsely}}
    \label{fig:arch:zhang2018sparsely}
\end{figure}

SG-GAN \cite{zhang2018sparsely} is a one-input multi-output network with encoder-decoder architecture.  The generator \(G\) of the model takes the input face image  \(x\) and it simultaneously generates all possible manipulations of the input image. Suppose \(x\) is a face image with \(n\) attributes  and each attribute has \(m\) possible values, the generator will produce $n\times m$ images where \({\widetilde{x}}^i_j\), \(0 \leq i < m\), \(0 \leq j < n\) is the edited version of the input image in which the $i$th attribute is set to the $j$th value as shown in Figure \ref{fig:arch:zhang2018sparsely}. The encoder and some layers of the decoder of the generator are shared among all attributes and there are separate networks for each attribute. SG-GAN is based on residual image learning in which the residual image $F$ generated by the network is concatenated with the input image $x$ and a convolutional layer is used to refine the result of the concatenation. All generated images are then fed into the discriminator for calculating the adversarial loss. In addition to adversarial loss, the reconstruction and classification losses are used for training the model.

\begin{figure}[htb]
    \centering
    \includegraphics[width=0.48\textwidth]{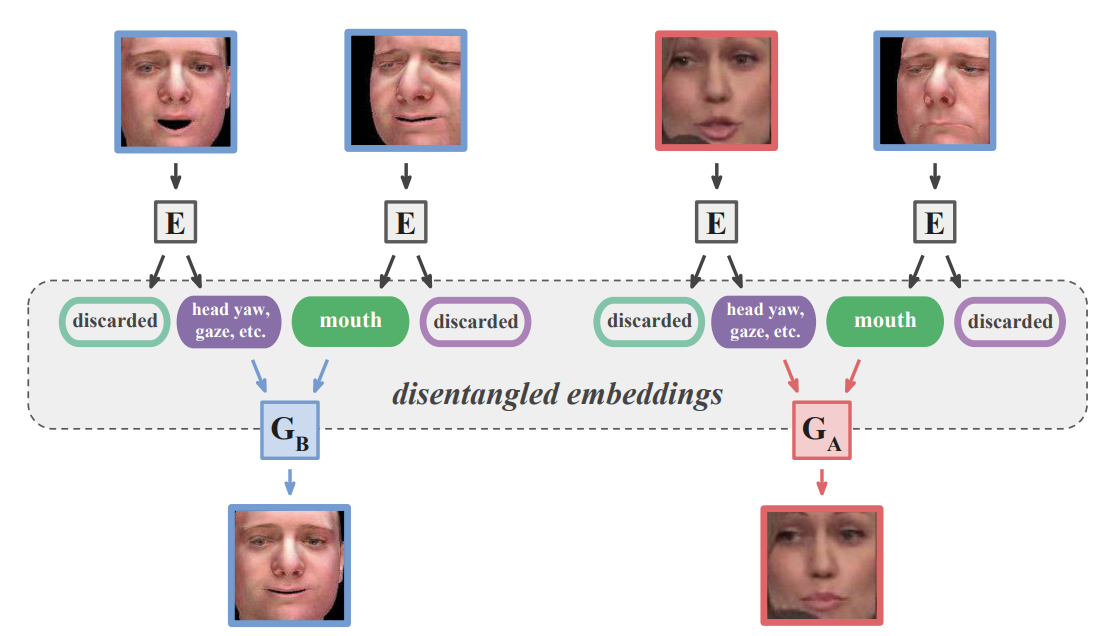}
    \caption{The architecture of PuppetGAN \cite{usman2019puppetgan}.}    \label{fig:arch:usman2019puppetgan}
\end{figure}

PuppetGAN \cite{usman2019puppetgan} is a cross-domain image manipulation model for transferring style between real and synthesized images. A domain-agnostic encoder (E) is utilized to map the images of both domains to a disentangled latent space in which the attribute of interest is isolated from the other facial attributes. Despite the common encoder, two different decoders one for the realistic domain ($G_A$) and the other for the synthetic domain ($G_B$) are employed in this model. Figure \ref{fig:arch:usman2019puppetgan} shows an example usage of the model for controlling the mouth state of realistic and synthesized target persons with synthesized driving images. The encoder and decoder networks are trained with reconstruction, disentanglement, cycle, and attribute cycle loss functions.

\begin{figure}[htb]
    \centering
    \includegraphics[width=0.48\textwidth]{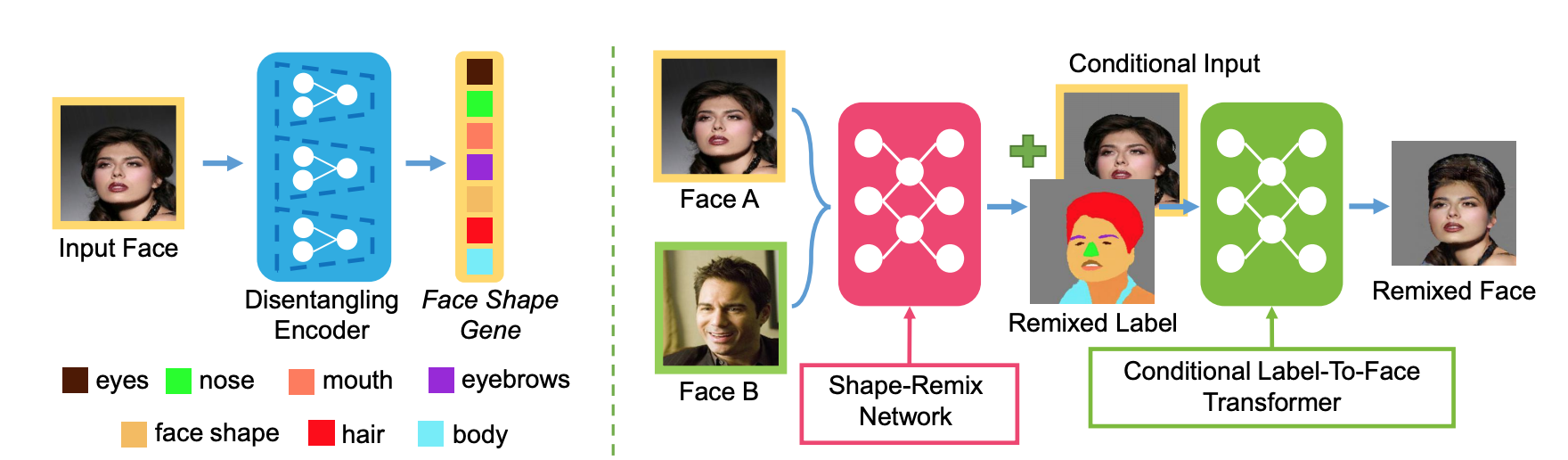}
    \caption{FaceShapeGene's architecture \cite{xu2019faceshapegene}.}    \label{fig:arch:xu2019faceshapegene}
\end{figure}

FaceShapeGene \cite{xu2019faceshapegene} provides another disentangled representation of the human face in which the shape of seven segments of a face (eyes, nose, mouth, eyebrows, face shape, hair, and body) are encoded in seven dimensions of a latent vector. Representation of each face part is learned separately using an encoder-decoder where each decoder learns to generate a label map for the corresponding part. FaceShapeGene representation is defined as the concatenation of these 1D part-wise features and another generator is trained to map these feature vectors to a whole-face label map. A shape-remix network is then developed to mix the latent representations of two faces and generate a remixed label. Finally, a conditional label to real face transformer is used to generate a face image with the shape of the remixed label and the identity of the conditional input face image (Figure \ref{fig:arch:xu2019faceshapegene}).

\begin{figure}[htb]
    \centering
    \includegraphics[width=0.48\textwidth]{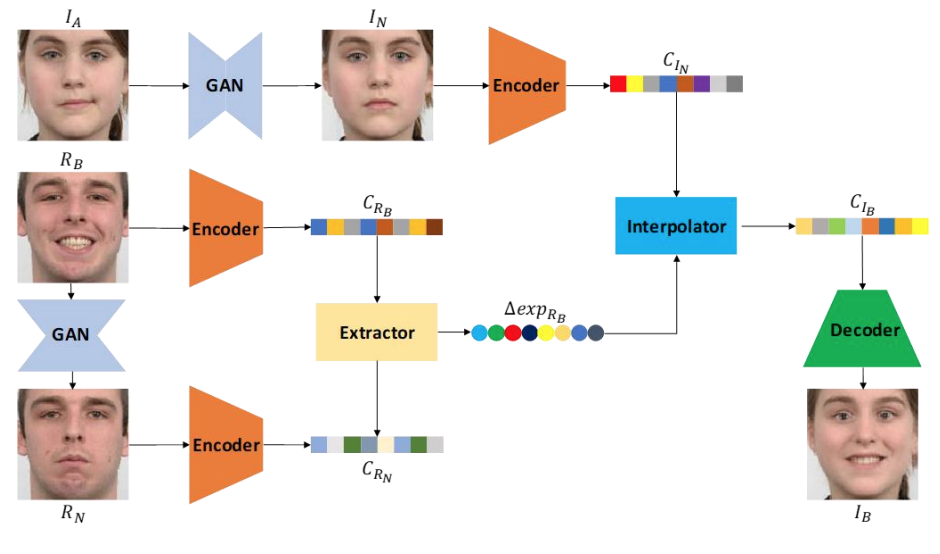}
    \caption{The overall architecture of LEED \cite{wu2020leed}.}    \label{fig:arch:wu2020leed}
\end{figure}

LEED (Label-Free Expression Editing via Disentanglement) \cite{wu2020leed} is a typical example of models which directly apply facial manipulation in the latent space of a face generator using the difference vector of the desired editing. As depicted in Figure \ref{fig:arch:wu2020leed}, an input image \(I_A\) and a reference (driving) image \(R_B\) are fed to the model. The goal is to generate an image \(I_B\) with the identity attributes of \(I_A\) and the expression attributes of \(R_B\). At first, the corresponding neutral faces \(I_N\) and \(R_N\) are obtained by a pre-trained GAN. After that the encoder respectively maps \(I_N\) , \(R_N\) , and \(R_B\) to latent vectors \(C_{I_N}\) , \(C_{R_N}\) , and \(C_{R_B}\) in the same latent space. So, the identity attributes of \(I_N\) and \(R_N\) with neutral expression attributes, and the identity and expression attributes of \(R_B\) are captured in the obtained vectors. An extractor component is then used to extract the difference vector of \(C_{R_B}\) and \(C_{R_N}\) as the attribute difference vector which is then combined with \(C_{I_B}\) in the Interpolator to generate a new latent vector \(C_{I_B}\) with the expression of \(I_B\) and the identity of \(I_A\). Finally, the Decoder translates \(C_{I_B}\) to the image space and generates the target image \(I_B\). The used loss functions in this model include reconstruction, siamese, and adversarial losses.

\begin{figure}[htb]
    \centering
    \includegraphics[width=0.45\textwidth]{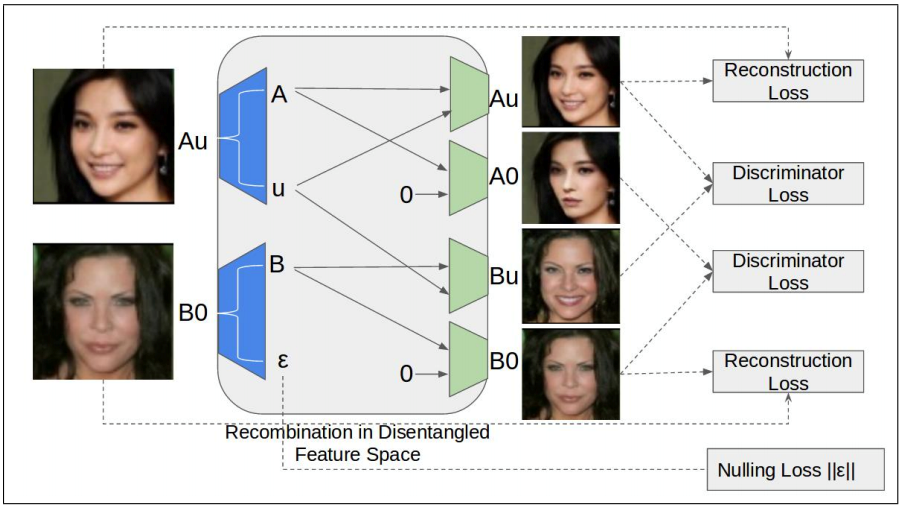}
    \caption{GeneGAN's architecture \cite{zhou2017genegan}.}    \label{fig:arch:zhou2017genegan}
\end{figure}

GeneGAN \cite{zhou2017genegan} also follows the idea of employing an encoder-decoder structure with a disentangled latent representation. As illustrated in Figure \ref{fig:arch:zhou2017genegan}, for two input images $Au$ (with identity $A$ and the target attribute $u$) and $B0$ (with identity $B$ and without the target attribute), each input image is encoded to a two-part representation separating the identity from the attribute value. During training, four codes \(Au\), \(A0\), \(Bu\), and \(B0\) are constructed as different combinations of complement codes of two input images. After that, the decoder reconstructs four legal recombinations for calculating reconstruction, adversarial, and parallelogram cycle losses. The nulling loss is also used for making the attribute vector zero for neutral images.

\begin{figure}[htb]
    \centering
    \includegraphics[width=0.48\textwidth]{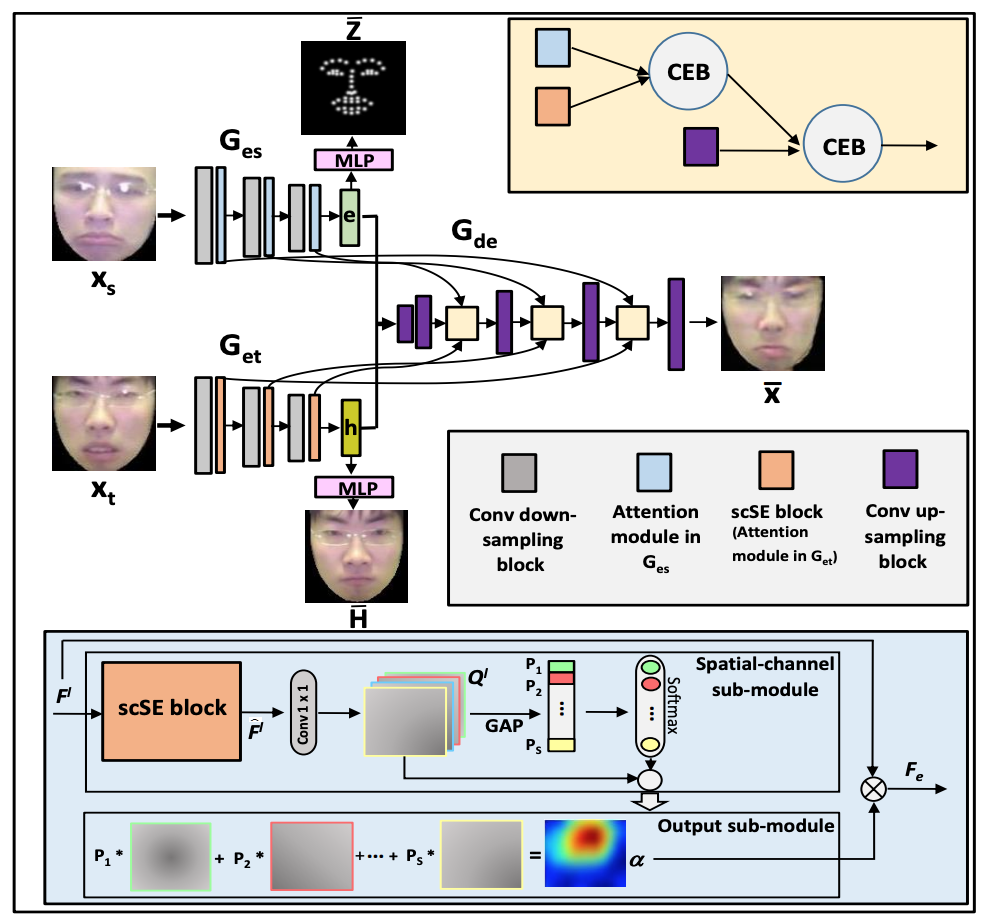}
    \caption{AIP-GAN's architecture \cite{ali2020efficient}.}    \label{fig:arch:ali2020efficient}
\end{figure}

AIP-GAN \cite{ali2020efficient} is an attention-based generative model for transferring expression from the source image \(x_s\) to the target image \(x_t\). As shown in Figure \ref{fig:arch:ali2020efficient}, the encoder \(G_{es}\) disentangles and extracts expression attributes from \(x_s\). To do this, the facial landmark image \(\bar{Z}\) is reconstructed using a supervised spatial-channel attention module. In addition to the landmark reconstruction loss, the expression classification error is used to train this module. In the other branch of the network, the encoder \(G_{et}\) learns to extract identity attributes ($h$) from \(x_t\) by trying to reconstruct the combined shape and appearance map \(\bar{H}\), i.e. the expression-agnostic identity preserving image, of the target face from  $h$. The decoder \(G_{de}\) combines the attributes learned by the intermediate layers with the disentangled attended expression and identity features from the final layers of both encoders. Various loss functions are used to train this model.

L2M-GAN \cite{yang2021n2m} manipulates the latent representation as well but first decomposes the latent representation of the input image into two related and unrelated parts using a style decomposer component. The components of the encoded style which are related to the target attribute are then transformed to a vector representing the target values of the corresponding attributes using a domain transformer module. The unrelated part of the initial style vector is then added to the transformed target style to create the style code of the edited image. 

There are lots of other models adopting the idea of swapping parts of latent codes in latent space to manipulate facial attributes, such as \cite{park2020swapping,gu2019ladn,natsume2018rsgan,lin2019exploring,awiszus2019learning,richardson2020encoding,he2019s2gan,li2019attribute,yang2019unconstrained,guo2020hierarchical}. The architecture of all these methods is similar to architectures of the models previously studied in this section.

The other main idea for editing facial attributes through latent code manipulation is to find a direction vector in the latent space that moving along that applies the desired changes on the corresponding face images. To do so, the source image is first mapped to a point in the latent space and then modified with a previously calculated direction vector. Finally, the resulting vector is transferred back to the image space via the face generator. Some models of this group are studied in the following. 

\begin{figure}[htb]
    \centering
    \includegraphics[width=0.48\textwidth]{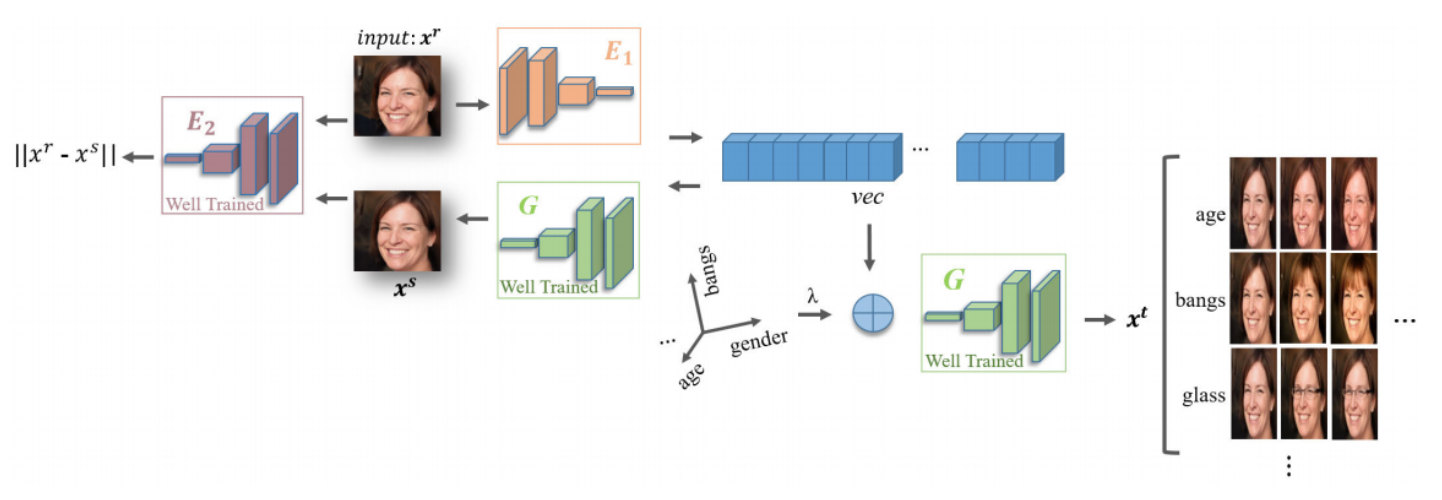}
    \caption{FEGAN's architecture \cite{ning2020fegan}.}    \label{fig:arch:ning2020fegan}
\end{figure}

FEGAN \cite{ning2020fegan} is a general model for applying continuous changes in different attributes of a target face image. The model employs a pre-trained generator \(G\) that maps each point of a latent vector space to a real face image and trains an inverse convolutional encoder \(E_1\) that embeds the input face image \(x^r\) to the latent vector \(vec\) (Figure \ref{fig:arch:ning2020fegan}). For each attribute \(y\), a \(y_{axis}\) vector is calculated as the attribute axis. Ideally, moving along an attribute axis and generating the sequence of corresponding face images only changes the target attribute of the faces. \(y_{axis}\) is obtained by a sequence regression model (\(y_{axis}=F(vec)\)) and orthogonal operations are utilized for reducing feature entanglement. Now, the latent code with the target attribute is obtained by \(vec_t = \lambda*y_{axis}+vec\) where \(\lambda\) is a controlling parameter to control the amount of attribute change. Finally, \(G\) is used to generate the target face image \(x^t\) from \(vec_t\). $E2$ is another convolution module used for training $E_1$ by calculating the feature distance between the input and generated images from \(vec\) latent vector.

Shen et al. \cite{shen2020interfacegan} propose the InterFaceGAN framework for analyzing the latent spaces learned by face GAN models and how facial attributes are coded in these spaces. The results of this study are also used in some face manipulation tasks. They assume that there is a separator hyperplane in the latent space for any binary attribute (e.g. gender). This means that the target attribute remains the same on each side of the hyperplane and is different from the other side of the boundary. To be more precise, five attributes are considered in their work including pose, smile (expression), age, gender, and eyeglasses. Also, a classifier is trained for each attribute using the annotations from the CelebA dataset with the ResNet-50 network. Next, a large number of images are generated from the utilized GAN, and attribute scores are assigned for these synthesized images. The corresponding scores are sorted for each attribute and 10K samples with the highest and the lowest scores are chosen as candidates to learn a linear SVM, resulting in a decision boundary. Finally, for each attribute, a direction vector orthogonal to the decision boundary is obtained that can be used for changing the value of that attribute.

\begin{figure}[htb]
    \centering
    \includegraphics[width=0.48\textwidth]{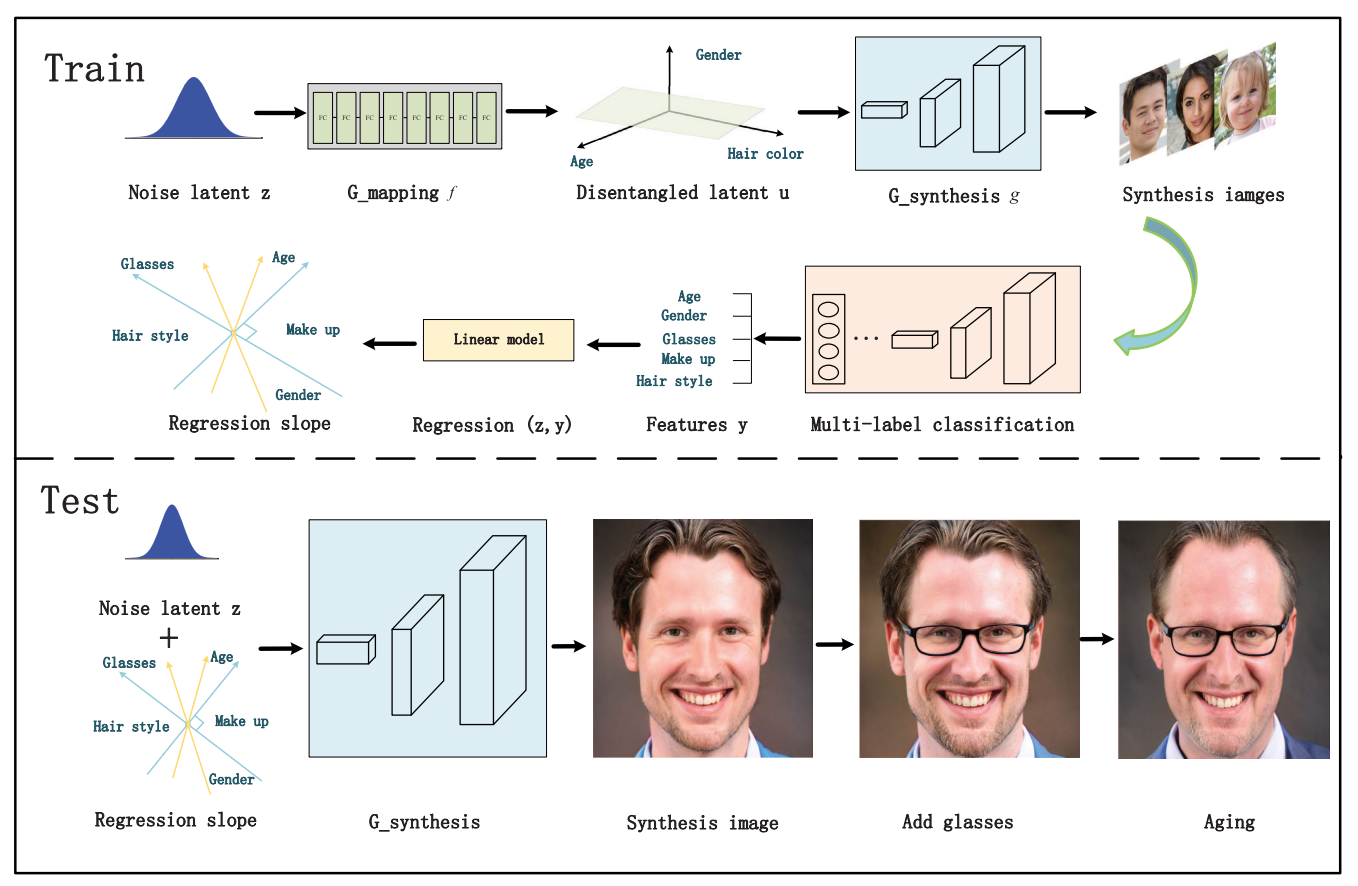}
    \caption{The architecture of the model proposed in \cite{yang2020generating}.}
    \label{fig:arch:yang2020generating}
\end{figure}

Yang et al. \cite{yang2020generating} proposed another model for face editing based on the direction vector in the latent space. In the training phase of this model, first, the mapping network $f$ generates the disentangled latent vector $u$ from the noise latent vector $z$. After that, the synthetic network $g$ generates a high-resolution image from vector $u$. Also, a multi-label classifier is trained to compute the attribute vector $y$ for the generated images. Then, to achieve the attribute axis and direction, the regression is applied on the latent vector z and the feature vector y, and the slope of the regression is considered as the attribute axis. In the test phase of this model, the input is a random noise latent vector $z$  to the synthetic network $g$ and the output is a synthesized image. By moving the latent vector in the direction of the extracted feature axis, the changes in the attribute are applied to the generated images through the network $g$. The architecture of this model along with the results of moving the initial noise vector along the glasses and aging axes are shown in Figure \ref{fig:arch:yang2020generating}. 

\begin{figure}[htb]
    \centering
    \includegraphics[width=0.48\textwidth]{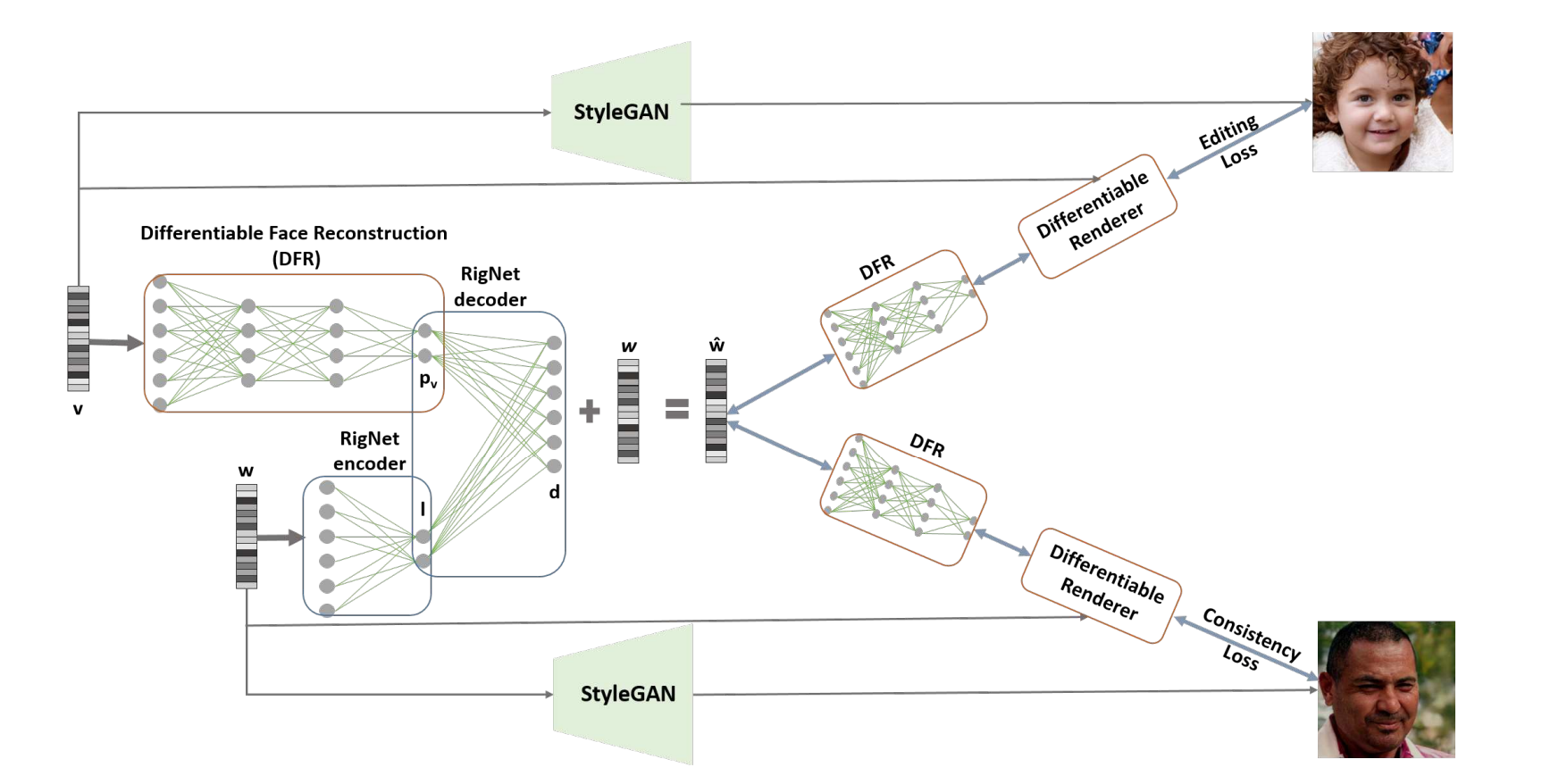}
    \caption{StyleRig's architecture \cite{tewari2020stylerig}.}    \label{fig:arch:tewari2020stylerig}
\end{figure}

StyleRig \cite{tewari2020stylerig} is another model that uses the attribute direction vectors in the latent space. However, the attribute direction vectors are calculated by a rigger network (RigNet). As the architecture of the model is shown in Figure \ref{fig:arch:tewari2020stylerig}, this model includes three main parts: a face generator (StyleGAN), a differentiable face reconstruction (DFR) unit, and a learned rigger network (RigNet). The DFR and StyleGAN networks are pre-trained and only RigNet is trainable in this architecture. The goal is to generate a latent code $\hat{w}$ that is translated to the target face with the control parameter encoded in latent code $v$ and the identity of $w$. $\hat{w}$ is computed as $\hat{w}= d+w$ where $d$ is the difference vector produced by RigNet. The RigNet consists of two encoder and decoder parts. The encoder transforms the $w$ to a lower dimensional representation $I$ and the decoder converts $I$ and the target attributes $P_v$, extracted from $v$ by the DFR network, to the difference vector $d$. The model is trained with reconstruction, edit, and consistency losses. 

There are some other models based on the main idea of the direction vector but with slight differences in details. For example, \cite{upchurch2017deep,dogan2020semi} achieve the attribute direction vector by calculating the difference of the latent representations of two groups of images with and without the desired attribute. Also, \cite{chen2018facelet} obtains the attribute direction by training a new encoder-decoder network from different image domains based on the desired attribute.

\begin{figure}[htb]
    \centering
    \includegraphics[width=0.48\textwidth]{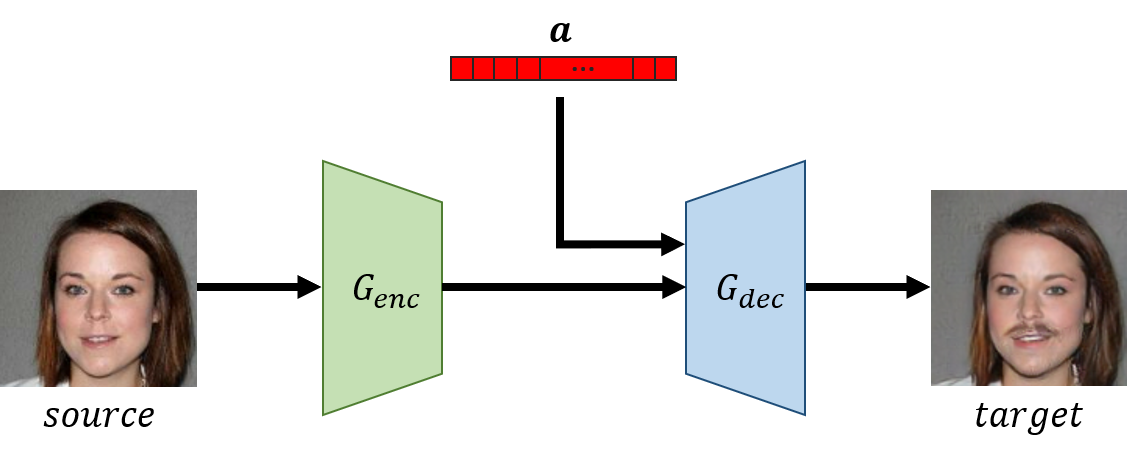}
    \caption{The overall architecture of conditional decoder models.}    \label{fig:arch:conditionaldecoderoverall}
\end{figure}
\textbf{Conditional decoder:}
This group of face editing models receive an attribute vector \(a\) along with the input image and modify the input image so as to possess the required attributes. Theses models are intrinsically a face generator with an encoder-decoder structure (Figure \ref{fig:arch:conditionaldecoderoverall}). The encoder maps the input image to a latent space representation \(z\) that along with the attribute vector \(a\) are fed into the decoder to synthesize the edited face image conditioned on \(a\). The most important component of these models is their \emph{conditional decoder}.

\begin{figure}[htb]
    \centering
    \includegraphics[width=0.48\textwidth]{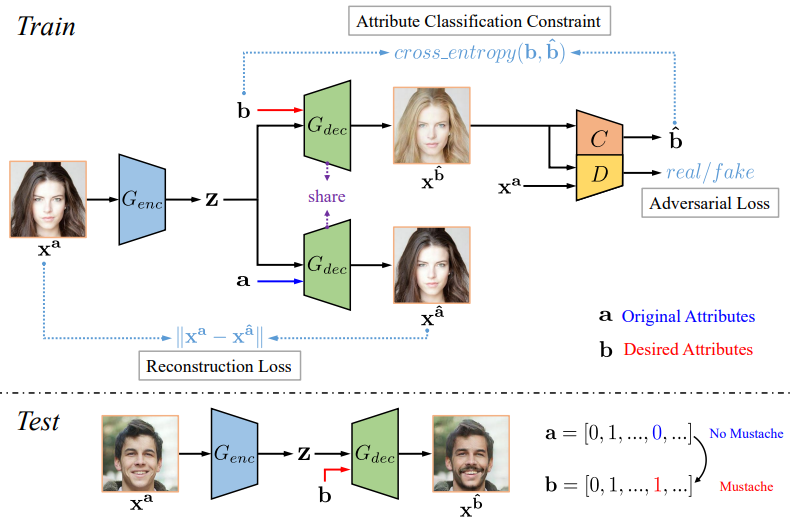}
    \caption{AttGAN's architecture \cite{he2019attgan}.}
    \label{fig:arch:attgan}
\end{figure}

AttGAN \cite{he2019attgan} is an exemplary model of this category. The train and test block diagrams of this model are shown in Figure \ref{fig:arch:attgan}. The test phase completely matches the general architecture of this group demonstrated above. Here, $(x^a)$ and $(x^{\hat{b}})$ are the source and target images, $(a)$ is the source image's attribute vector and $(b)$ is the desired attribute vector. Examples of $(a)$ and $(b)$ vectors are given in Figure \ref{fig:arch:attgan} which result in the addition of mustache to the original face image. $(G_{enc})$ is a stack of convolutional layers and $(G_{dec})$ is a stack of transposed convolutional layers with u-net symmetric skip connections. The model is trained with three different loss functions (Figure \ref{fig:arch:attgan}). In the training phase, the edited image $(x^{\hat{b}})$ is fed to an attribute classifier $(C)$ to extract its attribute vector $(\hat{b})$ and compare it to desired attributes $(b)$ to calculate the cross entropy loss. As in other GAN models, a discriminator $(D)$ is utilized to calculate the adversarial loss. In another branch, $(G_{dec})$ is fed with the attribute vector of the source image with the aim of reconstructing the original source image and the pixel-wise difference between $(x^a)$ and $(x^{\hat{a}})$ is considered as the reconstruction loss.

\begin{figure}[htb]
    \centering
    \includegraphics[width=0.48\textwidth]{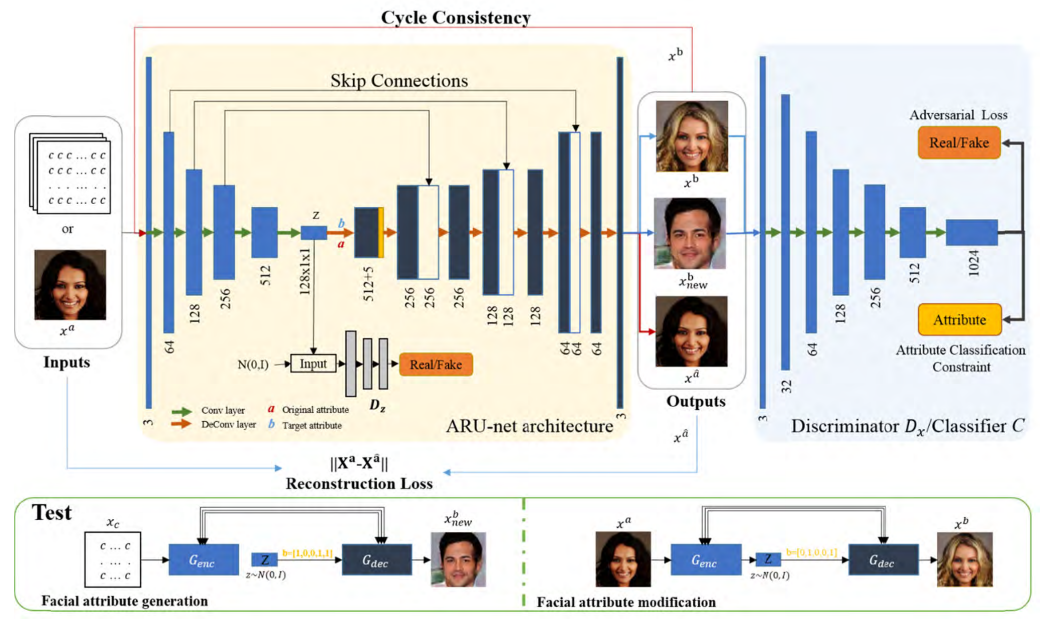}
    \caption{The architecture of the editing model proposed in \cite{zhang2019adversarially}.}
    \label{fig:arch:zhang2019adversarially}
\end{figure}

Zhang et al. \cite{zhang2019adversarially} have proposed an adversarially regularized U-net (ARU-net)-based generative adversarial networks (ARU-GANs) for facial attribute generation and modification. The general architecture of the model, as shown in Figure \ref{fig:arch:zhang2019adversarially} , completely fits the conditional decoder architectures: the input image \(x^a\) is encoded  as latent vector \(z\) and the decoder translate \(z\) conditioned on the attribute vector \(b\). As the name of the model implies, the model is based on a U-net structure with skip-connections and adversarial loss is used to guide the latent vector to match the prior distribution \(N(0,I)\). The decoder is conditioned on the desired attribute vector to map the latent space representation to the target image.
Cycle consistency and reconstruction losses are utilized to train the model in a manner that edits the image without altering its identity. A classifier \(C\) along with discriminator \(D\) is trained and leads the model to produce images which satisfy desired attributes.
\begin{figure}[htb]
    \centering
    \includegraphics[width=0.48\textwidth]{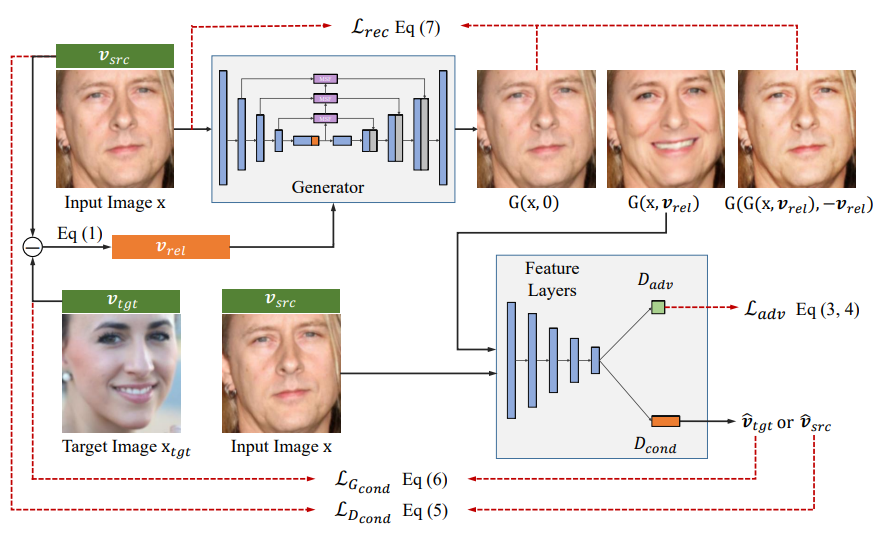}
    \caption{The overall architecture of the editing model proposed in \cite{ling2020toward}.}
    \label{fig:arch:ling2020toward}
\end{figure}

Ling et al. \cite{ling2020toward} propose a U-net based architecture for continuous facial attribute editing (Figure \ref{fig:arch:ling2020toward}) employing the notion of discrete relative action units (AUs). Relative action unit vector \(v_{rel}\), initially proposed by \cite{friesen1978facial}, precisely indicates the desired changes to facial muscles and provides the model with fine control over different parts of the face. As before, this relative AUs vector is fed into the generator along with the embedded input image. The other contribution of this work is that the skip connections of the U-net are replaced with multi-scale feature fusion (MSF) modules that fuse features from different scales and then transfer them from the decoder to the encoder side of the network. The loss functions of the model are reconstruction, cycle consistency, adversarial and condition losses. 

\begin{figure}[htb]
    \centering
    \includegraphics[width=0.48\textwidth]{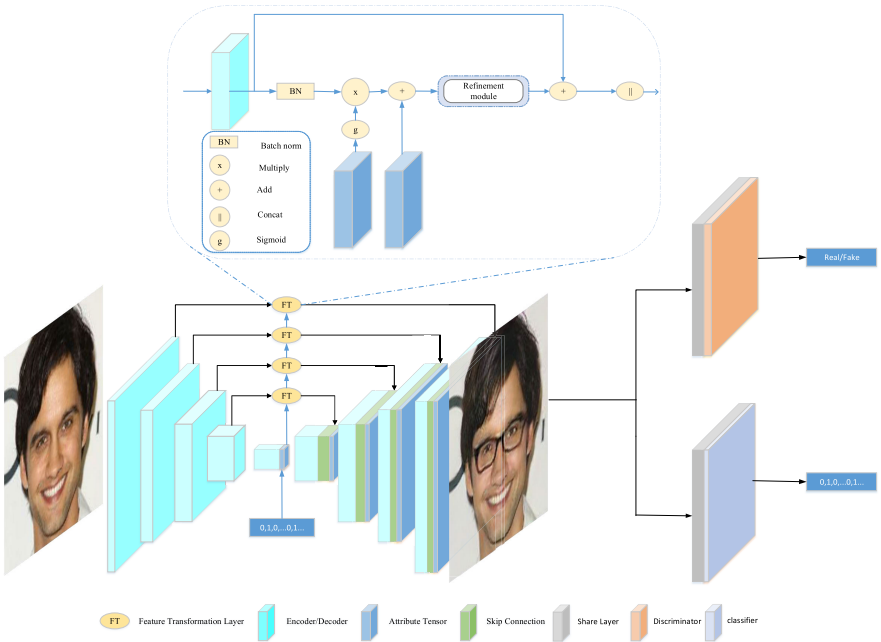}
    \caption{InjectionGAN's architecture \cite{ding2020injectiongan}.}
    \label{fig:arch:injectiongan}
\end{figure}

InjectionGAN \cite{ding2020injectiongan}, shown in Figure \ref{fig:arch:injectiongan}, is another U-net based conditional model in which the condition vector is concatenated to the embedding extracted by the encoding part of the model. This model proposed a Feature Transformation (FT) module which is embedded in skip connections. FT modulates normalized low-level feature maps with scaling and shifting parameters \(\alpha\) and \(\beta\) which are generated by two separate convolutional layers with the guidance of the target attribute vector. The refinement module takes modulated normalization layer, extracts salient features, and discards features not related to attribute changes. FT learns to apply coarse-to-fine semantic modification in which more global features are transformed at the outer layers and modification of local features is up to inner layers.
Discriminator \(D\) shares a stack of convolutional layers and two separated branches with the same architecture for \(D_{adv}\) and \(D_{cls}\) that computes adversarial and attribute classification losses, respectively. Also, an image reconstruction loss is used to preserve identity through modification.

\begin{figure}[htb]
    \centering
    \includegraphics[width=0.48\textwidth]{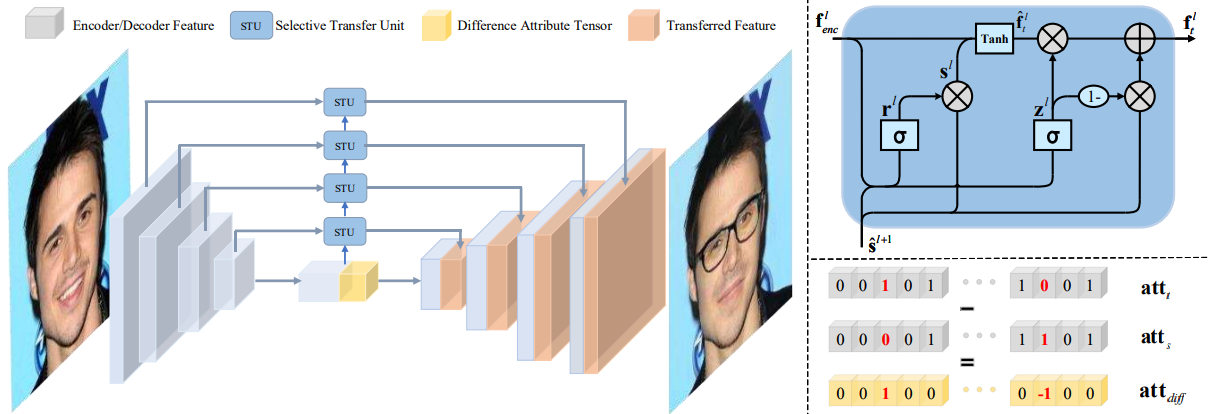}
    \caption{STGAN's architecture \cite{liu2019stgan}.}    \label{fig:arch:stgan}
\end{figure}

STGAN \cite{liu2019stgan} architecture is depicted in Figure \ref{fig:arch:stgan}. As shown in the left side of this figure, STGAN has an encoder-decoder architecture. \(G_{enc}\) encodes input image into an abstract representation via five convolution layers and \(G_{dec}\) contains five transposed convolution layers and generates output image from the abstract representation. However, in this model, difference attribute vector \(\textbf{att}_{diff}=\textbf{att}_t-\textbf{att}_s\) is given to model rather than the whole target attribute vector that may result in undesired changes in the output image. 
A discriminator \(D\) is utilized to train the model with two adversarial and attribute manipulation losses respectively calculated by \(D_{adv}\) and \(D_{att}\) branches of the discriminator. A reconstruction loss is used to verify that the image generated from \(att_{diff}=0\) is similar to the input image.
STGAN provides selective transfer units (STUs), which replace well-known skip-connections in U-Net structures and are applied after each encoder layer. STUs transfer only selected features from the encoder to the decoder. These features are compatible with decoder features and complement them. GRU structure is modified to build STUs.

\begin{figure}[htb]
    \centering
    \includegraphics[width=0.48\textwidth]{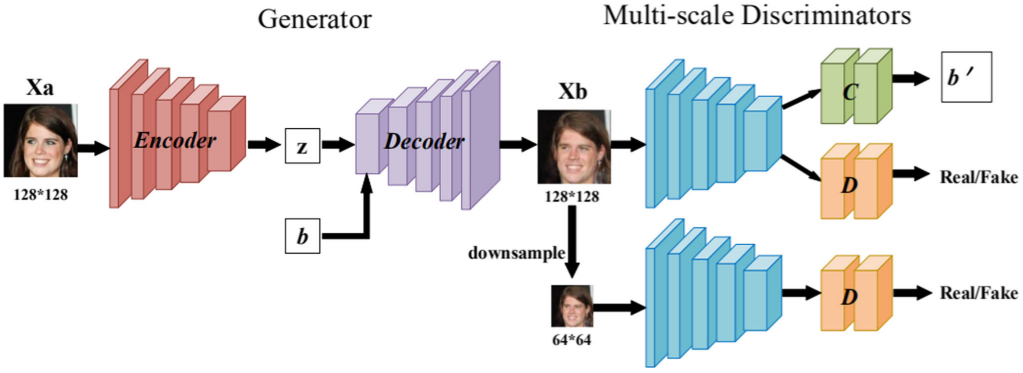}
    \caption{The architecture of the facial attribute editing model proposed by Song et al. \cite{song2020face}.}
    \label{fig:arch:song2020face}
\end{figure}

As depicted in Figure \ref{fig:arch:song2020face}, the architecture of the facial attribute editing model proposed by Song et al. \cite{song2020face} completely matches the base architecture of this category. An encoder network encodes input image \(X_a\) to a latent space representation \(z\) and decoder decodes \(z\) to the edited image \(X_b\) according to the target attribute vector \(b\) representing \(n\) binary attributes. Attribute classification and reconstruction losses are utilized during training. 
The main contribution of this model is its two-scale discriminator in which \(D_1\) is applied on \(128 \times 128\) images and capture detailed information, while \(D_2\) captures more global information on \(64 \times 64\) images.


FadNet \cite{lample2017fader} was one of the earlier facial attribute editing models with an encoder-decoder architecture. Encoder maps input image to the latent representation \(E(x)\) which is invariant to non-identity attributes. On the other side, the decoder takes this latent representation alongside the desired attribute \(y\) to generate the same person as input conditioned on the target attribute.
Lack of paired data from the same person with and without different attributes motivates the authors to use an adversarial penalty on the latent space in addition to typical adversarial loss used on image space. The intended discriminator tries to predict \(y\) given latent space representation \(E(x)\), hence, the encoder learns to result in a representation independent from attributes and makes it impossible for the discriminator to predict \(y\) from \(E(x)\). Adversarial and reconstruction losses are used in order to train the model.

\begin{figure}[htb]
    \centering
\includegraphics[width=0.48\textwidth]{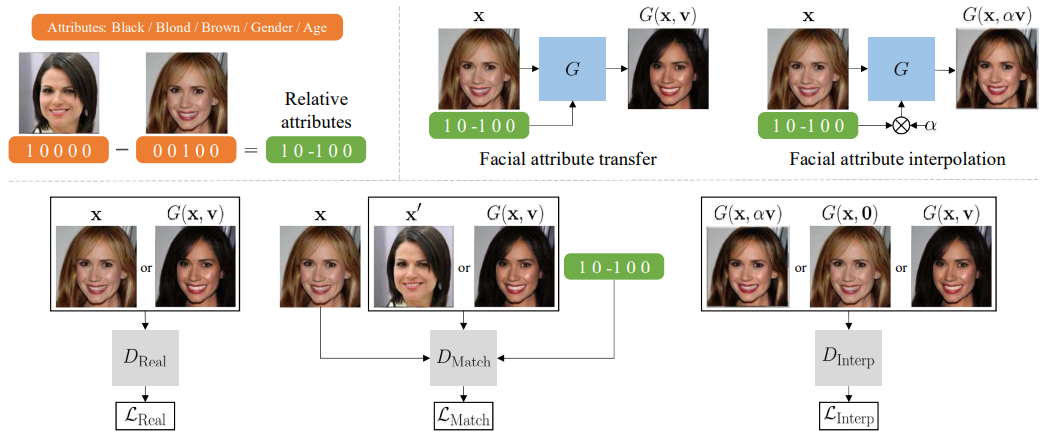}
    \caption{The architecture of RelGAN \cite{wu2019relgan}.}
    \label{fig:arch:relgan}
\end{figure}

RelGAN \cite{wu2019relgan} is another model of this category that utilizes a conditional decoder along with an attribute difference vector. As shown in Figure \ref{fig:arch:relgan}, relative attributes are defined as the difference between an n-dimensional vector $(a)$ representing attributes of the input image  and attribute vector $(b)$ corresponding to the target image as  $(v=b-a)$. The conditional generator of RelGAN is then used to modify the input image $(x)$ in accordance with relative attribute vector $(v)$ to give the final manipulated face image $(G(x,v))$. Relative attribute vector provides an easy way to interpolate between $(x)$ and $(G(x,v))$ as $(G(x,\alpha v))$ where $(\alpha)$ is an interpolation coefficient. 
Three different discriminators are contributing to model training. $(D_{Real})$ is used for adversarial loss and distinguishes real and fake images. Conditional discriminator $(D_{Match})$ aims to minimize the difference between condition vector $(v)$ and the vector obtained as the difference between input and generated images' attributes. 
The input of $(D_{Match})$ is the triplet $((x,v,x'))$ where $(x)$  is the real input image, $(x')$ is another real image or the fake image $(G(x,v))$, and $(v)$ is a relative attribute vector, and the output is $(+1)$ for real and matched triplets and $(-1)$ for fake or mismatched triplets.  There is also a reconstruction loss consisting of two penalties; one for cycle-reconstruction which aims to reproduce the input image by editing back the edited image and the other for self-reconstruction which aims to generate the same image as input by editing with $(v=0)$ condition. Since this model provides interpolation generation, $(D_{interp})$ aims to make interpolated image $(G(x,\alpha v))$ indistinguishable from non-interpolated images. To this end, $(D_{interp})$ estimates the degree of interpolation of an input image and should return zero for non-interpolated images.

\begin{figure}[htb]
    \centering
    \includegraphics[width=0.48\textwidth]{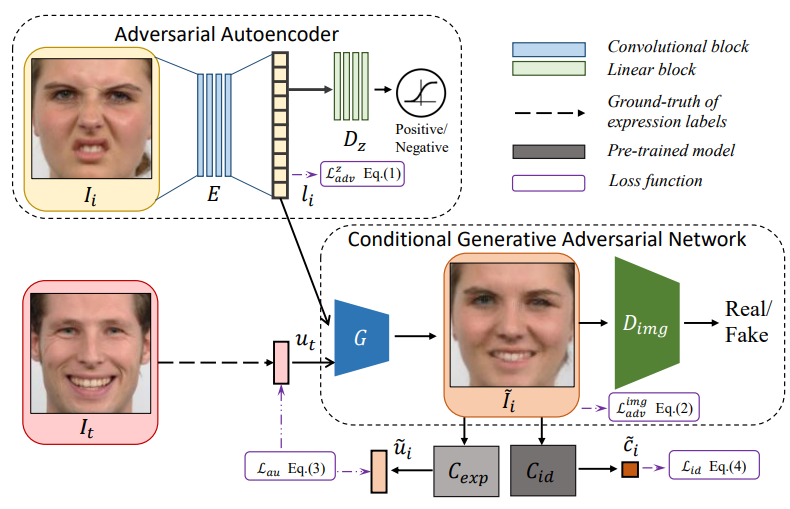}
    \caption{Architecture of fine grained expression manipulation model \cite{tang2020fine}}.
    \label{fig:arch:tang2020fine}
\end{figure}

Figure \ref{fig:arch:tang2020fine} belongs to \cite{tang2020fine} that follows the base architecture of this category to edit facial expression. An adversarial autoencoder encodes the input image \(I_i\) to the latent code \(l_i\) which is in local Euclidean space with structural. A discriminator \(D_z\) ensures that the generated codes follow a normal distribution. The conditional GAN module uses a target action units (AUs) vector as a condition to generate face image from the obtained latent code \(l_i\), and \(D_{img}\) is a typical GAN discriminator distinguishing between real and fake images.
\(C_{exp}\) and \(C_{id}\) are two classifiers whit the same architecture that lead the model to generate images with the expected expression and the original identity, respectively.

\begin{figure}[htb]
    \centering
    \includegraphics[width=0.48\textwidth]{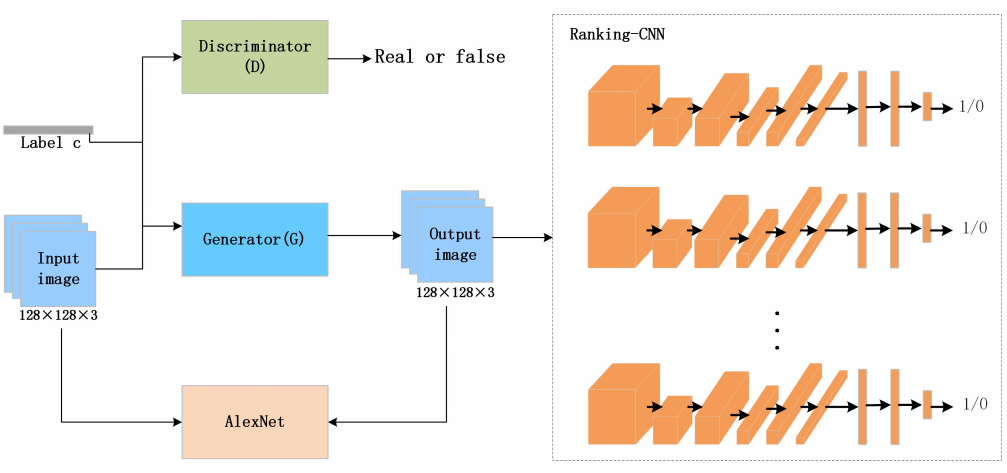}
    \caption{The architecture of a conditional GAN supervised by Ranking-CNN for face aging model \cite{sheng2020face}.}
    \label{fig:arch:sheng2020face}
\end{figure}

Figure \ref{fig:arch:sheng2020face} shows the architecture of the aging model proposed by Shen et al. \cite{sheng2020face} that employs a face generator \(G\) conditioned on the age group \(c\). Discriminator \(D\) discriminates real images from fake ones and provides the adversarial loss. A pretrained AlexNet is used to calculate perceptual features of the input and generated images and performs the the identity-preserving role. A ranking-CNN which consists of a series of basic CNNs is used to estimate the age of the generated face. The ranking-CNN is shown to provide more accurate results compared to a multi-class classifier.

\begin{figure}[htb]
    \centering
    \includegraphics[width=0.48\textwidth]{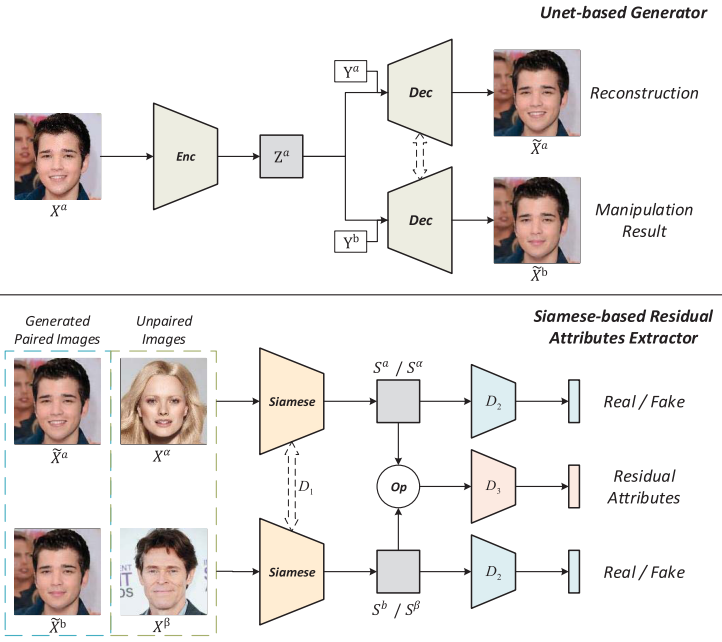}
    \caption{ResAttr-GAN's architecture \cite{tao2019resattr}.}    \label{fig:arch:resattrgan}
\end{figure}

ResAttr-GAN \cite{tao2019resattr}, as shown in Figure \ref{fig:arch:resattrgan}, comprises two main components: an encoder that maps the input image \(X^a\) with non-identity attributes \(a\) to latent representation \(Z^a\), and a conditional decoder that generates an output image with the same identity of \(Z^a\) and the set of attributes given by a \(Y\) vector; \(Y^a\) causes the model to reconstruct the input image and \(Y^b\) changes the attributes to the state determined by \(b\). As before, the model is trained with reconstruction and real/fake adversarial losses. The main contribution of this model is the introduction of a new loss function calculated by means of a residual attributes extractor (ResAttr) shown in Figure \ref{fig:arch:resattrgan}. Residual attributes are related to the non-identity attributes which can be the target of manipulation in this method and show the difference between the attributes of the two images (the generated image and the reference image). The aim of ResAttr discriminator \(D\) is to reduce the difference between the residual attributes of the generated image and the reference image.  As paired data with the same identity and different attributes are not available, the residual attributes extraction network is trained in a deep feature space using Siamese networks that extract high-level features from input images, discriminator \(D_2\) that distinguishes between data sources, and \(D_3\) network that extracts residual attributes from input images based on the corresponding extracted features.

\begin{figure}[htb]
    \centering
    \includegraphics[width=0.48\textwidth]{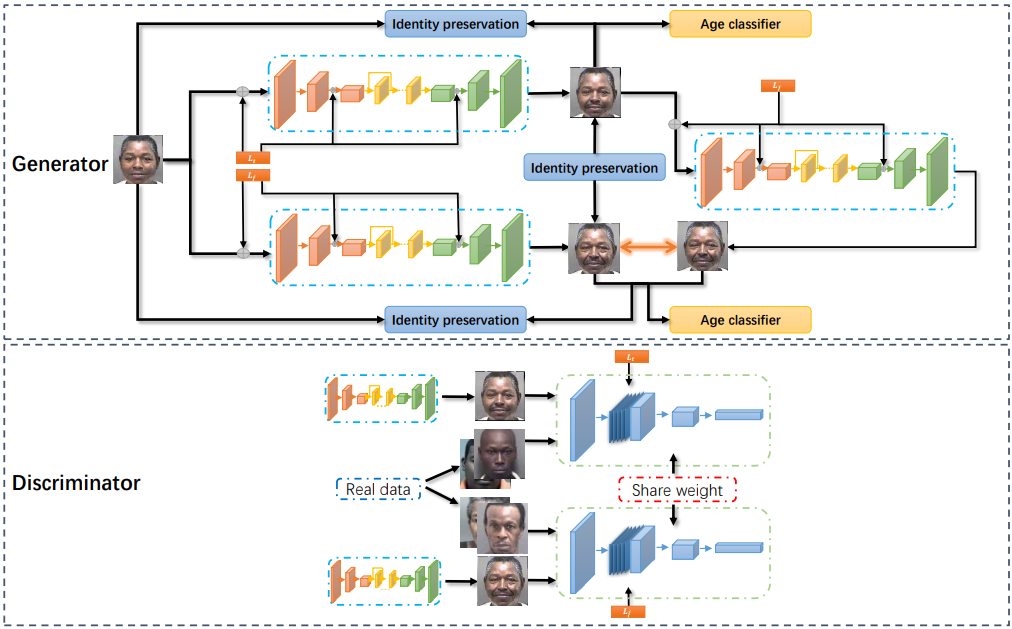}
    \caption{Triple-GAN's architecture \cite{fang2020triple}.}
    \label{fig:arch:triplegan}
\end{figure}

Triple-GAN \cite{fang2020triple} is an encoder-decoder based model which has been proposed for face aging. The input image is edited from two different paths conditioned on different one-hot vectors \(L_t\) and \(L_f\) determining two different age classes and results in two synthesized faces \(G(x,L_t)\) and \(G(x, L_f)\), respectively. Once again, the generator is applied to \(G(x,L_t)\) conditioned on \(L_f\) to generate \(G(G(x,L_t),L_f)\) which is expected to be equal to \(G(x,L_f)\).
Enhanced adversarial loss, age classification loss, and identity preservation loss guide the training process. Besides, the triple translation loss is introduced for this model to ensure that two generated images from the two aforementioned paths are the same. To do so, the \(L2\) norm between \(G(x,L_t)\) and \(G(G(x,L_f),L_t)\) is minimized.

\begin{figure}[htb]
    \centering
    \includegraphics[width=0.48\textwidth]{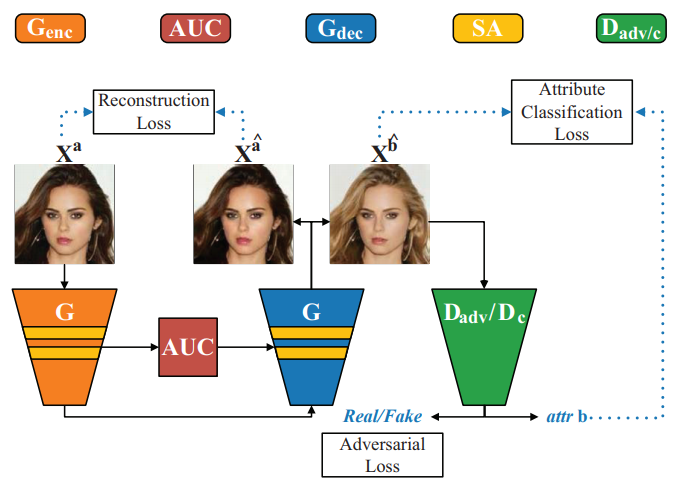}
    \caption{The architecture of MU-GAN model \cite{zhang2020mu}.}
    \label{fig:arch:mugan}
\end{figure}

MU-GAN \cite{zhang2020mu} is another conditional-decoder based model in which the target attribute vector \(b\) is concatenated with the representation of the innermost layer of the encoder and fed to the decoder(Figure \ref{fig:arch:mugan}). In this model, the base network of the generator is a symmetric U-net structure whose direct skip connections are replaced with attention U-net connections (AUCs).  AUCs improve the quality of the generated image and the preservation of the attribute-irrelevant details of the input image by selective transfer of decoder-side features to the encoder side. On the other hand, a self-attention mechanism supplements convolutional layers to enable log-range dependency on spatially separated regions and to consider global geometric information in these layers.
Discriminator \(D\) has two sub-networks \(D_{adv}\) for real and fake discrimination and \(D_c\) for attribute verification in the generated image. Common losses, i.e. adversarial, attribute classification, and reconstruction losses are also used to train the model.

PIRenderer \cite{ren2021pirenderer} employs two consecutive encoder-decoder structures to control the face motions based on the 3D morphable face models. The goal of the first (warping) network is to generate a coarse image with the desired pose and expression by producing a flow filed for warping the input image. The second (editing) network is just an image to image translation network that increases the resolution of the coarse image and generates the final image. The target style is first described through the parameters of the 3D face model and then embedded into a latent vector $z$ used in both networks to guide the production of the face images.

\begin{figure}[htb]
    \centering
    \includegraphics[width=0.48\textwidth]{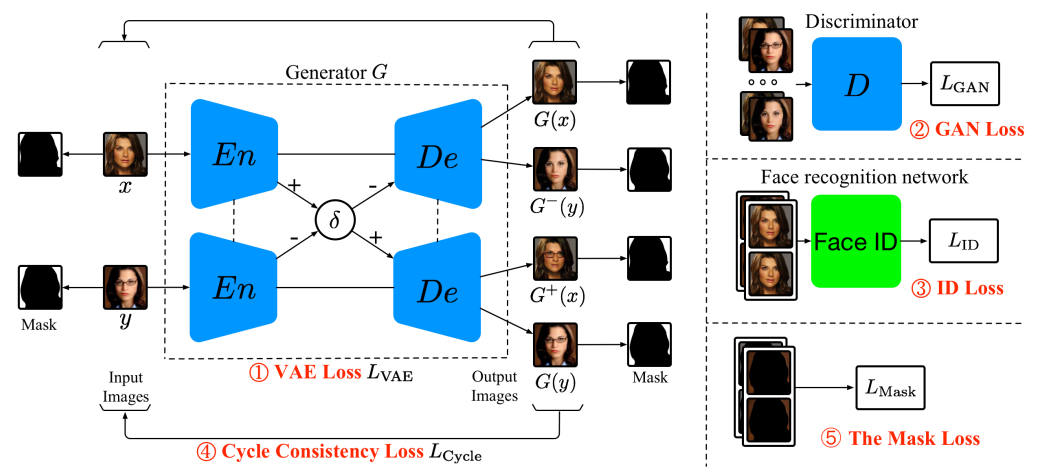}
    \caption{M-AAE's architecture \cite{sun2021mask}.}
    \label{fig:arch:maae}
\end{figure}

Sun et al. proposed Mask-Adversarial AutoEncoder (M-AAE) \cite{sun2021mask} for continuous manipulation of a target attribute. M-AAE, as shown in Figure \ref{fig:arch:maae}, is based on a VAE-GAN framework (Figure \ref{fig:arch:maae}) which its generator \(G\) is a VAE encoder-decoder. The input image \(x\) is transformed to a feature map by the encoder and then modified by applying a relative value \(\pm \delta\) in order to have the desired attributes. The modified features are translated to the final manipulated image by the decoder of VAE. The whole process can be described as \(G^+(x)= De(En(x) + \delta)\) for adding the desired attribute to the input image \(x\). The value of \(\delta\) controls the strength of the attribute and it has been discussed that modifying a small number of elements of the feature map helps the model to better preserve the details of the input image. 
M-AAE is trained by several loss functions. The VAE losses (prior regularization and reconstruction error) are used to train VAE components. The discriminator \(D\) provides the adversarial loss. Moreover, the face recognition (ID) and the cycle consistency losses improve the identity and detail preservation of the model. Finally, a mask-aware loss function is utilized to preserve the background of the input image by penalizing difference between mask-out regions in input and output images.
It should be noted that this model is not considered as a mask-guided model (defined in Section \ref{photo-guided}) in this survey as the mask is not used for manipulating the facial attributes. 

\cite{deng2020reference} converts face editing to an inpainting task in which a reference image and a source face image along with a face component mask are taken as input and the component determined by the mask is removed from the source image. A reference-guided encoder is then used to extract information of the desired component from the reference image and an inpaint generator is used to complete the corrupted input images based on this information. In this model, the encoded feature from the reference image is used as decoder's condition. 
 
\subsubsection{Image-to-image translation}
Some researchers have regarded semantic facial attribute editing as an image-to-image translation task in which the source and target domains correspond to face image sets with two different values of a specific attribute. Here, the condition (attribute) vector is no longer required as an input of the generator as the generator is trained to perform the same edit on all input images. In this section, these models are reviewed in two main categories: bi-modal models performing a special edit on single attributes and multi-modal models supporting the manipulation of several attributes.
\begin{figure}[htb]
\centering
\subfloat[]{\includegraphics[width=2.5in]{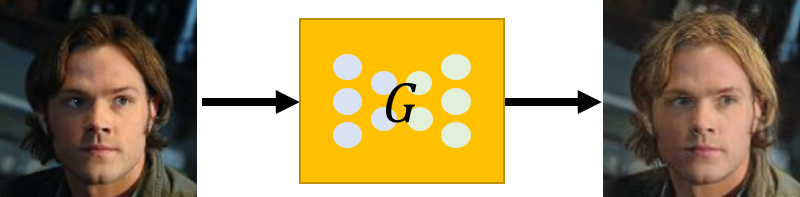}%
\label{fig:arch:oneway}}
\hfil
\subfloat[]{\includegraphics[width=2.5in]{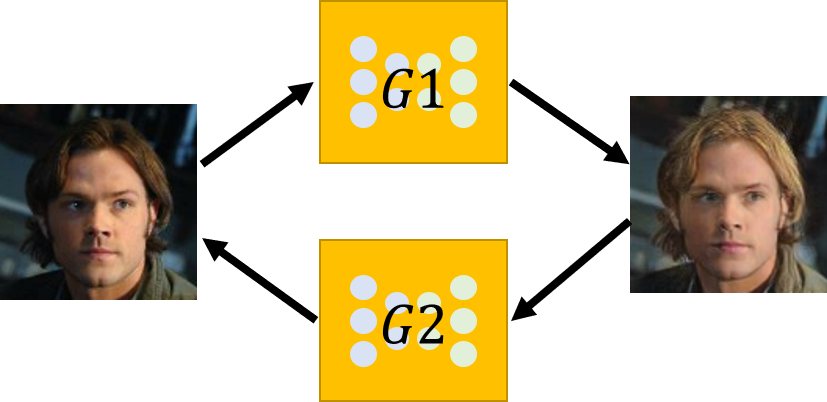}%
\label{fig:arch:twoway}}
\caption{The overall architectures of one-way and two-way bi-modal image-to-image translation models. (a) One-way. (b) Two-way.}
\label{fig:arch:bimodal}
\end{figure}

\textbf{Bi-modal image-to-image translation models:}
In the realm of facial attribute editing methods, bi-modal image-to-image translation models are utilized to change a specific attribute of the input face image to have a predefined value.  The target attributes of this category are either binary features (e.g., has-glasses) or considered as binary features (e.g., age treated as young/old). For example, a model may be trained to add glasses to a face photo or to increase its age. 
As shown in Figure \ref{fig:arch:bimodal}, the models of this category can be further divided into two main subcategories: one-way and two-way models. The one-way methods (Figure \ref{fig:arch:oneway}) have only one path to add/remove the particular attribute to/from an image while two-way methods (Figure \ref{fig:arch:twoway}) run in two opposite directions -one to add an attribute to the image and the other to remove the attribute.

\begin{figure}[htb]
    \centering
    \includegraphics[width=0.48\textwidth]{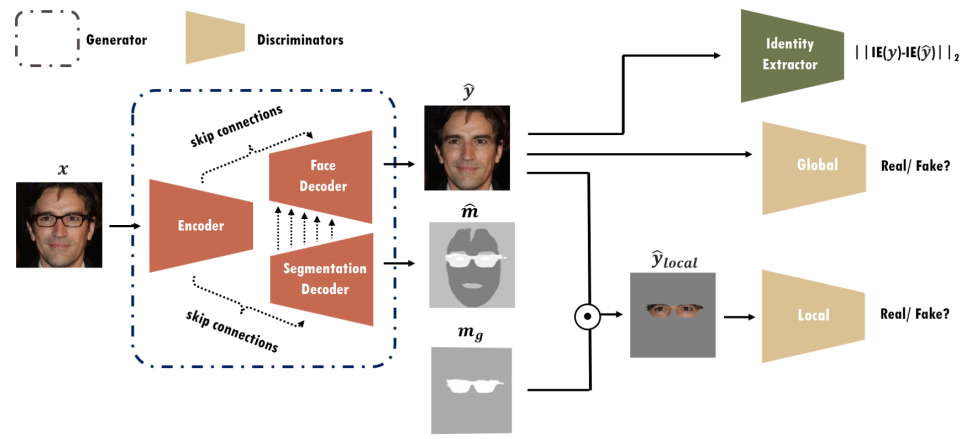}
    \caption{ByeGlassesGAN's architecture \cite{lee2020byeglassesgan}.}
    \label{fig:arch:byeglassesgan}
\end{figure}

ByeGlassesGAN \cite{lee2020byeglassesgan} is a one-way bi-modal image-to-image editing framework particularly developed for eyeglasses removal from face photos. As depicted in Figure \ref{fig:arch:byeglassesgan}, the model comprises a generator, an identity extractor, and two discriminators. Generator is composed of one encoder and two decoders: the face decoder that generates a glasses-free face from the embedded representation and the segmentation decoder that generates a mask for specifying the glasses region. The skip connections of the generator convey the characteristics of face pixels from the encoder to the decoder parts to help them to preserve consistency with neighboring pixels in inpainting the eye region.
Two global and local discriminators of ByeGlassesGAN make sure the whole generated image and the local inpainted region are realistic. The identity extractor module is used to minimize the distance between the identity feature vectors of the input and generated images. A set of synthesized paired images (with and without glasses) is constructed for training the model.

\begin{figure}[htb]
    \centering
    \includegraphics[width=0.48\textwidth]{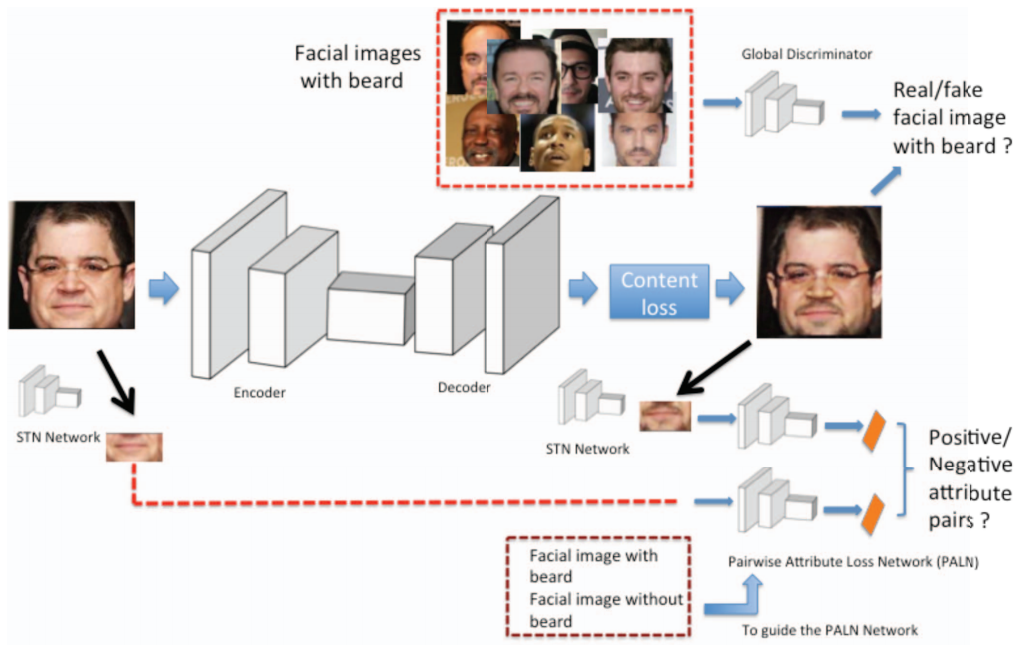}
    \caption{The architecture  of the Wang et al.'s model \cite{wang2018weakly}.}
    \label{fig:arch:wang2018weakly}
\end{figure}

Wang et al. \cite{wang2018weakly} provides a unified framework for learning to transfer images between different domains, e.g. from “hair” to “bald” or from “no beard” to “goatee beard”, given a set of unpaired reference images of each domain. Their proposed model, as shown in Figure \ref{fig:arch:wang2018weakly}, is a one-way bimodal image-to-image translation model with an encoder-decoder generator. The main contribution of their model is the introduction of a perceptual content loss and a local pairwise loss; the former tries to force the generated image to have similar content as the input image, and the latter causes the model to generate images with the desired attributes.  The content loss is defined as the average distance between the feature vectors extracted from the original and the generated images at different layers of the VGG network. The pairwise attribute loss is an adversarial loss realized through a spatial transform network (STP) \cite{jaderberg2015spatial} which determines the regions of an input image that are related to the target attribute, and a pairwise attribute loss network (PALN) which predicts the labels of the two regions extracted by STP from an image pair. It is worth mentioning that there is no need for the training image pairs to belong to the same person and any two faces from the two domains suffice.

\begin{figure}[htb]
    \centering
    \includegraphics[width=0.48\textwidth]{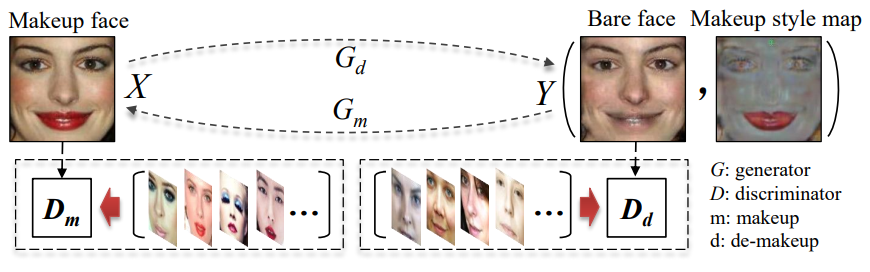}
    \caption{ The architecture of BTD-Net \cite{cao2019makeup}.}
    \label{fig:arch:cao2019makeup}
\end{figure}

The Bidirectional Tunable De-makeup Network (BTD-Net) \cite{cao2019makeup} is a deep bidirectional de-makeup network with a cycle structure (Figure \ref{fig:arch:cao2019makeup}). Though both makeup and de-makeup directions are considered, the ultimate goal of the model is only to remove makeup effects from the face images, and the other direction is only used to improve the learning of this task. 
As other two-way bimodal translation models, BTD-Net has two generative networks \(G_m\) and \(G_d\) which are responsible for adding and removing makeup, respectively. ResNet blocks are injected into the U-net structure of the generators to help them preserve the makeup-irrelevant information. Due to the lack of paired data, two distinct discriminators \(D_m\) and \(D_d\) are considered for each generator to enforce the generated images in each way to be consistent with the corresponding target domain. The makeup process is a one-to-many mapping as the cosmetics can be applied to different regions of the face. So, a makeup style map \(S\) is introduced as an additional output \(G_d\) and as an input condition guiding \(G_m\) in producing the face with the corresponding makeup. BTD-Net uses adversarial and cycle consistency losses during the training process.

Wang et al. \cite{wang2019face} follow the similar CycleGAN architecture for age progression. Youth and elder photos form the two image domains of the model and two generators, perform the image translations between them – as BTD-Net the focus of the paper is only on one generator (aging direction).  In addition to the common cycle loss, a new bias loss is introduced to reduce the artifact of the output images, which is defined as the difference between the output of the discriminator and a constant value obtained experimentally.

\begin{figure}[htb]
    \centering
    \includegraphics[width=0.48\textwidth]{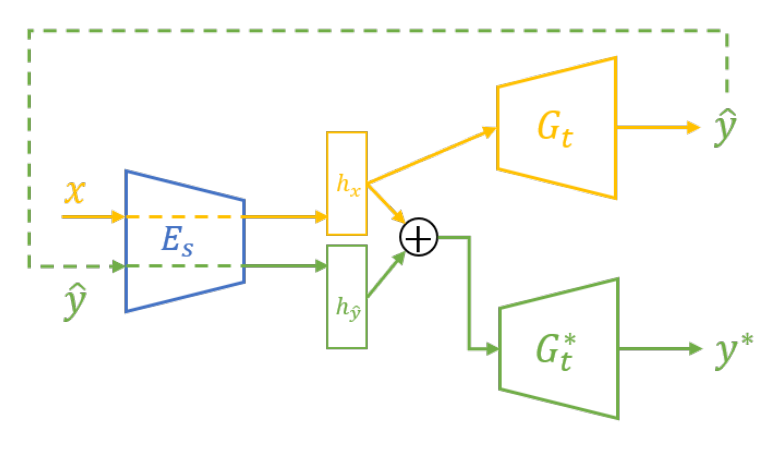}
    \caption{The architecture of the model proposed in \cite{he2019deliberation}. }
    \label{fig:arch:he2019deliberation}
\end{figure}

He et al. \cite{he2019deliberation} adopt the general framework of image-to-image translation models to carry out face manipulations as well as other image translation tasks. As usual, the input image  \(x\) is first encoded to  \(h_x=E_s(x)\) and then a generator is used to generate the target image  \(\hat{y}=G_t(h_x)\). The only difference of the proposed scheme is the introduction of a deliberation network \(G^*_t\)  inspired by the success of the same network in natural language processing tasks. 
As shown in Figure \ref{fig:arch:he2019deliberation}, the initial image generated in the target domain (\(\hat{y}\)) is encoded with the same network that processes the input image and the feature vectors of both images are fed into the deliberation network to give the final image as  \(y^*=G^*_t(h_x, h_{\hat{y}})\). The cycle and adversarial losses govern the training of the model, and it is shown that the model can be trained for multi-modal transformations following a StarGAN's architecture.

There are other works in this category that do not propose new architectures but provide novel ideas for generating training data or developing training algorithms of the existing methods. 
For example, \cite{viazovetskyi2020stylegan2} uses the pix2pixHD framework, but generates paired data for gender and age transformation. It uses styleGAN2 as a face generator and a pretrained network \(f\) for attribute prediction and generates paired data as follows. Firstly, it generates abundant face images using styleGAN2 and predicts the attributes for each image using \(f\) which returns the confidence of the face detection, too. Images with low confidence are filtered out. Then, the center of each class (male or female class for example) and the transition vector between two classes in the latent space are determined. In the following, it generates a random latent vector and moves it along the transition vector direction to generate five latent vectors and their corresponding images. Again, the attributes of the newly generated images are predicted and low-confidence images are filtered. Based on the classification results, the most proper pairs of images for each class are selected for training the transformer. 
Another synthetic dataset creation method is proposed in \cite{Kang2021eyeglass} for eyeglass removal. To do so, they add glasses to face images using either the available facial attribute editing models like starGAN \cite{choi2018stargan} or the landmarks of the input face image. They also consider the age, race, and gender of the input face in selecting the type of the glasses that are added to the photos. 

\begin{figure}[htb]
    \centering
    \includegraphics[width=0.4\textwidth]{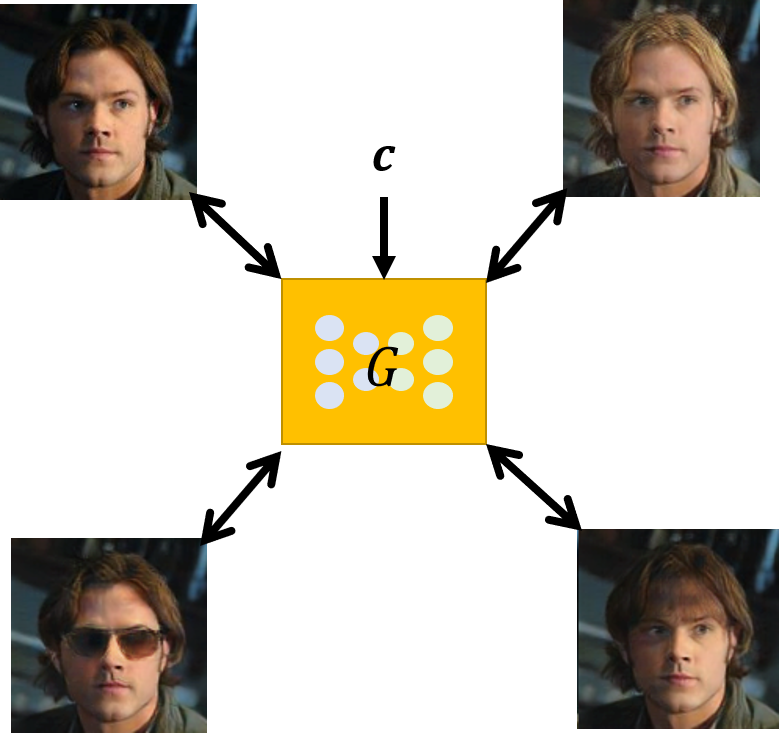}
    \caption{The overall architecture of multi-modal image-to-image translation models for facial attribute editing. }
    \label{fig:arch:multimodal}
\end{figure}
\textbf{Multi-modal image-to-image translation models:}
Bi-modal models are intrinsically designed to transform images of two modalities differing only in a single attribute and they need to be retrained for every different attribute. In multi-modal image-to-image translation architecture illustrated in Figure \ref{fig:arch:multimodal}, there are several domains each related to a specific attribute and a generator that converts the input image from any input domain to an image in a target domain specified by the domain label \(c\). In the following, some models of this group are studied.

\begin{figure}[htb]
    \centering
    \includegraphics[width=0.48\textwidth]{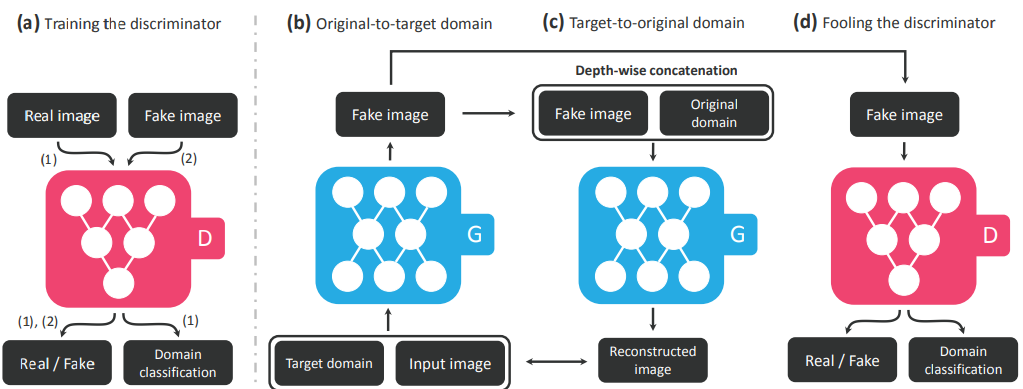}
    \caption{The StarGAN's architecture \cite{choi2018stargan}.}
    \label{fig:arch:stargan}
\end{figure}

StarGAN \cite{choi2018stargan} is a representative example of this category of image-to-image translation models whose architecture is shown in Figure \ref{fig:arch:stargan}.
StarGAN is a framework for multi-domain image-to-image translation which uses a single generator \(G\) that maps the input image from the source domain \(x\) to a target domain \(y\) conditioned on domain label \(c\) as \(y=G(x,c)\). The domain label \(c\) is peeked randomly during training, hence the generator will flexibly translate the input image to multiple domains. The domain label is spatially replicated and concatenated with the input image. The key advantage of starGAN is that it can be simultaneously trained on multiple datasets with different annotations. To this end, a mask vector \(m\) is added to the domain label of each domain that covers unrelated features to ignore unspecified labels and focus on labels available in each dataset.
In addition to adversarial and cycle-consistency losses, an auxiliary classifier is added on top of \(D\) and a domain classification loss contributes in both generator and discriminator to satisfy conditional training and to generate images on the determined domain.

Motivated by StarGAN, SemiStartGAN \cite{hsu2018semistargan} was proposed to tackle the lack of supervision problem. While StarGAN needs a lot of labeled images in each domain to train its multi-domain model, SemiStarGAN establishes a semi-supervised GAN network to train a multi-domain translation model with partially labeled data.
Kapania et al. \cite{kapania2019multiple} proposed another extension of StarGAN by introducing an auxiliary SVM classifier which classifies the generated images and produces a domain label whose difference with the true label of the target domain provides a domain classification loss for more effective training of the generator.

\begin{figure}[htb]
    \centering
    \includegraphics[width=0.48\textwidth]{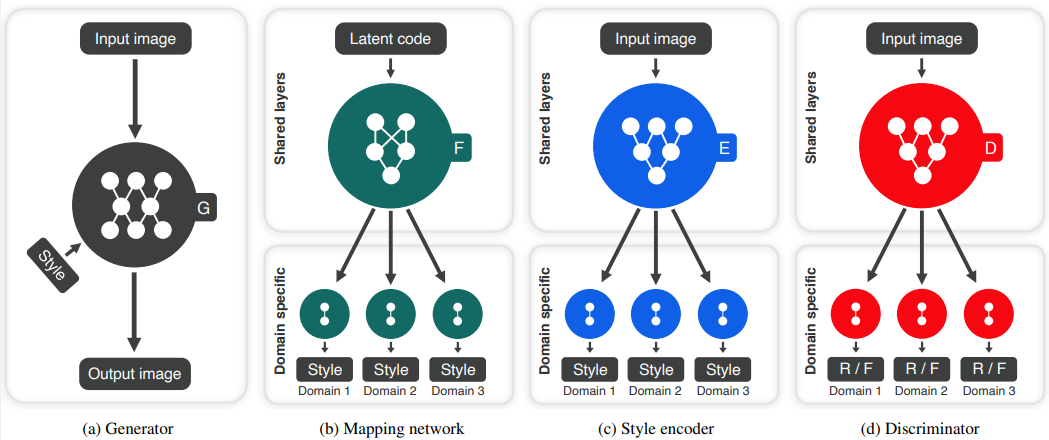}
    \caption{The starGANv2's architecture \cite{choi2020stargan}.}
    \label{fig:arch:stargan2}
\end{figure}

starGANv2 \cite{choi2020stargan} is a general multi-modal image-to-image translation method, specially adopted for face-to-face transformation, which tries to address two necessary properties of such models, i.e. diversity of the generated images and scalability over multiple domains. The model consists of four modules (Figure \ref{fig:arch:stargan2}). The generator \(G\) transforms the input image to an output image based on a style code that determines both the target domain and the style of the image in that domain. The style code is provided either by the mapping network (\(F\)) or by the style encoder (\(E\)). The mapping
network is an MLP network with output branches for each domain. Given a random vector and a target domain, the mapping network generates a random style code. The output of the style encoder is also a domain-specific style code which is extracted from an input image and enables the model to use reference images for performing style transfer. Finally, the discriminator determines if its input image is real or fake, again considering the specified domain of the input image. In addition to cycle consistency and adversarial losses, a style reconstruction loss is used to force the generator to generate images with desired style. Moreover, a diversity sensitive loss is employed to further increase the diversity of the images produced by the generator. 

\subsubsection{Photo-Guided} \label{photo-guided} 

\begin{figure}[htb]
    \centering
    \includegraphics[width=0.48\textwidth]{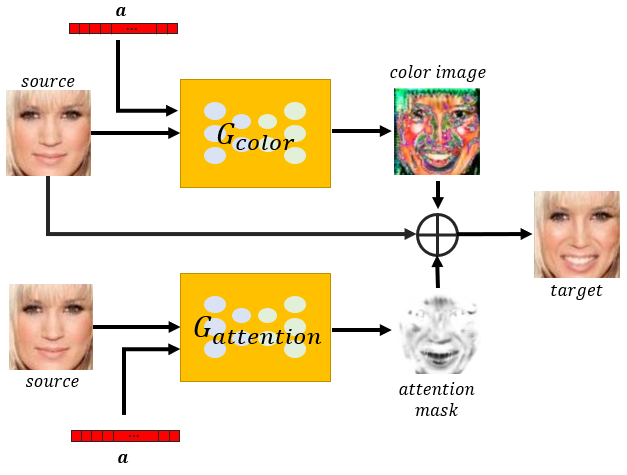}
    \caption{The overall architecture of the mask guided face editing models.}
    \label{fig:arch:maskguidedoverall}
\end{figure}

\textbf{Mask-guided models:}
Figure \ref{fig:arch:maskguidedoverall} demonstrates the generic architecture of the models of this category. The key ingredient of the architectures of these models is the use of a mask to guide the generator \(G\) to focus only on the regions of the input image that may be affected by the edit request and leave the remaining parts untouched. To this end, two generators \(G_{color}\) and \(G_{attention}\) are trained simultaneously, the former is responsible for editing the input image in accordance with the intended attributes and generating the corresponding color image \(I_c\) and the latter predicts an attention mask \(A\) determining the contributions of the edited and original images in synthesizing each pixel of the output image. Finally, the alpha blending operation is performed on the input and edited images under the supervision of the attention mask to generate the target image as \[target = A.I_c+(1-A).I .\]  This way, the generator doesn't require to produce static parts of the input image and can focus particularly on the attribute-specific regions to form more realistic changes.

\begin{figure}[htb]
    \centering
    \includegraphics[width=0.48\textwidth]{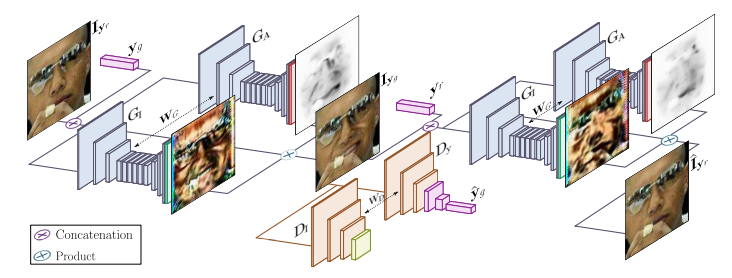}
    \caption{ The GANimation model \cite{pumarola2018ganimation}.}
    \label{fig:arch:ganimation}
\end{figure}

GANimation \cite{pumarola2018ganimation} is a well-known example of the mask-guided category whose architecture is given in Figure \ref{fig:arch:ganimation}. The goal of GANimation is to learn a mapping function that translates input image \(I_{y_r}\) into target image \(I_{y_g}\) conditioned on an arbitrary action unit \(y_g\) representing the goal expression. The input image and spatially replicated vector \(y_g\) are concatenated and fed to a generator consisting of two networks with shared weights in the encoding part. The generator produces two masks instead of a full image: \(G_I\) is responsible for generating a color mask \(C\) which demonstrates changes in input image focusing on regions that form the desired expression, and \(G_A\) is utilized to generate an attention mask \(A\). Each pixel of output image \(I_{y_g}\) is taken from either \(C\) or \(I_{y_r}\) based on the value of the attention mask at that pixel. 
The rest components of the model are used for training generator based on three different losses. First, PatchGan \cite{isola2017image} is adopted to form \(D_I\) that takes generated image as input and results in a matrix demonstrating the probability of each patch coming from a real image. Second, an auxiliary regression component is added to top of \(D_y\) to estimate the activation units of the output image which should be close to \(y_g\). Finally, the same generator is applied to the manipulated image concatenated with the expression vector of the input image \(y_r\) to reproduce the initial image as  \(\hat{I}_{y_r}\) and calculate the reconstruction error.

\begin{figure}[htb]
    \centering
    \includegraphics[width=0.48\textwidth]{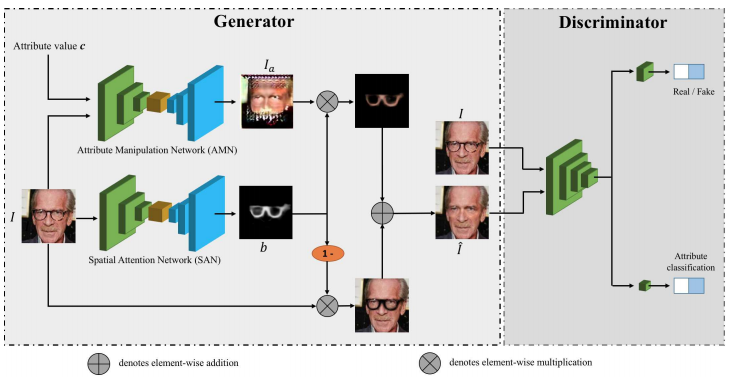}
    \caption{The proposed model of SaGAN \cite{zhang2018generative}.}
    \label{fig:arch:sagan}
\end{figure}

SaGAN \cite{zhang2018generative} is another instance of mask-guided models whose architecture (Figure \ref{fig:arch:sagan}) closely resembles the overall architecture of this category (Figure \ref{fig:arch:maskguidedoverall}). The generator comprises two modules: Attribute Manipulation Network (AMN) and Spatial Attention Network (SAN). AMN focuses on editing the input image based on the attribute value \(c\) and generates an edited face image \(I_a\). SAN takes the input image and predicts attention mask \(b\) that specifies the region in which \(c\) plays a role. \(I_a\) and \(c\) are fused together based on \(b\) to form the final edited image \(\hat{I}\).
Discriminator \(D\) has two roles: discriminating real and fake images which results in generating realistic images and classifying attributes of the edited images to make sure that the output image contains attribute \(c\). In addition, a reconstruction loss is employed to preserve attribute-irrelevant regions and make sure that the identity is not modified. 

The look-globally-age-locally model \cite{zhu2020look} is another typical mask-guided model used for aging faces based on a one-hot attribute vector determining the age range group. In addition to the adversarial and age classification losses, it employs an attention loss to prevent the saturation of the attention mask to zero.

\begin{figure}[htb]
    \centering
    \includegraphics[width=0.48\textwidth]{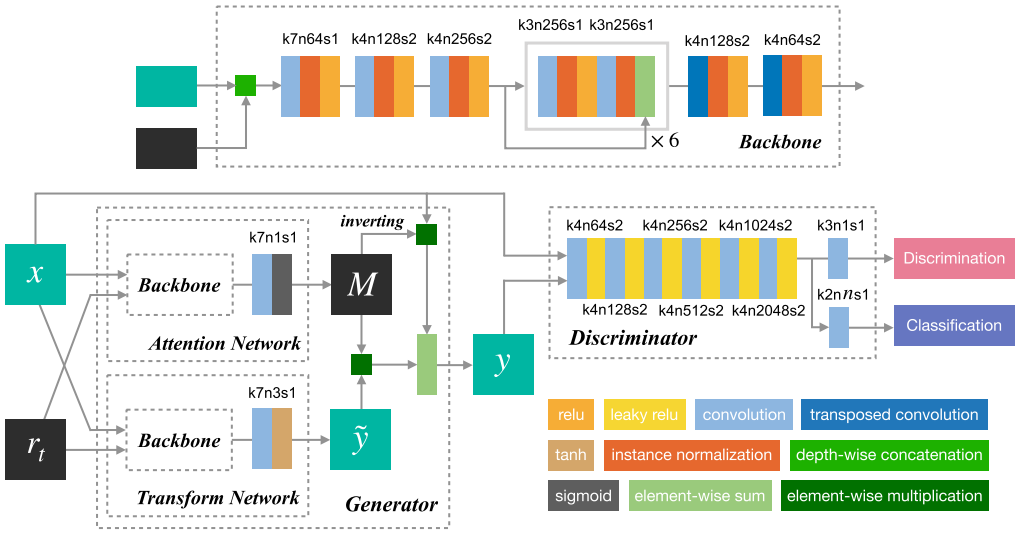}
    \caption{The RAG model's architecture \cite{zhang2019rag}.}
    \label{fig:arch:rag}
\end{figure}

RAG \cite{zhang2019rag} is also a mask-guided model that like the other models in this category, has a generator with two modules: the Transform Network (TN)  that generates an edited image \(\hat{y}\) and the Attention Network (AN) results in an attention mask \(M\) (Figure \ref{fig:arch:rag}).
The only difference with the base model of this category is that RAG \cite{zhang2019rag} uses a residual attribute vector \(r_t\) instead of directly feeding desired attribute vector \(c_t\) to the generator. The residual vector is defined as the difference between \(c_t\) and the attribute vector of the source image \(c_s\) and describes what should be changed in the source image to generate the target image. At the inference time, using residual vectors is more convenient than the target vectors since \(c_t\) is unknown. The fusion of the source image and the generated color image is the same as before and again adversarial, classification, identity and reconstruction losses are used to train the model.

\begin{figure}[htb]
    \centering
    \includegraphics[width=0.48\textwidth]{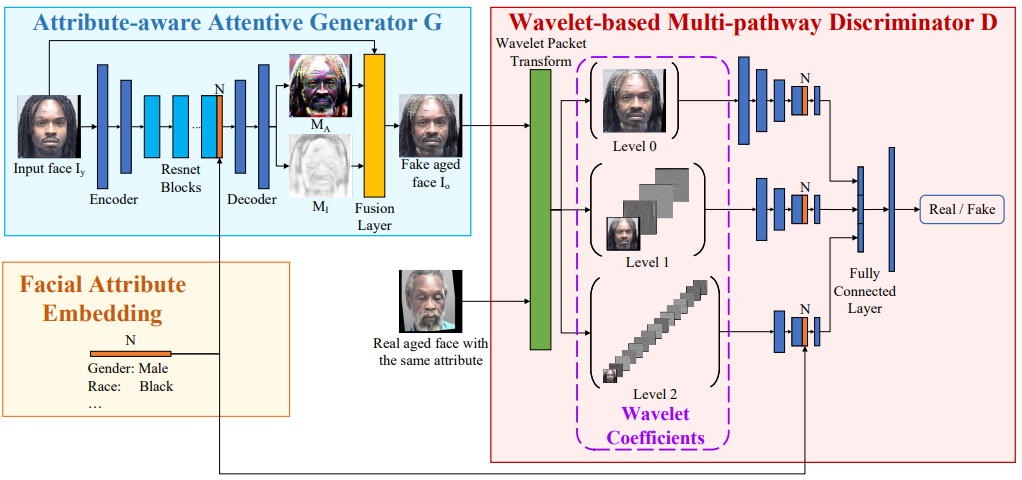}
    \caption{ The A\textsuperscript{3}GAN model \cite{liu20213}. }
    \label{fig:arch:a3gan}
\end{figure}

A\textsuperscript{3}GAN \cite{liu20213} is a purpose-built generative model for face aging. The model consists of two main components (Figure \ref{fig:arch:a3gan}): an attribute-aware attentive generator and a wavelet-based multi-pathway discriminator. The generator is a common mask-guided generator, except that it does not require an attribute vector for determining the required edits since the model is only used for increasing the age of the input face. Conversely, a facial attribute embedding is used to describe the features of the input face, e.g. gender and race, that should be preserved after aging- the reason why the model is called attribute-aware. As shown in Figure \ref{fig:arch:a3gan}, this facial attribute embedding which should be provided as conditional information, is concatenated to the latent representation of the input face in an hourglass-shaped fully convolutional network that outputs the attention mask and the image map. 
The discriminator employs wavelet coefficients of different scales of the input image as a representation of the textural features of the input face which concatenated with the facial attribute vector is used for deciding whether the image is real or fake. 

\begin{figure}[htb]
    \centering
    \includegraphics[width=0.45\textwidth]{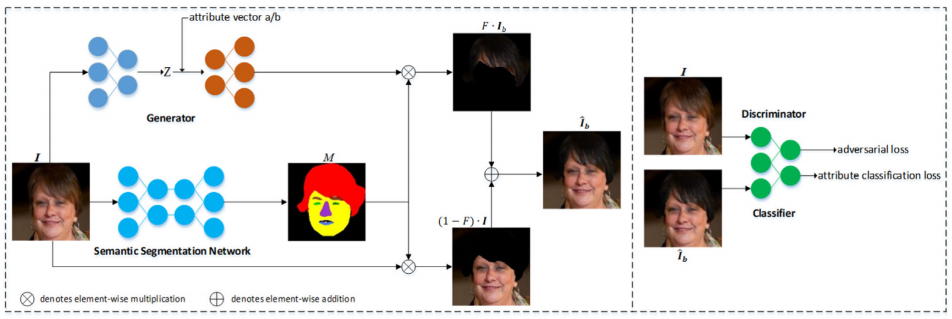}
    \caption{The structure of SM-GAN model \cite{chen2020harnessing}.}
    \label{fig:arch:smgan}
\end{figure}

Figure \ref{fig:arch:smgan} presents the overall structure of SM-GAN \cite{chen2020harnessing}. In SM-GAN a typical encoder-decoder generator is utilized to generate an edited image \(I_b\) according to the target attribute vector \(b\). However, the attention mask generation is different from the previous models. Here, a semantic segmentation network is trained and employed to provide seven semantic segmentation masks for different parts of a face. Each facial attribute is related to one or more of these regions by a set of human-defined rules. Therefore, each attribute \(j\) corresponds to one or more of segmented masks and a target region mask is formed for each attribute as the union of all corresponding segmented masks. As the model is able to edit multiple-attributes, the final target region mask is formulated as \(F=F_1 \cup F_2 \cup ... \cup F_k\) where \(k\) is the number of attributes to change. Final region mask \(F\) is used to merge the input image \(I\) and the edited image \(I_b\).

\begin{figure}[htb]
    \centering
    \includegraphics[width=0.48\textwidth]{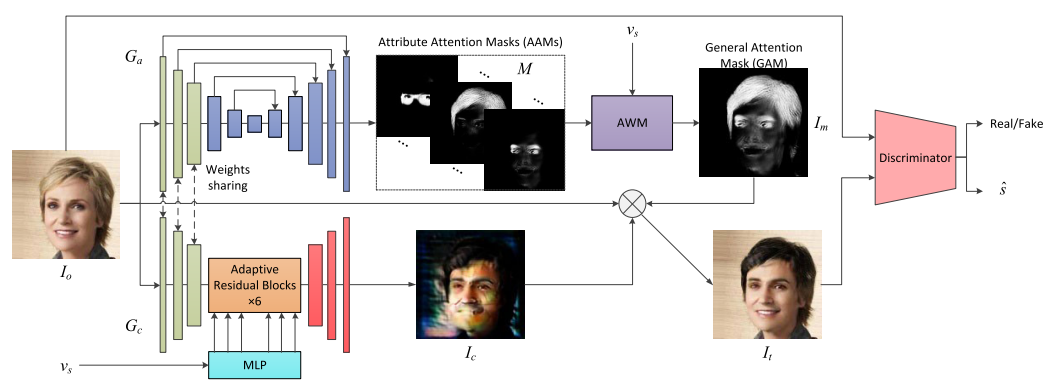}
    \caption{The self-attention-masking model's architecture \cite{xia2020self}.}
    \label{fig:arch:selfattention}
\end{figure}

Self-attention-masking model \cite{xia2020self} is also based on the idea of semantic segmentation and decomposition of the face region. As shown in Figure \ref{fig:arch:selfattention}, the generator of this model consists of two parts: \(G_c\) which generates a modified image as color mask and \(G_a\) which is responsible to produce an attention mask. \(G_a\) is trained to give M Attribute Attention Masks (AAMs) for different facial attributes like hair, skin, sunglasses, gender, and age.  An Attention Weighting Module (AWM) is then used to combine AAMs and form a general attention mask (GAM) based on the attribute vector \(v_s\). Absolute values of \(v_s\) are used to update AAMs and determine the strength of activation of each attribute mask. The final image is then generated based on the input image, the general attention mask and the edited attribute image \(I_c\).

\begin{figure}[htb]
    \centering
    \includegraphics[width=0.48\textwidth]{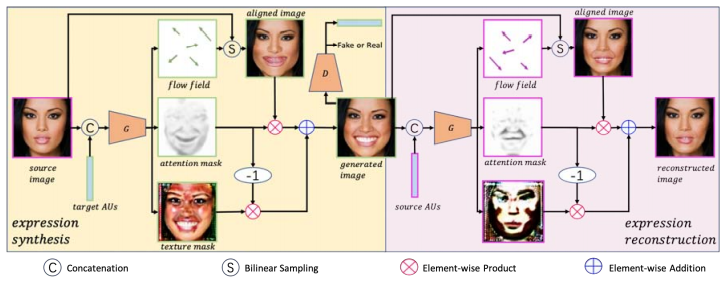}
    \caption{The DFT-Net's architecture \cite{Wang2019dft}.}
    \label{fig:arch:dft}
\end{figure}

DFT-Net \cite{Wang2019dft} (Figure \ref{fig:arch:dft}) follows the overall architecture of this group for manipulating expression in the given input face image. \(G\) is a generator composed of one encoder and two decoder networks. The source image \(X_s\) and target attributes in the form of the action units (\(AU\)s vector) concatenation is encoded to a latent space representation \(C\). 
The first decoder generates the flow field map and the attention mask and the second decoder produces the texture mask. The attention mask and texture masks are the same as attention mask and color image of the general architecture of mask-guided models but the flow field is the new component proposed in DFT-Net for warping the input image. Using a bilinear sampling operation, the aligned image is produced from the input image with the guidance of the extracted flow field and participates in the final merging operation instead of the source image. 
As before, a discriminator is employed for applying adversarial and conditional losses while the reconstruction phase is used to increase the similarity of the source and generated images and improve identity preservation through reconstruction loss. Moreover, total variation (TV) loss and attention regularization loss are deployed to make the flow field smooth and to prevent it from saturating to zero and nullifying the flow decoder.


\begin{figure}[htb]
    \centering
    \includegraphics[width=0.48\textwidth]{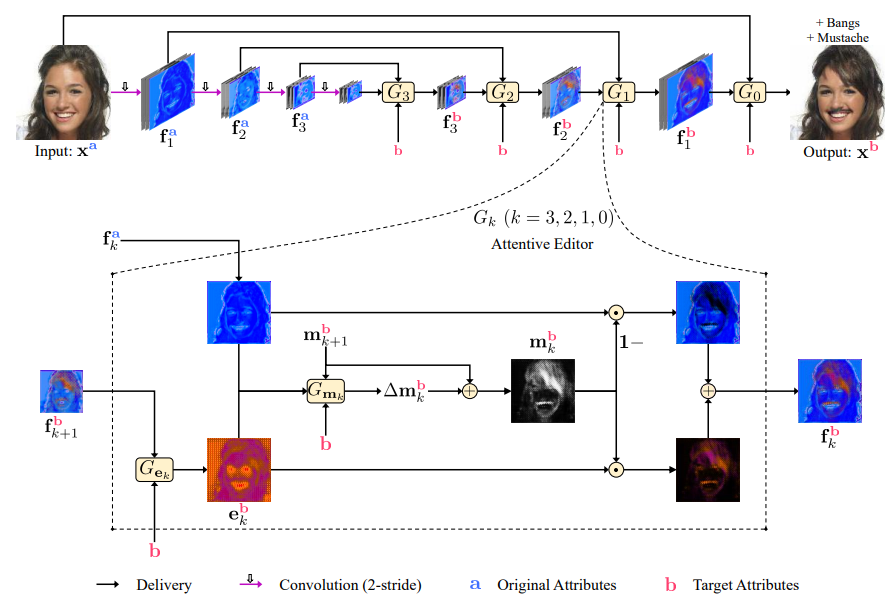}
    \caption{PA-GAN's architecture \cite{he2020pa}.}
    \label{fig:arch:pagan}
\end{figure}

PA-GAN \cite{he2020pa} proposes a solution for progressive facial attribute editing from high to low level features while at each level only the attribute area is modified. As shown in Figure \ref{fig:arch:pagan}, the model has four attentive editors \(G_k\) which accomplish editing in feature space. Each editor has a mask-guided pipeline; \(e_k^b\), generated by \(G_{e_k}\), is a feature map that contains the information of the target image with desired attributes and \(m_k^b\) corresponds to a typical attention map and is generated by \(G_{m_k}\) via a residual strategy. \(e_k^b\) is an improvement over the previous level's resulted map \(f_{k+1}^b\) and represents more precise target attributes. On the other hand, \(G_{m_k}\) is responsible for generating residual \(\Delta m_k^b\) and makes the attention mask more surpassing as \(m_k^b=m_{k+1}^b+\Delta m_k^b\). The prime attention mask is course because of global information gathered by highest feature level and as the feature level gets lower, information becomes more local and the attention mask becomes more precise. As in mask-guided baseline, input feature map \(f_k^a\) and refined feature map \(e_k^b\) are merged together with the guidance of attention mask \(m_k^b\) and the output feature map \(f_k^b\) is generated as robust editing result of this level.
In addition to attribute prediction and adversarial losses, two extra loss functions are introduced for learning a more accurate attention mask. \(l_{spa}\) constrains the sparsity of the attention mask forcing the mask to concentrate on the proper regions of the image instead of covering the whole image. Moreover, as some attributes have disjoint regions and their masks should have no overlap, \(l_{ovl}\) minimizes pixel-wise multiplication of those masks in order to keep their overlap minimum.

Local Mask-based Image-to-image Translation (LOMIT) \cite{cho2019and} exploits the idea of mask-guided editing to transfer style from a driving to a source image. To do so, LOMIT first generates two masks for the driving and source images based on the target attribute, e.g. the hair mask for the hair color attribute. Both images are then separated into the foreground and background based on the corresponding attention masks. The source image along with its attention mask and the separated background are then combined with the foreground of the driving image to perform the style translation from the driving to the source image in a special encode-decoder network, called highway AdaIN. In order to learn semantic consistent masks that capture the whole region of the attribute, i.e. consider entire hair region as a whole even under various lightening conditions, a regularization term minimizes the difference between mask values of those pixels. The reconstruction, adversarial, and classification losses are also used in the training process.

\begin{figure}[htb]
    \centering
    \includegraphics[width=0.48\textwidth]{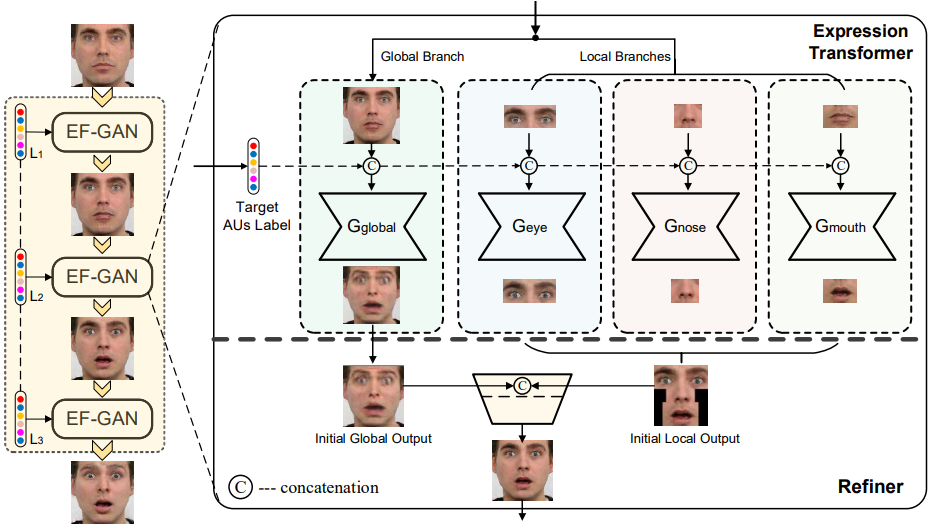}
    \caption{The pipeline of cascadeFE-GAN model \cite{wu2020cascade}.}
    \label{fig:arch:cascade}
\end{figure}

Wu et al. propose a cascade FE-GAN \cite{wu2020cascade} as a progressive model for handling extreme facial expression manipulations. The main pipeline of this model (Figure \ref{fig:arch:cascade}) consists of three consecutive EF-GANs each performing one step of expression editing on the output of the previous EF-GAN. In this model, the target face expression is given as Action Units (AUs) vector. Each EF-GAN receives an initial image and an AUs vector.  In addition to a global branch that produces the edited version of the whole face, three local branches are employed in EF-GANs to especially concentrate on the changes that should happen to the mouth, nose and eyes regions of the face. Each local and global branch of EF-GAN is a mask-guided model; it first computes an edited color image \(M_C\) and an attention map \(M_A\) and then fuses the input image and the edited image based on the attention map as  \(\mathcal{I}_{init}=M_A\cdot M_C+(1-M_A)\cdot I_{in}\). 


Initial outputs of local branches are then assembled and synthesized as a complete face. Two faces acquired from two types of branches are then concatenated and fed into a refiner network that generates the final edited image.

\begin{figure}[htb]
    \centering
    \includegraphics[width=0.48\textwidth]{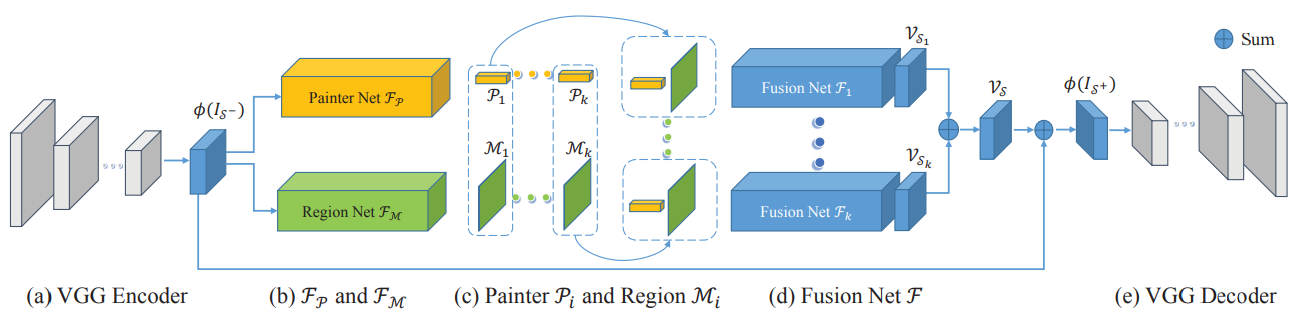}
    \caption{The semantic component decomposition model \cite{chen2019semantic}.}
    \label{fig:arch:scd}
\end{figure}

Semantic Component Decomposition \cite{chen2019semantic} leverages the idea of mask-guided models at the level of face feature maps instead of the images. To provide more user control over the edit process, the facial attributes are decomposed into multiple semantic attributes and a specific region is associated to each attribute. The complete pipeline of the model is illustrated in Figure \ref{fig:arch:scd}. 
\(I_{\mathcal{S}^-}\) is the input image without a particular attribute \(S\) and \(I_{\mathcal{S}^+}\) is the resultant face with attribute \(S\).
Firstly, negative sample \(I_{\mathcal{S}^-}\) is encoded in the feature map via a pretrained VGG encoder as \(\phi(I_{\mathcal{S}^-})\). The aim is to compute attribute tensor \(\mathcal{V}_\mathcal{S}\) that added to \(\phi(I_{\mathcal{S}^-})\) produces \(\phi(I_{\mathcal{S}^+})\), the feature map of the target face, which is then converted to the output image \(I_{\mathcal{S}^+}\) via a VGG decoder. To do so, \(\phi(I_{\mathcal{S}^-})\) passes through the painter network \(\mathcal{F}_{\mathcal{P}}(.)\) and the region network \(\mathcal{F}_{\mathcal{M}}(.)\) that correspond to the color network and the attention network of the mask-guided models, respectively. Each \(\mathcal{P}_i\) determines a specific effect whose corresponding region is given by \(\mathcal{M}_i\). In the following, every pair of \(\mathcal{P}_i\) and \(\mathcal{M}_i\) is fused under Fusion Network \(\mathcal{F}_i(.)\) resulting \(\mathcal{V}_{\mathcal{S}_i}\). Finally, the high-level attribute tensor \(\mathcal{V}_\mathcal{S}\) is calculated as a linear combination of all \(\mathcal{V}_{\mathcal{S}_i}\) components.

\begin{figure}[htb]
    \centering
    \includegraphics[width=0.48\textwidth]{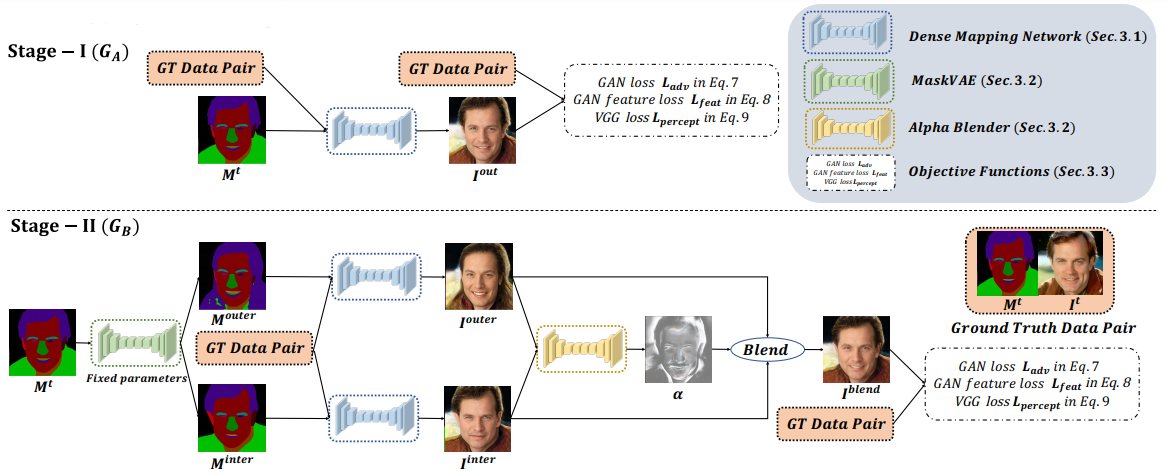}
    \caption{The structure of MaskGAN \cite{lee2020maskgan}.}
    \label{fig:arch:maskgan}
\end{figure}

MaskGAN\cite{lee2020maskgan} is another mask-based method for interactive facial image manipulation which can be utilized in other applications like transferring style from a source to a target image. However, it uses semantic masks of 19 facial component labels, e.g. eye, mouth, skin, and nose.
The model comprises three components: Dense Mapping Network (DMN), MaskVAE, and AlphaBlender. DMN (Figure \ref{fig:arch:maskgan}), as the only component used at the inference time, extends pix2pixHD \cite{wang2018high} architecture with an extra style encoder \(Enc_{style}\). \(Enc_{style}\) extracts style information from the input image \(I^t\) using the extra spatial information provided by the corresponding semantic mask \(M^t\) and generates spatial-aware style information which are injected into the residual blocks of DMN via AdaIN. To train the model, a large-scale high resolution dataset of face images and semantic maps, called CelebAMask-HQ, has been created. Moreover, MaskVAE and AlphaBlender components are utilized to simulate the editing behaviors of the users to train the model with diverse semantic masks. MaskGAN has been successfully used for style transfer.

\begin{figure}[htb]
    \centering
    \includegraphics[width=0.48\textwidth]{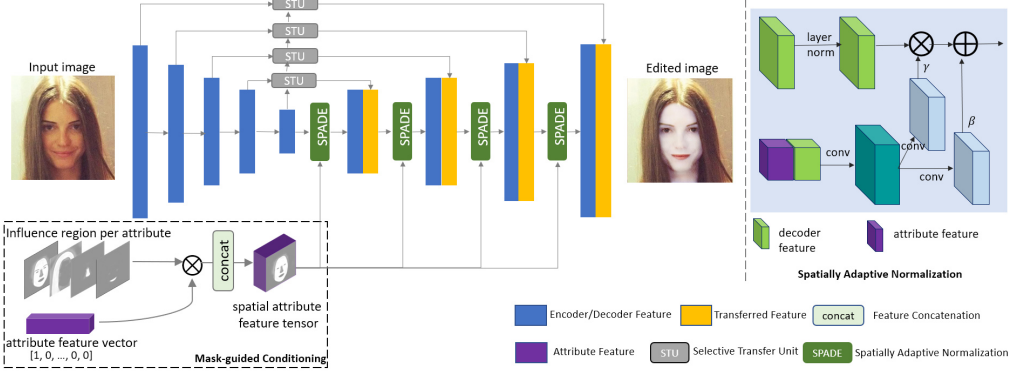}
    \caption{The MagGAN's architecture \cite{wei2020maggan}.}
    \label{fig:arch:maggan}
\end{figure}

Unlike the common mask-guided models which use the attention mask to improve the quality of the generated image by restoring unwanted regions via alpha-blending of the modified and original images, MagGAN \cite{wei2020maggan} directly uses masks in the generator to only alter desired parts from scratch. MagGAN (Figure \ref{fig:arch:maggan}) follows an encoder-decoder architecture using selective transfer units proposed in STGAN \cite{liu2019stgan} and uses adaptive layer normalization to avoid direct concatenation of the condition with the feature map inspired by StyleGAN \cite{karras2019style}.
Firstly, a segmentation model is trained to produce 19 masks from a face image. For each attribute \(a_i\), two probability masks \(M^+\) and \(M^-\) are dedicated which indicate influence regions such that \(M^+\) shows the regions affected by increasing \(a_i\) and \(M^-\) corresponds to the regions altered if \(a_i\) decreases. A Mask-Aware Reconstruction Error (MARE) is proposed to assess the quality of irrelevant region preservation.
From the perspective of how to inject attributes as a condition to the model, traditional methods replicate attribute vector spatially while it conveys no spatial information. But, MagGAN uses masks to localize attribute changes directly into the generator as a \(C\) channel mask-guided attribute condition tensor whose \(i\)th channel has 1s at the positions that may be modified due to the change of \(a_i\).
A series of multi-level discriminators are used to reach high resolution realistic results. The coarsest-level discriminator checks global consistency by getting a full downsampled image as input and finer-level discriminators check how realistic are patches in the high-resolution image. A classifier is also associated with each discriminator.

\begin{figure}[htb]
    \centering
    \includegraphics[width=0.48\textwidth]{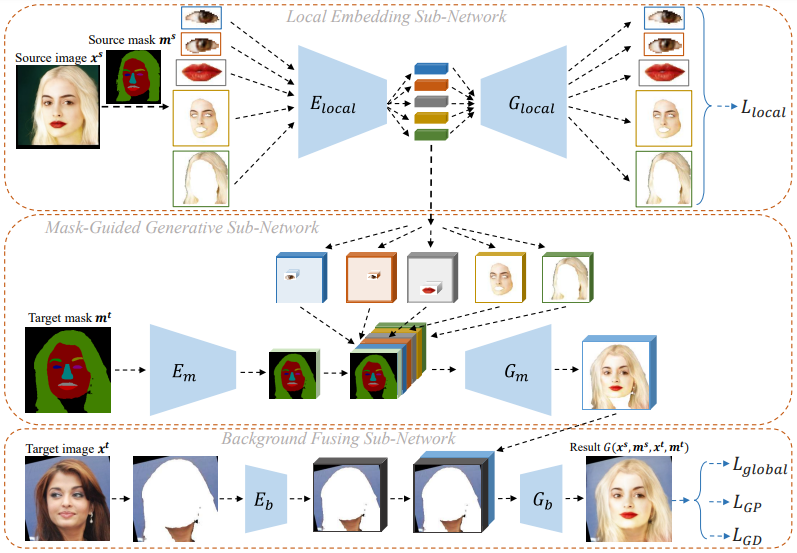}
    \caption{The architecture of the model proposed in \cite{gu2019mask}.}
    \label{fig:arch:maskguided}
\end{figure}

Gu et al. \cite{gu2019mask} propose another face editing model that doesn't completely fit the general architecture of the mask-guided models but as it uses masks to guide the local feature extraction and image generation, we review it as a mask-guided model.
As illustrated in Figure \ref{fig:arch:maskguided}, this model is made up of three sub-modules: a local embedding sub-network that extracts feature embedding of the source image with the guidance of source masks, a mask-guided generator sub-network which uses a target mask to concatenate facial features in order to generate the foreground face, and a background fusing sub-network to fuse generated foreground face and the background to from the target image. In the local embedding sub-network, the source mask \(m^s\) is produced from the source image \(x^s\) using a face parsing module. A distinct auto-encoder network is trained for every five facial components to learn separate embedding for each component. Using embedding at component-level improves model controllability to component-level.
At the mask-guided generation step, the target mask along with five feature tensors of the previous step are given to a generator to produce the foreground image. 
To put the resultant foreground face on the target background in an artifact-free manner, a background fusion sub-network is used. Target background is extracted with the target mask \(m^t\) and is encoded using (\(E_b\)) encoder. The final generator (\(G_b\)) gets the concatenation of foreground face and the background feature tensor and generates the final output.
Adversarial loss and two reconstruction losses at local and global levels are used to train the model. Furthermore, a face parsing loss is employed to keep the mask of the generated image close to the target mask.
 

\begin{figure}[htb]
    \centering
    \includegraphics[width=0.48\textwidth]{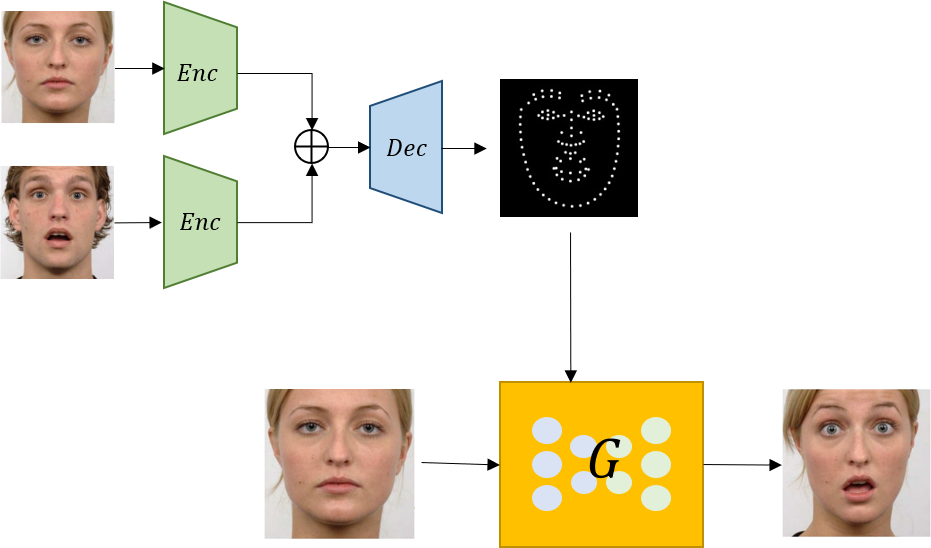}
    \caption{The overall architecture of landmark-guided face editing models.}
    \label{fig:arch:landmark-guided-overall}
\end{figure}
\textbf{Landmark-guided models:}
Figure \ref{fig:arch:landmark-guided-overall} depicts the generic architecture of landmark-guided face editing models. The main characteristic of the models of this category is the use of a landmarks image that is given to the generator \(G\) to be used as a constraint in synthesizing the target face image. These models are usually used for transferring the pose and expression of a driving image to a source image while the identity of the source image is preserved. To this end, there is a first step to extract landmarks of the driving image and intended information of the target image.  A conditional generator is then used to produce an image of the target person with the desired pose (landmarks). Some prominent models of this category are introduced in the following.

\begin{figure*}[htb]
    \centering
    \includegraphics[width=\textwidth]{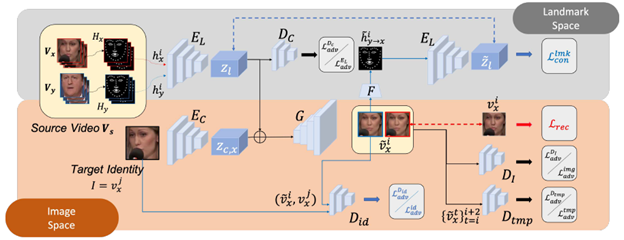}
    \caption{CrossID-GAN's architecture \cite{huang2020learning}.}
    \label{fig:arch:crossID-GAN}
\end{figure*}

CrossID-GAN \cite{huang2020learning} is a landmark-guided model to reenact a target image with a driving video to generate a moving video of the target face. CrossID-GAN is trainable in both supervised and unsupervised manners. This model completely follows the overall architecture  of landmarks-guided models (Figure \ref{fig:arch:landmark-guided-overall}) and consists of three main components: an encoder \(E_{L}\) that encodes ID-invariant driving facial landmarks into \(Z_{l}\), an encoder \(E_{C}\) that encodes the content of a source image into identity preserving representation \(Z_{c,x}\), and a conditional generator \(G\) that generates the final image.  The data flow between the three aforementioned components is shown in Figure \ref{fig:arch:crossID-GAN}. In supervised training, the training data is a sequence of landmarks extracted from a sequence of driving images (\(V_{x}\)) while the driving images and target image have the same identity and the goal is to reconstruct the input video. For training the model in this manner, the reconstruction and adversarial losses are utilized. In unsupervised training, driving images and the target image are from different identities. A few auxiliary components are used to train CrossID-GAN when there is no ground truth. One of these components is a discriminator \(D_{C}\) which classifies content/ID and is trained with \(E_{L}\) as generator. Another constituent of the model is a pre-trained network \(F\) for landmark extraction. By using \(F\) and incorporating an ID-invariant landmark consistency loss, the model learns to minimize the distance between landmarks of the generated image and those of the driving image. Moreover, three discriminators are employed: \(D_{I}\) to enforce the model to generate realistic images, \(D_{tpm}\) along with a temporal consistency loss to ensure the smoothness of the output reenacted video, and \(D_{id}\) to certify the identity consistency between the generated image and the target image.

\begin{figure}[htb]
    \centering
    \includegraphics[width=0.48\textwidth]{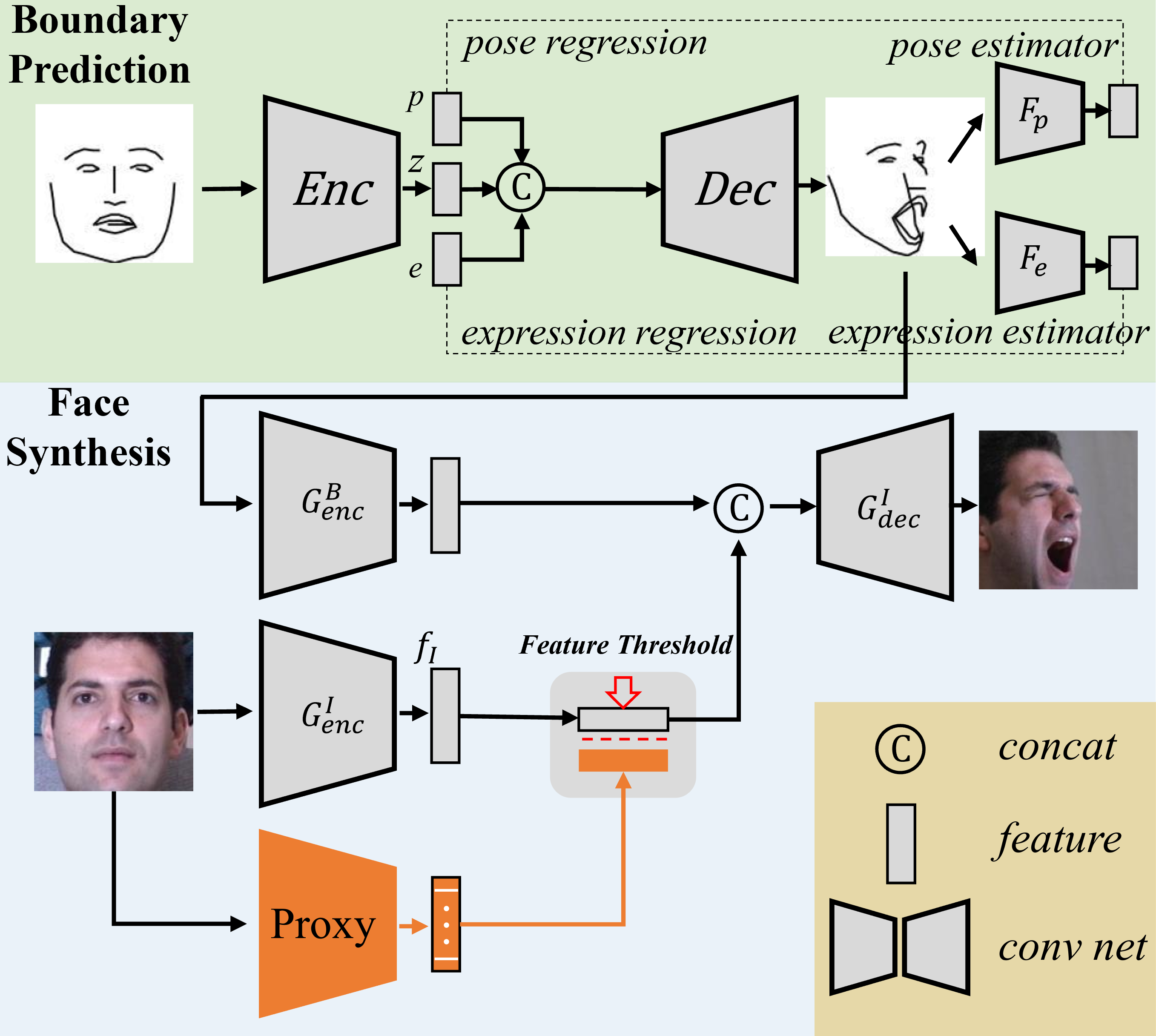}
    \caption{The boundary-based GAN model proposed in \cite{fu2021high}.}
    \label{fig:arch:boundarybased}
\end{figure}

Fu et al. \cite{fu2021high} have introduced a boundary-based model for pose and expression manipulation in which the landmarks image is replaced with a boundary map of the face. As illustrated in Figure \ref{fig:arch:boundarybased}, the model is formed of two correlated stages: boundary prediction and face synthesis. The first stage predicts the boundary of the desired output by using an encoder-decoder architecture. This stage acts as the first part of the overall architecture of landmarks-guided models, but in a more compact way in which the pose and expression vectors of the driving image are directly inserted into the decoder instead of using an encoder for extracting this information. The encoder encodes the input boundary image into \(z\), and the decoder generates the image boundary of the intended image based on a concatenation of \(z\) and the pose and expression vectors. Pixel-wise and conditional regression losses focusing on pose and expression are used in this stage. \(F_{p}\) and \(F_{e}\) in this stage are two pre-trained networks for predicting the pose and expression of the generated boundary image, respectively. The second stage consists of two encoders and one decoder. This stage demonstrates the generator part of the architecture of landmarks-guided models. The boundary encoder encodes the boundary image of the previous stage to \(f_{B}\) vector; meanwhile, the identity encoder encodes the input face image into \(f_{I}\) vector. Finally, the decoder generates the manipulated face image. By incorporation of a pre-trained proxy network and a feature threshold loss, the structure and texture are enforced to be disentangled in the latent space. Moreover, multi-scale pixel-wise, multi-scale conditional adversarial, and identity preserving losses are utilized for training the model. 

\begin{figure*}[htb]
    \centering
    \includegraphics[width=\textwidth]{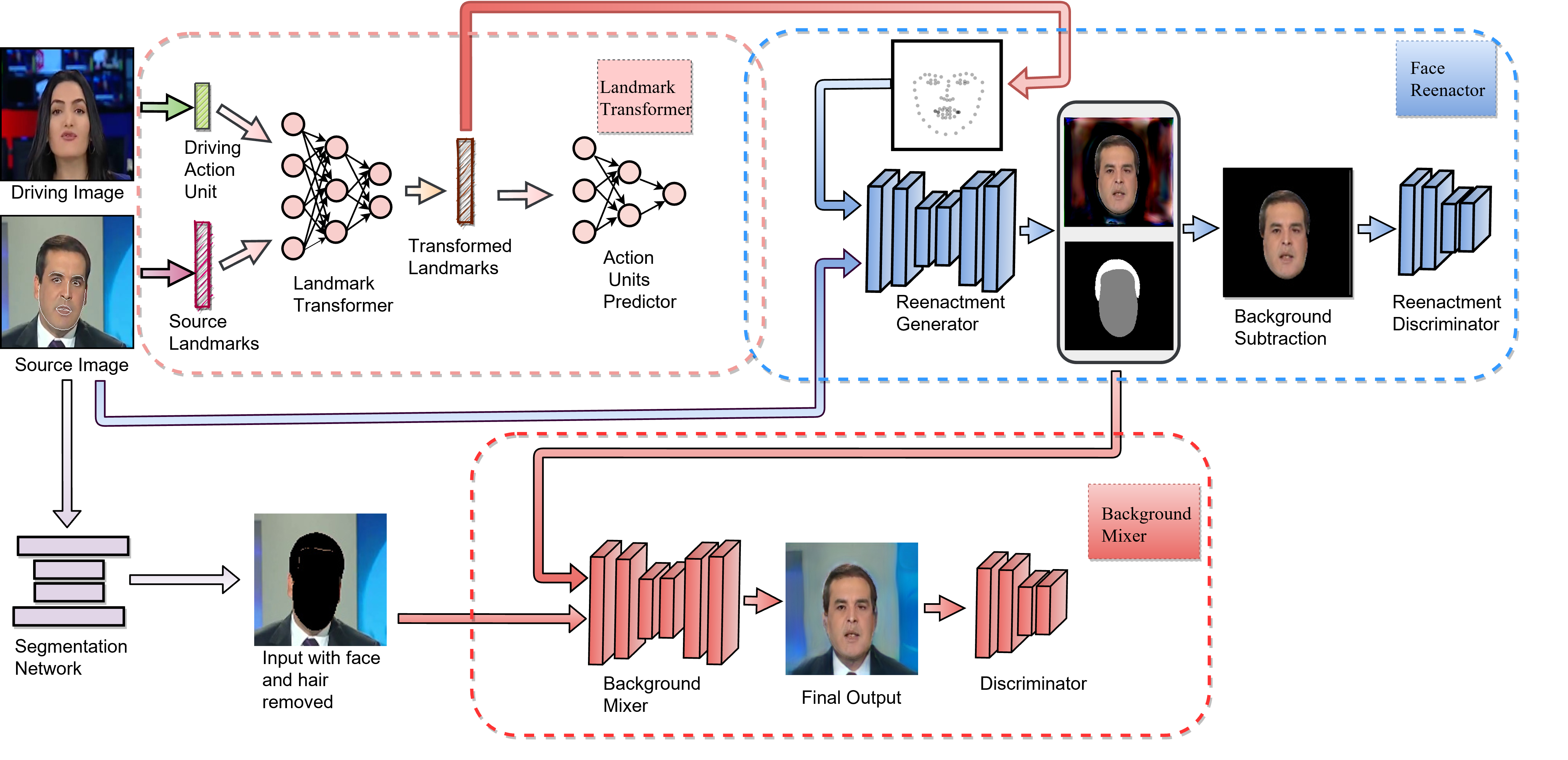}
    \caption{The FACEGAN's architecture \cite{tripathy2021facegan}.}
    \label{fig:arch:facegan}
\end{figure*}

FACEGAN \cite{tripathy2021facegan} is a reenactor model which transfers the expression and pose of a driving image to a target image. As depicted in Figure \ref{fig:arch:facegan}, FACEGAN is composed of three parts: landmark transferring, face reenactment, and background mixing. The two first stages are compatible with the generic architecture of landmarks-guided models, but the third stage is a specific network for improving the quality of the final output. In the landmark transferring step, the driving face's expression and pose   are extracted in the form of action units (AUs) by a pre-trained network. The extracted AUs and the landmarks of the driving face are inputted to a landmark transformer which generates a landmark embedding vector that contains the target face's structure and driving face's AUs. Successive frames of videos are used to train the landmark transformer, which is a fully connected network. Afterward, this generated landmark vector is converted to a one-channel gray-scale image. In the face reenacting step, a reenactment generator is used to produce the reenacted face image and the corresponding segmentation given the gray-scale landmark image and the source image. Finally, in the background mixing step, the background mixer network generates an image that combines the source image's background with the reenacted face. FACEGAN utilizes two discriminators in its training: one discriminator for training reenactment generator and another one to train the background mixer network. In addition, reconstruction loss, connectivity loss, perceptual loss, adversarial loss, standard cross-entropy, and pixel-wise loss are used in the training time.

\begin{figure}[htb]
    \centering
    \includegraphics[width=0.48\textwidth]{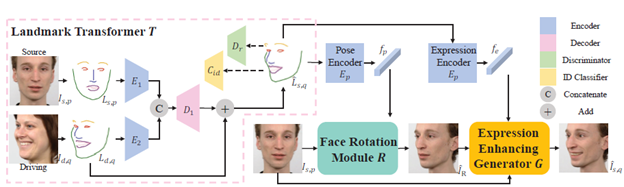}
    \caption{The outline of the LI-Net model \cite{liu2021li}.}
    \label{fig:arch:linet}
\end{figure}

LI-Net \cite{liu2021li} is another model for manipulating face expression and pose under the guidance of landmarks. As depicted in Figure \ref{fig:arch:linet}, LI-Net comprises three constituents: a landmark transformer, a face rotation module, and an expression enhancing generator. The landmark transformer includes two encoders for encoding source and driving landmarks, and a decoder for generating a landmark shift. The landmark shift is added to the driving landmark and makes an output landmark that contains the identity of the source face along with the pose and expression of the driving face. A discriminator and an identity classifier are utilized to guarantee the realness and identity consistency of the generated landmarks, respectively. Moreover, a cycle loss is employed to ensure that the transformed landmarks are equal to the source landmarks in an inverted path. The rotation module generates a rotated face under the guidance of the pose vector extracted from the generated landmarks. The last component applies the expression of driving face on the rotated face and generates the ultimate output. A pose prediction and a perceptual loss are used to maximize the accuracy of generated pose and minimize the gap between images based on their features. 

\begin{figure}[htb]
    \centering
    \includegraphics[width=0.48\textwidth]{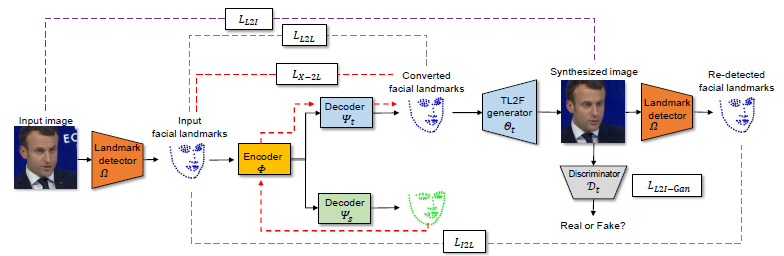}
    \caption{The LandmarkGAN's structure \cite{sun2020landmarkgan}.}
    \label{fig:arch:landmarkgan}
\end{figure}

LandmarkGAN \cite{sun2020landmarkgan} is a subject-specific model. As illustrated in Figure \ref{fig:arch:landmarkgan}, instead of using driving and source images for reenactment, LandmarkGAN utilizes a target-specific landmark to face generator (TL2F) and a target-specific decoder (shown in blue color in Fig \ref{fig:arch:landmarkgan}). In other words, the identity of the target person has been embedded in the weights of TL2F generator and the decoder network, and the driving image has been replaced with the input facial landmarks. So, in practice, LandmarkGAN's architecture conforms to the base architecture of this category except that the source image is not received as an input and is learned by the model. The landmark converter component of LandmarkGAN has an encoder-decoder structure and converts input facial landmarks to a vector. The encoder part is shared among all identities while as mentioned earlier, the decoder is target-specific. The TL2F generates the reenacted face with the target's identity. In addition to adversarial and reconstruction losses, a cycle loss is used in training the model.

\begin{figure}[htb]
    \centering
    \includegraphics[width=0.48\textwidth]{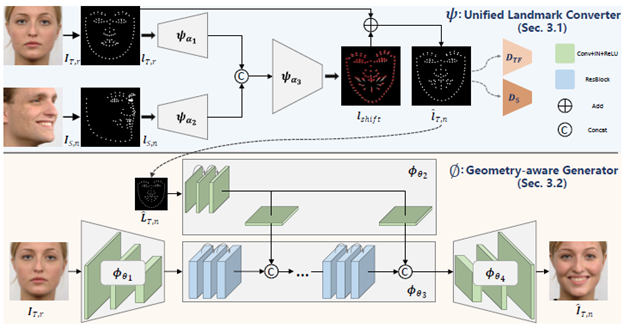}
    \caption{The FReeNet reace reenactment model \cite{Zhang2020freenet}.}
    \label{fig:arch:freenet}
\end{figure}

FReeNet \cite{Zhang2020freenet} is a landmark-guided model to manipulate facial expressions. As depicted in Figure \ref{fig:arch:freenet}, FReeNet comprises two stages: Unified Landmark Converter (ULC) and Geometry-aware Generator (GAG). ULC stage consists of one generator and two discriminators. The generator part leverages two encoders and one decoder to generate reenacted landmarks. One encoder is responsible for extracting the driving (source) face's landmarks, while the other is for extracting the target face's landmarks. The decoder fuses the outputs of the two encoders to generate the landmarks shift (\(I_{shift}\)). By adding landmarks shift to the driving landmarks the desired landmarks are obtained which contains the identity of the target face and the expression of the driving face. There are two discriminators in this part of the model: \(D_{S}\) by focusing on identity similarity between landmarks pair and \(D_{TF}\) by estimating if a set of landmarks is real or fake make the generator more efficient in accuracy and robustness. ULC uses point-wise \(L1\), cycle consistency, and adversarial losses in training time. GAG part of the model consists of two encoders, a transformer, and a decoder. One encoder for encoding generated landmarks in the previous stage, another for encoding source face image. The transformer decouples appearance and geometry information from encoders' outputs. Finally, the decoder generates desired output. In this stage pixel-wise \(L1\), adversarial, and triplet perceptual losses are used for training. 

\begin{figure}[htb]
    \centering
    \includegraphics[width=0.48\textwidth]{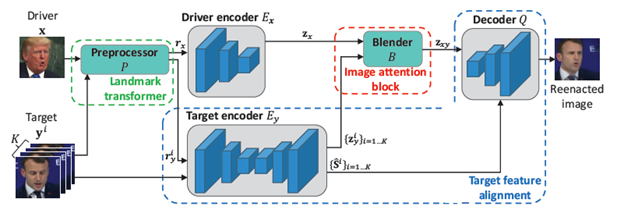}
    \caption{The architecture of MarioNETte reenactment model \cite{ha2020marionette}.}
    \label{fig:arch:MarioNETte}
\end{figure}

MarioNETte \cite{ha2020marionette} is a reenactment model working in a few-shot setting manner. It receives a driving image along with multiple images of the target face to transfer the pose and expression of the driving face to the target face. At a high-level view, MarioNETte consists of a conditional generator and a discriminator which is only used for training the generator.. The generator's architecture is shown in Figure \ref{fig:arch:MarioNETte}. Three main components have been injected into the generator: the landmarks transformer, the image attention block, and the target feature alignment module. The landmarks transformer tackles the intrinsic gap between the key points of the driving and target faces. Image attention block handles multiple feature maps of the target person and pays attention to the right position of each feature. Target feature alignment improves the quality of the reenacted images by applying multi-level feature warping operations. In addition to these novelties, the generator utilizes two encoders for extracting the feature maps of the driving and target images and one decoder for generating the desired reenacted image.

\begin{figure}[htb]
    \centering
    \includegraphics[width=0.48\textwidth]{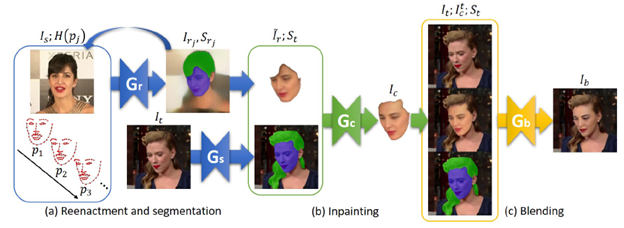}
    \caption{The FSGAN reenactment model \cite{nirkin2019fsgan}. }
    \label{fig:arch:FSGAN}
\end{figure}

FSGAN \cite{nirkin2019fsgan} is a subject agnostic reenactment model. As illustrated in Figure \ref{fig:arch:FSGAN}, FSGAN consists of three major parts: reenactment and segmentation, inpainting, and blending. In the first stage, two generators \(G_{r}\) and \(G_{s}\) are used for reenacting the source face, such that the expression and pose of the generated face are the same as the driving (target) face. In this stage, a sequence of facial landmarks of driving face in addition to the source image are fed into \(G_{r}\). The output of \(G_{r}\) is a reenacted source face \(I_{r}\) which has the pose and expression of the target face, and a segmentation of \(I_{r}\) into hair and face segments. Also, \(G_{s}\) segments target image into face and hair segments. The face segment of \(I_{r}\) is placed in the face segment of the target image. As expected, this may cause some missing pixels. So, in the second stage, by applying the inpainting network \(G_{c}\), missing pixels are colorized. Finally, in the last stage, the output of the previous stage is put on the target image and by the corporation of a blending network, the desired output will be generated. Some loss functions which are used for training FSGAN are domain specific perceptual loss, reconstruction loss, and adversarial loss. 

\begin{figure}[htb]
    \centering
    \includegraphics[width=0.48\textwidth]{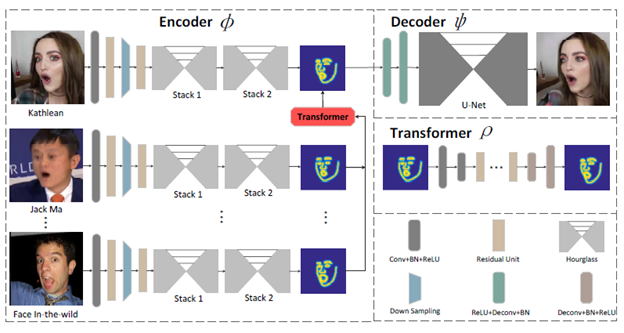}
    \caption{The structure of ReenactGAN \cite{wu2018reenactgan}.}
    \label{fig:arch:reenactGAN}
\end{figure}

ReenactGAN \cite{wu2018reenactgan} is a subject-specific reenactment model. As depicted in Figure \ref{fig:arch:reenactGAN}, ReenactGAN comprises three major components: an encoder, a transformer, and a decoder. The encoder maps the source and driving images into a landmarks latent space. Then the transformer blends the latent representations of the two sets of landmarks and creates a landmarks image that has the identity of source image and expression of the driving image. Finally, the decoder generates the reenacted image from the landmarks image. The transformer and decoder are target-specific and must be learned for each source identity. Cycle loss, adversarial loss, and shape loss are used in training the transformer. Also, reconstruction and adversarial losses are used in training the decoder.

\subsection{Loss Functions}
Several loss functions have been used for training face editing models. Suppose $(x^a)$ is an input image with attribute vector $(a)$. The editing model manipulates $(x^a)$ and generates $(x^{\hat{b}} = G(x^{a},b))$ via its generator $(G)$ regarding the desired attribute vector $(b)$. In order to generate more realistic and fidelity edited images while preserving the attribute-irrelevant portions, the some loss functions have been broadly used to train the editing models, regarding four main criteria: 1) the quality of the synthesized image, 2) the reconstruction error for the input image, 3) the correctness of the desired attribute changes, and 4) the preservation of irrelevant attributes and regions.

\textbf{Adversarial loss:}
As all adversarial networks, adversarial loss is applied to both generator and discriminator networks in order to generate more realistic output images. It also tries to make the output images' distribution close to that of the real images. The adversarial loss for discriminator $(d)$ and generator $(g)$ is defined as follows:

\begin{equation}
\begin{aligned}
\mathcal{L}_{adv_d} = &-\mathbb{E}_{x^a \sim p_{data}} log( D(x^a))\\
&-\mathbb{E} _{x^{\hat{b}} \sim p_{g}, b \sim p_{attr}}  log (1 - D(x^{\hat{b}}))
\end{aligned}
\end{equation}

\begin{equation}
\mathcal{L}_{adv_g}=-\mathbb{E}_{\hat{x}^b \sim p_{g},b\sim p_{attr}}[log(D(x^{\hat{b}})],
\end{equation}
where \(D\) is the discriminator network, \(p_{data}\) shows the distribution of the real world images, \(p_{attr}\) represents the distribution of the attribute vector, and \(p_g\) demonstrates the distribution of the generated images.

\textbf{Conditional attribute loss:}
In an editing task, the generated image should possess the condition attributes as well as being realistic. To reach this end, discriminator network of these models has two branches each providing a different output. One for adversarial loss described above and the other for the classification error.
Conditional attribute loss ensures that the requested edits are applied correctly and the generated image has all desired attributes determined by the attribute vector \(b\). In most cases, the attribute vector is represented as a categorical vector and the loss is called \textit{attribute classification loss} while, in some cases, e.g. in the case of using action units (AUs) as facial expression descriptors, the attributes' values are continuous and \textit{attribute regression loss} is used instead. This loss function is formulated as:

\begin{equation}
\begin{aligned}
\mathcal{L}_{cls} = &\mathbb{E}_{x^{\hat{b}} \sim p_{g},b \sim p_{attr}}[-log D_{cls}(b|x^{\hat{b}})]\\
&+ \mathbb{E}_{x^b \sim p_{data}}[-log D_{cls}(b|x^b)]
\end{aligned}
\end{equation}
where \(D_{cls}\) demonstrates the classification branch of the discriminator. The first term of the above loss function calculates the classification error for generated images and is used for training the generator and the second term gives the error over real data and is only used for training the attribute classifier itself. 

\textbf{Reconstruction loss:}
When the input image is edited based on its original attribute vector $(a)$, the generated output image $(x^{\hat{a}})$ is expected to be the same as the input image $(x^a)$. This constraint guarantees that excluded attributes will remain untouched during the editing process. This objective function is formulated as:
\begin{equation}
\begin{aligned}
\mathcal{L}_{\text {rec}} =\mathbb{E}_{x^a \sim p_{data}, x^{\hat{a}} =G(x^a,a)} \bigg [ \left\|x^a-x^{\hat{a}} \right\|_{1} \bigg ]
\end{aligned}
\end{equation}


\textbf{Consistency loss:}
consistency loss, also called \textit{recovery loss} or \textit{identity loss}, evaluates how similar are \(G(G(x^a; b);a)\) and the input image \(x^a\). It is defined as:

\begin{equation}
\begin{aligned}
\mathcal{L}_{consistency} = \mathbb{E}_{x^a \sim p_{data}, b \sim p_{attr}} \bigg [ \left\|G\left(G\left(x^a , b \right) , a \right)-x^a\right\|^{2} \bigg ]
\end{aligned}
\end{equation}

\textbf{Perceptual loss:}
To check the content similarity of the input image and the edited image in the semantic level, perceptual loss penalizes the difference between features of the two images extracted by a pre-trained feature extractor network, e.g. VGG19.

\textbf{Attention loss:}
In mask-guided approach, as stated before, an attention mask \(A\) is responsible for specifying the regions which need modifications. However, the attention mask can easily saturate to 1 and lose its effect. So, the attention loss is introduced to regularize the mask and make it as sparse as possible to prevent saturation. It is defined as L1 or L2 norm of the generated mask.

\textbf{Total variation loss:}
Total variation (TV) loss is not a common loss function among the facial attribute editing models. However, in some models, this loss function has been used to encourage models to generate spatially smooth images.

\textbf{Other losses:}
Some works have also proposed their own new loss functions to make the training procedure more stable or to generate output images with higher quality. 
For example, Kim et al. \cite{kim2021not} introduce CAM-consistency loss which overcomes cycle-consistency loss's weakness and can be applied to all editing models. The cycle-consistency loss cannot preserve details in regions that are not relevant to the target attributes. While the input and the regenerated images have differences and cannot be compared directly, CAM-consistency only considers attribute-irrelevant parts of two images in the comparison. To this end, an auxiliary classifier attends to the model to determine the regions of interest. Grad-CAM is a visualization method which specifies where on an image a classifier attends more regarding a specific class. The output of Grad-CAM can be used as a mask of attribute relevant parts on the input image. The CAM-consistency loss is formulated as:
\begin{equation}
    \mathcal{L}_{CAM}=\mathbb{E}_{x^a \sim p_{data}, x^{\hat{b}} =G(x^a,b)} \bigg [ 1-M_{CAM} \odot ||x^a-\hat{x}^b||_1 \bigg ]
\end{equation}
where \( M_{CAM}\) is the attention mask generated by Grad-CAM.

\section{Datasets and Evaluation Metrics} \label{sec4}
\subsection{Datasets}

\begin{table*}[!htbp]
\centering
\caption{The list of the datasets related to facial attribute editing tasks.}
\resizebox{\textwidth}{!}{%
\begin{tabular}{|c|c|c|c|c|c|c|c|}
\hline
\textbf{No} &
  \textbf{Dataset} &
  \textbf{\#Features} &
  \textbf{Year} &
  \textbf{\#Samples} &
  \textbf{\#Identities} &
  \textbf{Uncontrolled/   Controlled Environment} &
  \textbf{Annotation} \\ \hline
1 &
  CAS-PEAL   \cite{gao2007cas} &
  3 &
  2003 &
  99594 &
  1040 &
  Controlled   environment &
  Pose,   Expression \\ \hline
2 &
  MORPH  \cite{ricanek2006morph} &
  1 &
  2006 &
  55134 &
  13617 &
  Controlled   environment &
  Age \\ \hline
3 &
  LFW   \cite{huang2008labeled} &
  1 &
  2007 &
  13233 &
  5749 &
  Uncontrolled   environment &
  Identity \\ \hline
4  & FaceTracer   \cite{kumar2008facetracer}  & 10 & 2008 & 15000 & 15000 & Uncontrolled   environment & Facial   attribute, Expression      \\ \hline
5 &
  PubFig   \cite{kumar2009attribute} &
  73 &
  2009 &
  58797 &
  200 &
  Uncontrolled   environment &
  \_ \\ \hline
6 &
  MultiPIE   \cite{gross2010multi} &
  2 &
  2010 &
  $\sim$750000 &
  337 &
  Controlled   environment &
  Pose,   Expression \\ \hline
7 &
  RaFD   \cite{langner2010presentation} &
  3 &
  2010 &
  8040 &
  67 &
  Controlled   environment &
  Expression,   Pose, Gaze direction \\ \hline
8 &
  FaceWarehouse   \cite{cao2013facewarehouse} &
  1 &
  2013 &
  3000 &
  150 &
  Controlled   environment &
  3D   Expression \\ \hline
9 &
  CACD   \cite{chen2014cross} &
  1 &
  2014 &
  163446 &
  2000 &
  Uncontrolled   environment &
  Identity \\ \hline
10 &
  FGNET   \cite{fu2014interestingness} &
  1 &
  2014 &
  1002 &
  82 &
  Uncontrolled   environment &
  Age \\ \hline
11 &
  CelebA   \cite{liu2015deep} &
  40 &
  2015 &
  202599 &
  10177 &
  Uncontrolled   environment &
  40   facial attributes+ 5 landmarks \\ \hline
12 &
  IMDB-WIKI\cite{rothe2018deep} &
  2 &
  2015 &
  523051 &
  20284 &
  Uncontrolled   environment &
  Age,   Gender, Identity \\ \hline
13 &
  IJB-A   \cite{klare2015pushing} &
  1 &
  2015 &
  5712 &
  500 &
  Uncontrolled   environment &
  Identity \\ \hline
14 &
  CFP \cite{sengupta2016frontal} &
  1 &
  2016 &
  7000 &
  500 &
  Uncontrolled   environment &
  Identity,   Landmarks \\ \hline
15 & celebA-HQ   \cite{karras2017progressive} & 40 & 2017 & 30000 & 30000 & Uncontrolled   enviroment & 40   facial attributes+ 5 landmarks \\ \hline
16 &
  PPB   \cite{buolamwini2018gender} &
  3 &
  2017 &
  1270 &
  1270 &
  Controlled   environment &
  Gender,   Age \\ \hline
17 &
  Face   aging \cite{liu2017face} &
  1 &
  2017 &
  15030 &
  15030 &
  Uncontrolled   environment &
  Age \\ \hline
18 &
  UTKFace   \cite{zhang2017age} &
  3 &
  2017 &
  $\sim$20000 &
  20000 &
  Uncontrolled   environment &
  Age,   Gender, 68 Landmarks \\ \hline
19 &
  AgeDB\cite{moschoglou2017agedb} &
  1 &
  2017 &
  16488 &
  568 &
  Uncontrolled   environment &
  Age \\ \hline
20 &
  CLF   \cite{deb2018longitudinal} &
  1 &
  2018 &
  3682 &
  919 &
  Uncontrolled   environment &
  Age \\ \hline
21 &
  WFLW   \cite{wu2018look} &
  \_ &
  2018 &
  10000 &
  10000 &
  Uncontrolled   environment &
  98   Landmarks \\ \hline
22 &
  IMDB-Face   \cite{wang2018devil} &
  1 &
  2018 &
  1700000 &
  59000 &
  Uncontrolled   environment &
  Identity \\ \hline
23 &
  FairFace   \cite{karkkainen2019fairface} &
  3 &
  2019 &
  108501 &
  108501 &
  Uncontrolled   environment &
  Race,   Gender, Age \\ \hline
24 &
  MVF-HQ   \cite{fu2021high} &
  3 &
  2019 &
  120283 &
  479 &
  Controlled   environment &
  Pose,   Expression \\ \hline
25 &
  CelebAMask-HQ   \cite{lee2020maskgan} &
  19 &
  2020 &
  30000 &
  30000 &
  Uncontrolled   environment &
  Facial   component masks\\ \hline
26 &
  MAAD-Face   \cite{terhorst2020maad} &
  47 &
  2020 &
  3300000 &
  9100 &
  Uncontrolled   environment &
  47 Binary facial attributes \\ \hline
\end{tabular}%
}
\label{tabel:dataset:datasets}
\end{table*}

Numerous datasets have been proposed for face-related researches in the past decades. In this section, we review the datasets with facial attribute annotations that can be used in facial attribute editing tasks although most of them have been initially proposed for other tasks like face recognition or facial attribute classification. Based on the type of their annotations, the related datasets can be categorized into four main groups: multi-attribute, identity, age, and pose-expression. In the datasets of the first group, images have been labeled with many facial attributes and have been mainly used in facial attribute classification and editing models. In the identity datasets, there are a number of distinct identities and several face images of each identity which makes them suitable for face recognition and verification tasks. Datasets in the third group have been annotated with the age of the individuals and therefore are appropriate for developing age-invariant face recognition/verification systems and face aging models. The last group of the face datasets is dedicated to pose and expression features usually recorded under laboratory conditions and can be used for developing face reenactment models. In the following, a number of datasets in each category are reviewed. The datasets are also summarized in Table \ref{tabel:dataset:datasets}.

\textbf{Multi-attribute:}  CelebA \cite{liu2015deep} is a well-known dataset that has provided a diverse set of annotations for all of its 202599 images. The annotations include 40 binary attributes such as \emph{gender}, \emph{wavy hair}, \emph{eyeglasses}, \emph{bangs}, \emph{oval face}, \emph{smiling}, \emph{mustache}, and a lot of other binary attributes. In addition to these 40 attributes, each image in CelebA dataset has 5 landmark locations. CelebA is useful for training facial attribute classifiers. CelebA-HQ \cite{ karras2017progressive} dataset is a high-quality version of the CelebA which has 30000 images each upsampled from an image of CelebA. Similar to CelebA, all images in CeleA-HQ have 5 landmark points along with 40 attributes annotations.  CelebAMask-HQ \cite{lee2020maskgan} dataset is a version of CelebA-HQ which provides a mask photo for each image in CelebA-HQ dataset. The mask image locates 19 facial components, such as eyes, eyebrows, nose, skin, hair, hat, eyeglasses, ears, mouth, lip, earrings, necklace, cloth, and neck. PubFig \cite{kumar2009attribute}, Helen \cite{le2012interactive}, and FaceTracer \cite{kumar2008facetracer} are other datasets that cover multi-attributes. 

\textbf{Identity:} LFW (Labeled Face in the Wild) dataset \cite{huang2008labeled} is a quite popular dataset containing 13,233 images of 5,749 individuals, of which 1,680 identities have two or more images. All images have been collected from the web and have been annotated with people's names. LFW is widely used in unconstrained face recognition problems. IMDB-Face \cite{wang2018devil} is a more recent large-scale noise-controlled dataset that contains 1.7M face images of 5.9K identities that are captured from the IMDb website. IJB-A \cite{klare2015pushing} dataset consists of 5,712 images and 2,085 videos from 500 identities and has been designed for a specific face recognition challenge. CFP (Celebrities in Frontal-Profile) \cite{sengupta2016frontal} is another dataset of this category that comprises 7000 face images from 500 identities.  

\textbf{Age:} All datasets of this category provide the age label for their face images. Morph \cite{ricanek2006morph}  contains 55,134 photos of 13,617 individuals between 16 to 77 years old and is a suitable dataset for facial age estimation tasks. Face Aging \cite{ liu2017face} is another similar dataset in which 15,030 images are divided into 7 age groups including 0-10, 11-18, 19-30, 31-40, 41-50, 51-60 and 60+. The ages of the face images of FGNET \cite{fu2014interestingness} vary from 0 to 69. AgeDB \cite{moschoglou2017agedb} dataset contains 16488 photos of 568 subjects that are 0 to 100 years old and are divided into 10 groups. CACD (Cross-Age Celebrity Dataset) \cite{chen2014cross} is a collection of 163,446 images from 2,000 celebrities. Each image is annotated with the year which the photo was taken (2004-2013). CLF (Children Longitudinal Face) \cite{deb2018longitudinal} is devoted to age groups between 2 to 18. CALFW (Cross-Age LFW) dataset \cite{zheng2017cross}  covers ages between 0 to 60. FairFace \cite{karkkainen2019fairface}, PPB \cite{buolamwini2018gender}, UTKFace \cite{zhang2017age}, and IMDB-WIKI \cite{rothe2018deep} are other qualified datasets for facial age-related tasks.

\textbf{Pose and Expression:} The focus of this group of datasets is on the pose and expression of the face. MultiPIE (Multi Pose, Illumination, Expressions) \cite{gross2010multi} dataset's images have been recorded under certain conditions with 19 flashlights in different locations. More than 750,000 images of 337 subjects in this dataset cover various poses and expressions. RaFD (Radboud Faces Database) \cite{langner2010presentation} encompass 8,040 face images of 67 subjects which display 8 different expressions in three different gaze directions and five poses. MVF-HQ \cite{fu2021high} covers 13 poses and three different expressions and has been recorded in the laboratory conditions. FaceWarehouse \cite{cao2013facewarehouse} is a 3D dataset, in which all images have been labeled by one of 20 expressions. CFEE\cite{du2014compound} dataset has divided its face images into 22 expression groups including \emph{disgusted}, \emph{happily surprised}, \emph{happily disgusted}, \emph{sadly fearful}, \emph{sadly angry}, and \emph{sadly surprised}. EmotioNet \cite{fabian2016emotionet} dataset has also focused on expression attribute. The main difference between the EmotioNet dataset and all previous expression datasets is that the images of EmotioNet dataset are not taken in certain conditions or a laboratory. 

\subsection{Evaluation Metrics}
The evaluation metrics employed for assessing the goodness of the proposed facial image manipulation methods can be divided into qualitative and quantitative measures. The qualitative evaluations are mainly based on the human appraisal of the manipulated images while the quantitative methods assign a numerical score to each generated image which expresses the quality of the image itself or the edit operation performed on the image. In the following, the main evaluation metrics of each category are presented.

\subsubsection{Qualitative Analysis}
As the output of facial image editing is generally produced for human audiences, qualitative evaluations and comparisons are quite prevalent in facial image editing literature. Two main methods of this category of metrics are visual analysis and user study. Almost all of the studied models report sample outputs of their proposed models and visually compare them with the results of other methods or report the results of some user studies on these photos. 

\textbf{Visual Analysis}: In this evaluation method, sample outputs of the proposed models are presented and general quality aspects of the images or specific target attributes are investigated. The presence of artifacts, blurriness/sharpness of the image, and having natural skin color and texture are some general factors usually considered in the visual analysis of the manipulated images regardless of the target of manipulation. However, an important part of visual evaluations is based on the goal of the image manipulation task. As mentioned earlier, the goal of these models is to change the value of categorical attributes or to control the intensity of continuous attributes \cite{song2020face}. The outputs of the facial attribute editing models should have the target attribute values, e.g. smiling, while the non-target attributes remain unchanged compared to the input image which shows the disentanglement of the attributes in the proposed models \cite{li2020novel}. The aim of the visual analysis is to check these properties of the reported results. 
While the visual analysis of the results is essential in this field, it is not sufficient as the evaluation is limited to a small subset of manipulated images.

\textbf{User Study}: In user studies, a relatively large group of people are asked to give their opinions about the quality of the outputs of a model or to compare the results of two or more models. The aggregated outcomes of the study are then reported as an evaluation metric. In a simple scenario, the results of the models on a number of images are shown to the participants to select the best results for each image. The number of images for which a model is selected as the best performing model is used as a measure of the goodness of that model. To prepare a user study, one has to decide on the methods to be compared, the test images, how the outputs are going to be evaluated or compared (e.g. image quality, identity preservation, or naturalness), the group of evaluators, and the statistical measures that are going to be calculated from the users' opinions. There are some tools to perform these user studies, such as Amazon Mechanical Turk (AMT) and Google Form, while it also possible to use self-produced applications. Most of the works studied in this survey have conducted some sort of user studies to compare their results with other models from human points of view (e.g. \cite{or2020lifespan,ghosh2020gif,tan2020michigan,kowalski2020config,chang2018pairedcyclegan,li2018beautygan,liu2019stgan,li2019attribute,zhu2019ugan,guo2019mulgan}).

It should be noted that the user studies can be considered either qualitative or quantitative evaluation metric as on one hand, the quality of the generated images is evaluated based on human judgments (as in the qualitative measures), and on the other hand, some result is reported as numerical scores (as in the quantitative measures). The lack of other appropriate quantitative evaluation metrics has encouraged the use of user studies as a quantitative metric in many researches. 

If the test images, the evaluating population, and the polling strategies are set up carefully, the results of the user studies are more reliable than the visual analysis. The set of test images should be diverse enough to properly cover different challenges of the related task. The evaluating groups should not be biased towards a special model and must answer the questions carefully. Finally, the questionnaire should be devised carefully; for example, it is a good practice to shuffle the results of different models for each test image so that the answerers are not biased toward a special choice through the evaluation process. However, visual analysis provides a more tangible understanding of the quality of the edited images and in practice it is better to how both visual analysis and user studies in your work.

\subsubsection{Quantitative Analysis}
There are some quantitative metrics for assessing the performance of various facial attribute editing models. These measures assign numerical scores to each manipulated image based on a mathematical formula. Compared to human-based qualitative evaluations, quantitative measures are reproducible and are not subject to human biases. However, it is difficult to have perfect quantitative measures for some desired qualities.   

The quantitative metrics can be divided into two main groups. The first group includes image quality assessment metrics trying to measure the quality of a resulting face image independent of the requested edits while the second group investigates a pair of input and output face images to analyze the performance of the model with respect to the expected modifications.

\textbf{1) Image Quality Assessment (IQA):} This group of metrics aims to quantize the quality of the edited images. Two main types of IQA metrics are non-reference and full-reference quality metrics. In non-reference IQA, the evaluation algorithm just gets one input (generated image) and estimates the quality of the image, while in full-reference IQA, the resulting image is compared with a ground-truth and a score is assigned according to the similarity of the resulting image to the expected target image. While these two types of quality measures are usually calculated by predefined formulas, some neural networks have been proposed for this purpose where the metric can be defined by a set of labeled images. All of these evaluation metrics and methods are studied in the following.

\textbf{Non-reference IQA:} 
\begin{itemize}
    \item \textbf{Inception Score (IS)}
     has been introduced by Salimans et al. \cite{salimans2016improved} for evaluating the quality of synthesized images by GANs. IS is usually used in training a GAN or an image-to-image translation model to guide the models to generate high-quality images \cite{zhang2019adversarially}. A modified version of IS is Conditional Inception Score (CIS) \cite{huang2018multimodal} which is more applicable in multi-modal image translation works. Although both IS and CIS seek to capture the high quality and diversity of real data, they are not usually used in evaluating facial attribute editing models. A drawback of IS is that it does not compare the statistics of real-world samples with those of the synthetic samples which has been solved in Frechet Inception Distance (FID)score.
    \item \textbf{Frechet Inception Distance (FID)} score has been introduced by \cite{heusel2017gans} and is a standard metric for evaluating the quality of generated images (here, edited images) by GANs. FID score is an improved version of the inception score and demonstrates the distance between the distribution of the synthesized images and that of the real images. FID, also known as Wasserstein-2 distance, takes its name from the Inception model and Frechet Distance. At first, feature vectors of two datasets (1024 real images and 1024 synthesized images) are extracted using the inception network. Then, the means and covariances of these two datasets are calculated. Finally, FID is calculated as the Frechet distance of two sets of feature vectors' distributions. FID is formulated as equation \ref{equ:fid} and its lower values demonstrate the better quality of manipulated images. It is notable that experiments have shown that FID values are generally consistent with human judgment \cite{heusel2017gans}. Many studies have utilized FID score to evaluate the quality of the manipulated face images, such as \cite{zhang2019adversarially, ying2019clsgan, guo2019mulgan, zhu2019ugan, abdal2021styleflow, chu2020sscgan, esser2020disentangling, collins2020editing, viazovetskyi2020stylegan2, kowalski2020config, tan2020michigan, li2020novel}. If the feature vectors of the images generated by the GAN model have multinomial Gaussian distribution \(N(\boldsymbol{m}, \boldsymbol{C})\) and the features of the real world images have \(N(\boldsymbol{m}_{w}, \boldsymbol{C}_{w})\), FID score is defined as:
    
    \begin{equation}
    \begin{aligned}
    \label{equ:fid}
    d^{2}\left((\boldsymbol{m}, \boldsymbol{C}),\left(\boldsymbol{m}_{w}, \boldsymbol{C}_{w}\right)\right) &=\left\|\boldsymbol{m}-\boldsymbol{m}_{w}\right\|_{2}^{2} \\ &+\operatorname{Tr}\left(\boldsymbol{C}+\boldsymbol{C}_{w}-2\left(\boldsymbol{C C}_{w}\right)^{1 / 2}\right)
    \end{aligned}
    \end{equation}
    
\end{itemize}

\textbf{Full-Reference IQA:} The structural similarity  (SSIM) index, peak signal-to-noise ratio (PSNR), and learned perceptual image patch similarity (LPIPS) are some of the most common full-reference IQA metrics utilized in most papers. In the following, these measures are described in more details. 
  \begin{itemize}
    \item \textbf{SSIM} \cite{wang2004image} is based on the human visual system which is adapted for obtaining structural information from the scene. SSIM can provide an approximation to image quality perceived by human. This metric is based on the concepts of local luminance, contrast, and structure in two images. If $x$ and $y$ are two non-negative images, the luminance, contrast, and structure concepts are defined in equations \ref{equ:luminance}, \ref{equ:contrast} and \ref{equ:structure}, respectively. 
    
    \begin{equation}
    \label{equ:luminance}
        l(\mathbf{x}, \mathbf{y})=\frac{2 \mu_{x} \mu_{y}+C_{1}}{\mu_{x}^{2}+\mu_{y}^{2}+C_{1}}
    \end{equation}
    
    \begin{equation}    
    \label{equ:contrast}
        c(\mathbf{x}, \mathbf{y})=\frac{2 \sigma_{x} \sigma_{y}+C_{2}}{\sigma_{x}^{2}+\sigma_{y}^{2}+C_{2}}
    \end{equation}
    
    \begin{equation}  
    \label{equ:structure}      
        s(\mathbf{x}, \mathbf{y})=\frac{\sigma_{x y}+C_{3}}{\sigma_{x} \sigma_{y}+C_{3}}
    \end{equation}
    
    where $\mu_{x}$, $\mu_{y}$, $\sigma_{x}$, $\sigma_{y}$ and $\sigma_{x y}$ are the mean intensities of $x$ and $y$, the standard deviations of $x$ and $y$, and correlation of $x$ and $y$, respectively. Also, $C_{i}$s are small constants defined as $C_{i}=\left(K_{i} L\right)^{2}$, where $L$ is the dynamic range of the pixel values and $K_{i}<<1$ ($1 \leq i \leq 3$) are small constants. Finally, SSIM index is defined as the combination of these three quantities as follows:
    
    \begin{equation}
    \label{equ:ssim_1}
        \operatorname{SSIM}(\mathbf{x}, \mathbf{y})=[l(\mathbf{x}, \mathbf{y})]^{\alpha} \cdot[c(\mathbf{x}, \mathbf{y})]^{\beta} \cdot[s(\mathbf{x}, \mathbf{y})]^{\gamma}
    \end{equation} 
    
    where $\alpha > 0$, $\beta > 0$ and $\gamma > 0$ adjust the relative importance of the three components. To simplify the equation, the parameters are set as $\alpha=\beta=\gamma=1$  and  $C_{3}=C_{2} / 2$ in \cite{wang2004image} and the following equation is obtained which is the most common form of SSIM:
     
    \begin{equation}
    \label{equ:ssim_2}
        \operatorname{SSIM}(\mathbf{x}, \mathbf{y})=\frac{\left(2 \mu_{x} \mu_{y}+C_{1}\right)\left(2 \sigma_{x y}+C_{2}\right)}{\left(\mu_{x}^{2}+\mu_{y}^{2}+C_{1}\right)\left(\sigma_{x}^{2}+\sigma_{y}^{2}+C_{2}\right)}
    \end{equation}
    
    This metric is used in many papers such as \cite{li2020novel,tewari2020pie,ning2020fegan,ning2021continuous,song2020face,guan2020collaborative,wei2020maggan,chen2020coogan,liu2019stgan,ying2019clsgan,zhang2020mu}. There are also some other versions of SSIM index such as MSSIM (mean SSIM) and MS-SSIM (multiscale SSIM). MSSIM is the average SSIM of studied images and MS-SSIM \cite{wang2003multiscale} considers the image details at different resolutions. In MS-SSIM, a low-pass filter is applied to the input images and the filtered images are iteratively downsampled by a factor of 2. The original images are indexed as scale 1, and the highest scale is as scale $M$, obtained after $M-1$ iterations. The contrast and structure components are calculated for each scale $j$ as $c_j(x,y)$ and $s_j(x,y)$, respectively, while the luminance is only computed at scale $M$ ($l_M(x,y)$). Finally, MS-SSIM is obtained by the following equation:
    
    \begin{equation}
        \operatorname{MS-SSIM}(\mathbf{x}, \mathbf{y})=\left[l_{M}(\mathbf{x}, \mathbf{y})\right]^{\alpha_{M}} \cdot \prod_{j=1}^{M}\left[c_{j}(\mathbf{x}, \mathbf{y})\right]^{\beta_{j}}\left[s_{j}(\mathbf{x}, \mathbf{y})\right]^{\gamma_{j}}
    \end{equation}
    where $\alpha_{M}$, $\beta_{j}$ and $\gamma_{j}$, similar to equation \ref{equ:ssim_1} adjust the relative importance of different components \cite{natsume2018rsgan}.

    \item \textbf{PSNR} is computed between two 8-bit grayscale images $x$ and $y$ \cite{hore2010image}, both of size $M \times N$ as follows:

     \begin{equation}
     \label{equ:psnr}
        \operatorname{PSNR}(x, y)=10 \log _{10}\left(255^{2} / \operatorname{MSE}(x, y)\right)
    \end{equation}
    where MSE is defined as:
     \begin{equation} 
        \operatorname{MSE}(x, y)=\frac{1}{M N} \sum_{i=1}^{M} \sum_{j=1}^{N}\left(x_{i j}-y_{i j}\right)^{2}
    \end{equation}
    
    When the MSE tends to zero the PSNR will goes to infinity. This means that whenever the high PSNR shows high quality of the generated images and low PSNR demonstrates the high distinction between the two images. MSE is also another metric that is used to compare a resulting image with its reference. Both of these metrics are good at assessing the quality of the noisy images based on the amount and intensity of the noise but do not properly reflect the structural content of the image. Many authors have used MSE \cite{richardson2020encoding} and PSNR to evaluate their proposed facial image manipulation models \cite{li2020novel,tewari2020pie,ning2020fegan,ning2021continuous,song2020face,guan2020collaborative,wei2020maggan,chen2020coogan,liu2019stgan,zhang2020mu}.

    \item \textbf{LPIPS}, introduced in \cite{zhang2018unreasonable}, measures perceptual-level similarity and is claimed to be consistent with human perception \cite{guan2020collaborative}. This metric is based on the distance of the projected deep features obtained from pre-trained networks such as VGGFace \cite{parkhi2015deep}. Lower values of LPIPS show that the evaluated images are close in the feature space and similar in the perceptual level. To be more precise, for two images $x$ and $y$ and the feature extractor network $F$, LPIPS is computed as equation \ref{equ:LPIPS}.
    
    In this equation, firstly, the feature stack from $L$ layers is extracted and unit-normalized in the dimension of the channel as $\hat{x}^{l}, \hat{y}^{l} \in \mathbb{R}^{H_{l} \times W_{l} \times C_{l}}$ for layer $l$. Then the activations are scaled channelwise by vector $w^{l} \in \mathbb{R}^{C_{l}}$ and the $L2$ distance is computed. Finally, spatial averaging and channel-wise summation is applied. This metric is used in \cite{li2019attribute,guan2020collaborative,richardson2020encoding,li2020novel}.
    
    \begin{equation}
    \label{equ:LPIPS}
        d\left(y, x\right)=\sum_{l} \frac{1}{H_{l} W_{l}} \sum_{h, w}\left\|w_{l} \odot\left(\hat{y}_{h w}^{l}-\hat{x}_{hw}^{l}\right)\right\|_{2}^{2}
    \end{equation}

\end{itemize}
  
\textbf{IQA with networks} 
Another approach for image quality assessment is to use a neural network to classify edited images in two or more quality levels. Some researchers use employ pre-trained networks for this purpose while the others train their own domain-specific models. Both approaches are common among quality assessment tools \cite{tripathy2020icface}. As well, these networks could work in full-reference or non-reference manner. For example, DeepIQA \cite{bosse2017deep} performs full reference quality assessment while CNNIQA \cite{kang2014convolutional} is a network for non-reference quality estimation. 

\textbf{2) Attribute edit/preservation analysis:}
In addition to generating high-quality face images, facial attribute editing models should correctly fulfill the editing task that is 1) accurately apply the requested changes to the target attributes of the input face image, and 2) do not change the other attributes of the face. So, the performance of a facial attribute editing model should be assessed from these two aspects: accurate target attribute editing and non-target attribute preserving. 

\textbf{Accurate Editing:} 
To evaluate attribute editing accuracy, a multi-label attribute classifier is employed to predict attribute values for generated images. The editing accuracy of the model is then estimated by comparing the attribute values of the resultant face image with the desired values. Many works such as \cite{he2019attgan,liu2020gan,ning2020fegan,ning2021continuous,wei2020maggan,shen2020interfacegan,chu2020sscgan,chen2020coogan,liu2019stgan,ying2019clsgan,zhang2019adversarially,kwak2020cafe,yu2021pose} use this approach as a quantitative metric to assess the performance of their attribute editing models.
In the face aging models like \cite{sheng2020face,shi2020can,he2019s2gan,yao2020high}, an age predictor model (usually the publicly available Face++ API) is used to estimate the age of the edited image, and the aging task is assumed to have been successfully done if the estimated age of the edited face image is the same as the desired age group. 
Some special-purpose face editing models have employed specific measures for realizing the accuracy of the modifications. For example, \cite{tang2020fine} uses mean squared error (MSE) and Pearson's correlation coefficient (PCC) between action units (AUs) of input and generated images to quantify expression manipulation accuracy.
Also, Tewari et al. \cite{tewari2020pie} use the landmark space instead of the image space to measure the angular distance between the target and generated head poses. Average \(L2\) distance between 66 facial landmarks of desired and generated images is used as the alignment error to measure the performance of the head pose editing model.

\textbf{Attribute Preservation:} 
As mentioned earlier, these group of evaluation metrics aim to assess the ability of the facial attribute editing model in preserving irrelevant attributes and regions of the input face image which is also known as identity preservation property in expression and pose editing tasks. 

A common approach to this goal is to perform face verification between the input and the modified face images. To do so, the distance between the feature vectors of the initial and manipulated face images, extracted by pretrained face recognition models, are measured using a Euclidean or cosine distance and compared with a predetermined threshold to verify if the two images belong to the same person or not as in \cite{sheng2020face,sharma2020improved, abdal2021styleflow, natsume2018rsgan, shen2020interfacegan, zhu2019ugan, shi2020can, sheng2020face,yu2021pose}. Chang et al. \cite{chang2018pairedcyclegan} firstly frontalize all images and then apply the face verification.

Unlike aforementioned works which compare the input and edited images, Kowalski et al. \cite{kowalski2020config} considers an attribute and assigns two opposite values to it and generates a positive (\(I_+\)) and a negative (\(I_-\)) images using the model under investigation. They then compare these two images and expect them to differ only on the chosen attribute.
In \cite{nitzan2020disentangling}, attribute preservation is assessed in three aspects: Euclidean distance between normalized 2D landmarks of reference and output images which evaluates expression preservation, Euclidean distance between Euler angles of reference and output images which measures the pose preservation, and the result of face recognition which assesses the identity preservation.
Finally, Wei et al. \cite{wei2020maggan} propose a mask-aware reconstruction error to measure the ability of the model to preserve irrelevant regions specified by the mask.

\section{Applications} \label{sec5}
Semantic facial attribute editing has a great potential to be employed in a vast variety of real-world applications either as the whole solution or as a part of another system. In fact, whenever a modification to a face image is required, semantic face editing is among the best options. In this section, we briefly review some semantic face editing usages organized by the application areas. 
\subsection{Face recognition}
Face Recognition (FR) systems are ubiquitous computer vision programs in which a face image extracted from a photo or movie frame is matched against a database of face images of known identities. The applications of such systems range from photo tagging to access control. The success and accessibility of the GAN-based face editing models during the past years have stimulated many researchers to adopt these models either to promote the performance of FR systems or even to attack them. In the rest of this section, some applications of face editing models in the field of face recognition are reviewed.
\subsubsection{Face data augmentation}
Face data augmentation is the process of enriching a dataset of face images by inserting new machine-created face photos from those available in the dataset. The performance of DNN-based face processing models strongly depends on the size and variation of their training datasets. Insufficient training data, the lack of labeled samples, and unbalanced data distribution (dataset biases) will cause the overfitting and low performance of these models. On the other hand, collecting and labeling adequate samples with high quality and balanced distributions is laborious, expensive, and error-prone. Face photo augmentation is one of the solutions to tackle these limitations.

Face data augmentation tools can be classified into general image augmentation techniques such as cropping, flipping, and rotation of the images, and specific face editing methods which alter the attributes and components of the image via sophisticated face manipulation techniques including GANs. The augmented faces generated by controllable generative models can have specific facial attributes and hence make it possible to reach balanced datasets with more intra-class variations. Moreover, the ground truth labels of these new samples are automatically generated. GAN-based face photo manipulation models are ideal tools for face data augmentation as the edit of any facial attribute like expression, makeup, age, eyeglasses, and hairstyle produces new precious data for face recognition systems \cite{romero2019smit, zhang2018generative,wang2020survey}.

Deng et al. \cite{deng2018uv} succeeded to achieve 94-percent accuracy in face recognition on CFP dataset due to data augmentation. In this work, the focus is on increasing pose variation in training data by generating complete facial UV maps via the combination of a 3D Morphable Model (3DMM) and a GAN-based deep neural network. 

In addition to face recognition, face data augmentation has found applications in face retrieval tasks with the aim of finding all face images of a database that are related to a specific subject. For example, it may be needed to retrieve images of a person from an enormous database of face images captured by CCTV cameras. As an example of such models, Deep Face Hashing with GAN (DFH-GAN) is a hash-based approach for face image retrieval \cite{zhou2021dfh}. It consists of three modules: a generator to perform data augmentation, a discriminator with a shared CNN to learn multi-domain facial attributes, and a hash encoding component to generate hash codes. The generator is used to generate synthesized face images for training the main retrieval engine.

\subsubsection{Face frontalization and neutralization}
Performance of the modern face recognition systems dramatically degrades with pose and expressions of the target face image. It would be beneficial if the input face photo is converted to a neutral frontal one before processing by these systems. This specially gives a boost to the performance of face recognition from uncontrolled images.  The use of face attribute editing for creating frontal neutral face images is almost straightforward. DR-GAN \cite{tran2017disentangled} and TC-GAN \cite{cheng2019tc} are two purpose-built models developed for this application.
DR-GAN injects changes into both generator and discriminator of the usual GANs. The generator part has an encoder-decoder structure which empowers the model to learn generative and discriminative representations. The discriminator part, as in semi-supervised GANs, classifies the pose and identity of the image in addition to real/fake discrimination. This model calls for labeled faces to generate identity-preserving faces and labels the face pose by one-hot codes to ensure the correctness of generated faces' poses. TC-GAN, on the other hand, is composed of two generators \(G_{c}\), \(G_{f}\), and one discriminator \(D\) in a triangle network architecture. TC-GAN converts a profile image to the corresponding frontal contour by \(G_{c}\) and then adds the face features by \(G_{f}\) to generate a photo-realistic frontal view of the input image without unintentionally changing the identity or expression. 

pSp \cite{richardson2020encoding} is a general image-to-image translation model which has been successfully trained for face frontalization task  with weak supervision. Here, an encoder is trained to map the input image to the style vectors of a style-based generator which produce the final image. The key idea of pSp is to randomly flip training images that forces the model to learn a fixed frontal image close to both states of the input image. In another attempt, Luan et al. \cite{luan2020geometry} focus on facial pattern preserving in their face frontalization and recognition system and try to reach this goal by designing a discriminator which leverages self-attention blocks and a series of sub-discriminators. In another attempt in this field,  a self-attention-based generator and a face-attention-based discriminator are proposed for long-pose face frontalization\cite{yin2020dual}. The generator combines local features with long-range dependencies, and therefore, produces identity-preserved faces while the discriminator highlights local features of face regions and consequently strengthens the realism of generated frontal faces.

\subsubsection{De-identification}
The goal of face de-identification is to remove the identity of a face in a photo or video in order to preserve the privacy of individuals while the other attributes (e.g. sex, age, pose, expression, and illumination) remain unchanged\cite{cao2021personalized}. For example, though collecting huge datasets of images or videos in the wild is a necessary step in most of the computer vision tasks, protecting the privacy of the people captured in these images/videos is indispensable too. The need for reliable de-identification tools becomes more apparent when large image collections are going to be published on web either as a training set like ImageNet \cite{yang2021study} or through an application like Google Street View.
 
Blurring the face areas of the people in an image is a simple yet effective way for hiding the identities of the face images but these obfuscated faces render the images unpleasant for human users. A more intelligent way for performing de-identification is to replace the original face with a randomly synthesized face fitting the target person's body. In \cite{li2019identification}, the face photos of a number of donors are used for replacing the target faces and a facial attribute transfer model is used to map non-identity facial attributes of the original face to the donor's face photo. In a different approach, PP-GAN \cite{wu2019privacy} trains a U-net based generator for mapping an input face image to a de-identified face with similar structure. For training this generator, in addition to the original discriminator, a verificator and a regulator component are used. The former provides a contrastive loss based on the identities of the two faces and the latter gives the structural similarity loss to measure the similarity of two faces.
\subsection{Video generation}
Recently deepfake videos, as a direct outcome of the facial attribute editing models, have received great attention in social neural networks. The possible applications of facial attribute editing in the movie and animation industries are enormous and here we will look at some of them. 

\subsubsection{Movie, animation and deepfake production}
As mentioned before, several generative models have been proposed for transferring style from a so-called driver to a target image to create moving characters. The movement may happen in the lips, face, head or the full body and has caused great attention towards these models due to their applicability in animation, realistic animation and movie (deepfake) production. In the following, we will look at just a few examples of the large body of literature devoted to these applications. 

Deepfake refers to the use of deep learning techniques for creating fake images or vides, hence the name. In fact, deepfake is a kind of image/video manipulation which has experienced tremendous progress during the past years thanks to the rapid development of generative models, especially GANs. Deepfakes are generated in different ways including entire face swap, expression manipulation, face and body reenactment, and background editing. Regarding the face swap models, RSGAN \cite{natsume2018rsgan} makes use of two distinct networks to separately learn latent space representations of face and hair. It then uses a composer component to reconstruct swapped faces by combining the hair of one person with the face of the other. In addition to the adversarial loss, the composer is trained with another discriminator that checks if the local patches of the input images are realistic or not. FaceShifter \cite{li2019faceshifter} is another prominent face swap model based on a two-stage framework which also deals with facial occlusions. The first stage generates swapped faces using an identity encoder, a multi-level attributes encoder, and an Adaptive Attentional Denormalization (AAD) generator. The information of the two first components are integrated in the ADD-generator to create swapped faces. The main focus of the second stage of FaceShifter is on addressing the occlusion problem where a Heuristic Error Acknowledging Refinement Network (HEAR-Net) is designed to solve this issue by the use of a face segmentation network in a fully self-supervised way. 

Expression editing is the other method of creating fake images and videos in which an input image is modified to have a particular expression, e.g. furious, happy, or sad. For example, Wu et al. \cite{wu2020cascade} progressively manipulate the expression of face by cascading EF-GAN blocks each composed of two components: an expression transformer and a refiner. The expression transformer has four branches, one branch for global and three others for eye, nose, and mouth, expression manipulation. The refiner combines outputs of the expression transformer and refines them to the final editing. In some models, as in Labeled Free Expression Editing (LEED) \cite{wu2020leed}, the input image is first mapped to a corresponding neutral face and then the target expression is added to it. 

Motion transfer is almost the most popular technique among deepfake generators which has also caught the attention of movie and animation producers. As stated above, here, the motion patterns of a driver are transferred to the target person while the face identity attributes and body shapes remain intact. Face reenactment is a common form of motion transfer which may be used to produce a deepfake speech from a politician, revolutionize the movie dubbing industry or create a talking head from a picture of a dead person. Tripathy et al. \cite{tripathy2020icface} developed a face reenactment model which is trained on frames of videos and learns to transfer facial non-identity attributes from one face image to another image of the same identity at the train time but to a different identity in the test phase. The model, named ICface, is based on two generators: one for generating a neutral image from the target person and one for manipulating it as per the extracted non-identity attributes of the input face. As another example of these models, DAE-GAN \cite{zeng2020realistic} learns separate pose and identity representations for generating talking faces. A conditional GAN is then employed to perform the style transformation.

\subsubsection{Text or speech to video}
Automatic video generation has become a promising line of research. While the current generative models are still far from generating the whole body of characters with high resolution and all details, talking face video generation as a special case of video generation seems to be reachable by the face attribute manipulating models. Recently, generating talking faces based on the input text or audio have received attention due to its applicability in lip synchronization in dubbing and deepfake generation\cite{lu2021live}. Chen et al. \cite{chen2019sound} propose a temporal GAN network for generating audio-driven talking face videos. Their model firstly generates landmarks corresponding to the input audio and then synthesizes the desired output based on the generated landmarks. Yi et al. \cite{yi2020audio} have also introduced a similar model and have attempted to minimize the artifacts produced due to head rotation. To tackle this problem, they utilize a 3D face reconstruction network and a memory-augmented GAN. In addition to the generator and discriminator, the memory-augmented GAN has a memory network for remembering representative identity features during training. At the test time, the memory network retrieves best-match features and sends them to the generator. This model is capable to generate videos with personalized head poses. Text2Video \cite{zhang2021text2video} is another network that generates a video from a text source. In Text2Video, the input text is first converted to audio and in parallel, a series of landmarks are generated from input text by a phoneme-pose dictionary. Finally, by the help of vid2vid GAN, the synthesized audio and the landmarks series are transferred to the target talking face video.

\subsection{Face construction and reconstruction}
As mentioned earlier, in many applications, it is required to create a human either randomly or based on a description as in virtual reality generation, movie and animation production, de-identification and facial composite creation by forensic technicians. In most of these applications, the target face is fully or partially determined by a set of facial attribute values or a textual description. A straightforward strategy in these cases is to create an initial portrait and then adjust or manipulate it based on the modifications suggested by the person having the final face in his mind, e.g. the producer of movie or an eye-witness of a scene.  In the following, some applications of this scope are briefly reviewed.
 
\subsubsection{Cartoon generation}
The transformation of human photos to cartoon characters, comic book characters, caricature portraits or emoji faces is an interesting application of AI powered tools for social media users. The result is ready with just few clicks without any laborious or expensive manual work. The task may be formulated as an image-to-image translation with probable extra editing required by the users \cite{pinkney2020resolution}. As expected, GAN based models are ideal tools for the creation and manipulation of such images. One class of existing works like CartoonGAN \cite{chen2018cartoongan}, U-GAT-IT \cite{kim2019u}, AutoToon \cite{gong2020autotoon}, and ReStyle \cite{alaluf2021restyle} just translate a real face/scene photo to a cartoon-domain one. The other methods, like the one proposed by Wang et al. \cite{wang2021cross}, allow the edit of the face attributes in both real and cartoon domains. In this model, the StyleGAN2 model is fine tuned to generate cartoon faces from the latent style vectors. Regarding the attribute manipulation, two networks are employed for mapping the latent codes and the corresponding 3DMM parameters to each other and the attribute editing is performed on the 3DMM face model.

\subsubsection{3D face reconstruction}
3D face reconstruction is the task of creating a 3D model, e.g. 3D Morphable Model (3DMM), or a 3D mesh of a person from a photograph of his. These 3D models provide a basis for performing different edits such as changing the pose of the head or applying various lighting conditions, hence considered as a prerequisite of many face manipulation related applications like virtual reality. 
Luo et al. \cite{luo2021normalized} have introduced a GAN-based framework for learning a model that can generate a 3D relightable and animation-friendly avatar face from a single unconstrained photograph. Their framework is trained in two inference and refinement stages. At the first stage, they combine a non-linear 3DMM with StyleGAN2 to generate a normalized 3D face. At the second stage, they attempt to refine the model by reconstructing the input face image from the 3D face model generated in the first stage and the face-mask corresponding to the input face which is extracted by PSPNet. The final output of the model is a textured 3D face with neutral expression and normalized lighting conditions that can be used for further modifications. Lee et al. \cite{lee2020uncertainty} have also designed an encoder-decoder structured model for the same purpose. The uncertainty-aware encoder extracts features and their distribution from the input face photos and the decoder is a combination of Graph CNNs and GANs to reconstruct high-quality 3D faces from single in-the-wild face images. Yin et al. \cite{yin2021weakly} reconstruct 3D faces by designing a generator for UV maps which are 2D representations of the 3D models that can be learned by GANs. This model utilizes a UV sampler to synthesize the incomplete UV maps from the input face. Then, the generator network generates a complete \(UV_{perd}\) map from the incomplete one.

\subsection{Face photo post processing}
\subsubsection{Makeup and de-makeup}
Prominent makeup applications (e.g., Taaz and PortraintPro) can transfer selected makeup styles to a face photo, but the set of transferable makeup styles are limited and preconfigured. GAN-based face manipulation models have paved the way for creating more effective makeup transfer apps in which any makeup style from an input face can be transmitted to a target photo. BeautyGAN's \cite{li2018beautygan} is one of these tools that has cycle-GAN architecture and with its two cooperating GANs applies the makeup of a face image to another face photo. Nguyen et al. \cite{nguyen2021lipstick} extract UV maps from two inputs, then by transferring the color and pattern from makeup style to target image, generate the desired output. SCGAN \cite{deng2021spatially} splits makeup style into three parts (i.e. lip, eyes, and skin), making it possible to have makeup style transfer in different regions of the face. 

Some makeup effects make severe changes in facial appearance that causes a mismatch between the face with and without makeup. Forasmuch as this mismatch affects the performance of automatic face recognition systems used in security applications, de-makeup tools are also required for removing makeups from the input images. BTD-net \cite{cao2019makeup} is a bidirectional makeup/de-makeup network that can learn on unpaired data. This model uses one generator and one discriminator in each direction and is specially designed for removing makeup effects from the input face photos.

\subsubsection{Accessory removing and adding}
As stated early in this paper (Section \ref{sec2}), the existence of fashionable or necessary face accessories, (e.g., hat, glasses, sunglasses, jewelry, and mask) in a face photo is regarded as a facial attribute. Most of the general facial attribute editing methods studied in this paper support the addition or removal of some sorts of accessories, e.g. glasses – an interesting application for accessory suppliers and of course for the general public. The removal of face accessories is also important in face recognition systems as these accessories affect the representation of the face by the corresponding models and degrades the performance of the recognition algorithms.  One relatively successful model for this aim is ByeGlassesGAN \cite{lee2020byeglassesgan} whose generator consists of an encoder-decoder and a segmentation decoder. The encoder extracts non-accessory information of face to be used by the decoder to reconstruct the input face without accessories and the segmentation decoder tries to locate the region of eyeglasses on the input face and predict the corresponding mask to guide the reconstruction process. It is shown that a significant increase in the face recognition accuracy is achieved by the use of ByeGlassesGAN. 

\subsubsection{Aging and rejuvenation}
Age is a global continuous feature of a face and as any other attribute may be the subject of facial attribute editing. However, manipulating a face photo so as to look younger or older is a complicated task that requires sophisticated changes of the image. Recently, GAN-based aging applications have become quite popular among the users of social networks. For example, Or et al. \cite{or2020lifespan} have classified the age range of  0 to 70 into 6 intervals and have trained a model to perform age progression and regression on the input face image. Due to the need for generating images in different age groups, their model has a conditional generator which is a multi-domain image-to-image translation network. 
Yao et al. \cite{yao2020high} focus on the resolution of the generated age-edited faces. They use an encoder-decoder structure along with an age modulation layer. The encoder and decoder attempt to learn an age-invariant latent space, while the age modulation layer codes the target age into a 128-dimensional vector which is applied to the decoder. 
A3GAN \cite{liu20213} consists of an attribute-aware attentive generator, a wavelet-based multi-pathway discriminator, and a facial attribute embedding network. The generator learns to generate elderly face images by the use of a discriminator that works on multi-scale wavelet coefficients and learns to identify generated images from real ones. The role of the facial attribute embedding network is to decrease the matching ambiguity that exists in unpaired data. Both generator and discriminator use the output of the facial attribute embedding network. A3GAN is a unidirectional model and does not work for rejuvenation.

\section{Challenges}
Despite all of the efforts put into the development of semantic facial attribute editing models, there is still a tremendous amount of work to do to achieve ideal models for this purpose. In this section, the most important challenges of the face attribute editing tasks along with the limitations of the current state-of-the-art models in facing these challenges are discussed. 

\subsection{Attribute entanglement}
Attribute entanglement is one of the most well-studied challenges of face image generating and facial attribute manipulation methods. In this phenomenon, altering the value of an attribute through a model also changes other attributes. For example, by requesting a model to increase the age of a person,  glasses are also added to the image. This attribute entanglement either comes from the biological relation between the attributes (e.g. race and skin color) or is a result of the biases present in the datasets used for training the models (e.g. gender and smile) \cite{han2021disentangled}. 

Regardless of the origin of these connections between attributes, it is desired to have models capable of independently manipulating each facial attribute. Learning disentangled latent representation for face images has been a hot topic during the past years, and many of the works studied in the previous sections claim to have this property. However, attribute entanglement remains a demanding challenge in this realm.

\subsection{Uncommon face conditions}
A close look at the facial attribute editing models reviewed in this paper reveals that most of these models can be applied only to face images similar to their training images. Most of the images of these datasets are captured in common conditions including a frontal pose without extreme pose or expressions, and not severely covered by body parts or external objects. As a result, when you ask the current stat-of-the-art models to add smile to a side view picture (profile shot) of a person the outcome is not satisfactory. The same happens for any unusual face photo with few or non similar images in the training dataset. 
Except for some special models like FaceShifter \cite{li2019faceshifter}, the face attribute manipulation models can not separate the occluding objects from the face behind them and produce degenerated images in the presence of occluding objects.

\subsection{Low data resources}
As mentioned earlier, almost all of the available facial attribute editing models employ deep data-hungry neural networks as their encoders and decoders. The whole knowledge of these models about the human face comes from their training data and their generated images follow the statistical distribution of these data. A complete list of the datasets used for training these models was provided in Table \ref{tabel:dataset:datasets}. These datasets suffer from different shortcomings. First of all, these datasets are not diverse enough, for example, many human races are not covered or some combinations of attributes are not present in these datasets. The other drawback of the current datasets is their innate biases. For example, most of the sad face photos belong to the kids. Moreover, the attribute tags are imbalanced for example there are a large number of smiling and neutral faces while there are very small angry faces. As much of these datasets are collected from the web or under the common situations, the unusual face statuses discussed in the previous subsection like extreme poses and expressions and occluding objects are not covered by them. Finally, the number of images and the quality of these images are two important factors of a proper dataset for training facial attribute editing models.

\subsection{Evaluation metrics}
A number of quantitative and qualitative measures used for evaluating the performance of facial attribute editing models were reviewed in Section \ref{sec4}. The qualitative metrics are based on human judgements and the quantitative measures either estimate the quality of the generated image or the rate of the presence/absence of the target attribute. In fact, all of these measures are some proxies for the required evaluation metric, i.e., how good is the requested edit performed?
The investigation of the results by humans is still the most common metric used in this field and no quantitative special-purpose metric has been proposed for evaluating the outcome of a face attribute editing model.

\subsection{Quality of the edited/generated images}
Despite the great improvements in face generation and manipulation methods, the resolution and quality of the generated images are not still enough for many real-world applications\cite{yi2020animating}. Most of the studied models perform edit on low-resolution images. For example, the size of the input images of
ICface \cite{tripathy2020icface} and AttGAN \cite{he2019attgan} are $128\times 128$ and $384\times 384$, respectively. Most of the other models with higher output resolutions, e.g., \cite{wu2021stylespace}, are based on the pre-trained face generators like StyleGAN. Moreover, the images generated by these methods are still prone to various artifacts and deficiencies.
Improving the resolution and quality of the edited images requires larger datasets and higher computational power for training the models.

\section{Conclusion} \label{sec6}
Facial attribute editing has received great attention during the past years due to its wide range of applications in different areas. Following the recent successes in generative adversarial networks and deep neural networks, a large number of models have been proposed for semantic facial attribute editing. In this paper, we provided a comprehensive survey of these work and tried to investigate them from different perspectives. In these models, the target attributes are usually given by an attribute vector or a driving image. Each of the proposed models generally follow one of the three main architectures of this field, namely encoder-decoder, image-to-image translation, and photo-guided methods. The models are trained on paired or unpaired datasets using a diverse set of loss functions including reconstruction, attribute classification, and adversarial losses. Despite invaluable achievements, the current state-of-the-art models yet are unable to overcome all challenges of the semantic editing task. Low resolution and image quality, attribute entanglement, and working only on common face images are the most important of the drawbacks of the current models. Preparing large unbiased labeled datasets of facial images, defining special-purpose metrics for evaluation, and learning disentangled representations seems to be important requirements of the current models.  

\bibliographystyle{cas-model2-names}

\bibliography{main}

\end{document}